\documentclass{article}

\pdfoutput=1

\usepackage{PRIMEarxiv}

\usepackage[utf8]{inputenc} 
\usepackage[T1]{fontenc}    
\usepackage{hyperref}       
\usepackage{url}            
\usepackage{booktabs}       
\usepackage{amsfonts}       
\usepackage{nicefrac}       
\usepackage{microtype}      
\usepackage{lipsum}
\usepackage{fancyhdr}       
\usepackage{graphicx}       
\usepackage[linewidth=0.5pt]{mdframed}

\graphicspath{{media/}}     

\usepackage{amsmath,amsfonts,bm}
\usepackage{amsthm}  
\usepackage{multirow}
\usepackage{subcaption} 
\usepackage{amssymb}
\usepackage{wrapfig}

\newtheorem{theorem}{Theorem}                   

\newtheorem{lemma}[theorem]{Lemma}

\def\eqref#1{equation~\ref{#1}}

\def\1{\bm{1}}

\DeclareMathAlphabet{\mathsfit}{\encodingdefault}{\sfdefault}{m}{sl}
\SetMathAlphabet{\mathsfit}{bold}{\encodingdefault}{\sfdefault}{bx}{n}

\title{Knowledge-Aware Evolution for Task-Free Streaming Federated Continual Learning with Arbitrary Class Overlap}

\author{
  Sixing Tan\thanks{First author} \\
  Faculty of Computing \\
  Harbin Institute of Technology \\
  \texttt{hit\_tsx@163.com}\\
   \And
  Xianmin Liu\thanks{Corresponding author}\\
  Faculty of Computing \\
  Harbin Institute of Technology \\
  \texttt{liuxianmin@hit.edu.cn}\\
}

\begin{document}
\maketitle

\begin{abstract}
  Federated Continual Learning (FCL) leverages inter-client collaboration to better balance new knowledge acquisition and old knowledge retention on non-stationary data.
  However, existing FCL methods struggle to adapt to streaming scenarios where sequential and ephemerally accessible data chunks lack task identifiers and exhibit arbitrary class overlap, leading to confusion between old and new knowledge and an inability to sustain local inference on all encountered classes.
  To address this, we propose FedKACE with three components: 
  1) an adaptive mechanism that determines when to switch the inference model from the local to the global one to improve client-side inference performance; 
  2) a responsive gradient-balanced replay scheme that utilizes the ratio of the squared L2 gradient norms to balance client-specific knowledge between new acquisition and old retention; 
  3) a holistic buffer maintenance strategy that preserves highly informative and boundary-significant samples to enhance knowledge retention under class overlap.
  Experiments across multiple scenarios and theoretical analysis demonstrate the effectiveness of FedKACE.
\end{abstract}

\section{Introduction}
Federated learning (FL) \cite{21} enables multiple clients to collaboratively train models without sharing raw data, yet its static data assumption leads to severe performance degradation when faced with dynamic class-incremental scenarios or distribution shifts \cite{28}.
To address this, federated continual learning (FCL) \cite{31} extends FL to such dynamic environments via continual learning mechanisms, more effectively balancing new knowledge acquisition with prior knowledge retention through cross-client collaboration \cite{30}.

However, in real-world settings, edge clients with constrained storage receive sequential data chunks exhibiting class overlap and lacking task identifiers, where the continuous data influx forces the discarding of each chunk after limited iterations to avoid memory overflow \cite{29}.
This reality exposes two limitations of traditional FCL methods by violating their core assumptions:
First, class overlap and missing task identifiers cause confusion between old and new knowledge \cite{38}, breaking the premise of mutually exclusive class sets across tasks.
Second, the streaming data paradigm not only limits iterations per data chunk but also requires inference on all seen classes \cite{39}, contradicting the reliance on static intra-task data for tasks spanning multiple FL rounds.
These challenges expose key limitations of existing FCL methods \cite{30}: 
sample replay methods struggle to balance old and new samples within the same class in constrained buffers; 
generative replay methods suffer from poor generation quality with restricted iterations;
and parameter isolation methods overallocate parameters and inflate model size without explicit task boundaries. All of the above collectively limit practical FCL deployment.

To address these limitations in FCL, we focus on the streaming FCL setting where clients receive data chunks that are non-overlapping across FL rounds, are accessible solely during their respective FL rounds, and exhibit arbitrary class overlap.
This setting poses three core challenges for clients: (1) knowledge confusion from class overlap across data chunks without task identifiers; (2) unknown, dynamically evolving data heterogeneity among clients; (3) the requirement for real-time inference on all encountered classes per FL round.
Preserving the low training complexity of sample replay, we propose the \textbf{Fed}erated \textbf{K}nowledge-\textbf{A}ware \textbf{C}ontinual \textbf{E}volution (FedKACE) framework. 
First, an adaptive mechanism determines when to switch the inference model from the local to the global one, motivated by their distinct convergence dynamics, implemented by monitoring the rate of change in the discrepancy between the global model's average accuracy and prediction confidence evaluated on the local buffer, to improve client-side inference performance.
Second, a responsive gradient-balanced replay scheme that dynamically adjusts loss weights for buffer samples, based on the ratio of the squared L2 norms of buffer samples' gradients to those from new samples, to achieve client-specific balance of knowledge between new acquisition and old retention.
Finally, a holistic buffer maintenance strategy that jointly optimizes the informativeness, decision boundary significance, and feature space dispersion of buffer samples, through a two-stage screening process, to enhance knowledge retention under class overlap scenarios.

Our contributions are as follows:

1. We extend FCL to the streaming FCL setting, which exhibits arbitrary class overlap between data chunks, lacks task identifiers, and requires real-time inference on all encountered classes after each FL round.

2. We propose FedKACE framework, integrating an adaptive inference model switching mechanism to achieve real-time inference, along with a responsive gradient-balanced replay scheme and a holistic buffer maintenance strategy to improve knowledge retention under arbitrary class overlap and without task identifiers.

3. Experiments across diverse scenarios and theoretical analysis validate the effectiveness of FedKACE.

\section{Related Works}
\subsection{Federated Continual Learning}
Federated Continual Learning (FCL) extends FL to adapt to dynamic data distributions while balancing stability and plasticity. 
Existing FCL methods fall into five types:
Rehearsal (Sample Replay) methods \cite{24,25,26} mitigate forgetting by storing key samples but struggle to balance the representation of old and new samples for overlapping classes; 
Generative Replay methods \cite{32,33} avoid raw data storage yet are constrained by generation quality and computational costs; 
Parameter Isolation methods \cite{22,34} reduce storage needs but require explicit task boundaries and inflate the model size;
Regularization methods \cite{35,36} require no extra storage but assume similar class sets across clients at each time step;
Dynamic Collaboration methods \cite{37,23} incur no additional storage overhead yet depend on clients' class sets jointly cover all possible classes per FL round.
Moreover, most FCL methods assume explicit task identifiers, full data availability for each task throughout its multiple FL rounds, and mutually exclusive class sets across tasks, limiting their ability to handle knowledge retention in real-world settings like streaming FCL.

\subsection{Online Continual Learning}
Online Continual Learning (OCL), at the intersection of continual and online learning, processes streaming data restricted to task-specific classes within each task \cite{39}.
Existing OCL methods fall into four types:
Rehearsal (Sample Replay) methods \cite{47,48} retain prior knowledge by storing key samples but are highly dependent on buffer capacity; 
Gradient Optimization methods \cite{49,50,10} balance stability and plasticity by employing specialized update trajectories and thereby incur high computational costs and poor scalability;
Module Enhancement methods \cite{52,53} improve adaptation to non-stationary environments by introducing task-specific modules but require explicit task identifiers;
Regularization methods \cite{54,55} preserve prior knowledge by constraining parameter updates yet heavily rely on initialization from pretrained models.
Moreover, OCL methods struggle on real-world data streams due to explicit task boundary assumptions, and overfit locally skewed data without global regularization, limiting their capacity to retain knowledge in real-world settings like streaming FCL.

\section{Notation and Problem Formulation}
In the streaming FCL setting, we assume static client data within each FL round.  
Consider a $T$-round FL process with a client set $\mathcal{K} = \{1, \dots, K\}$. In each FL round $t \in [T]$, each client $k \in \mathcal{K}$ receives a new data chunk $\mathcal{D}_k^{t} = \{(x_i, y_i)\}_{i=1}^{n_k^t}$ that is accessible solely during the $t$-th FL round, where $n_k^t$ is the number of samples, $x_i$ is an input and $y_i$ is its label from the class set $\mathcal{C}_k^{t}$ of $\mathcal{D}_k^{t}$. 
Moreover, for each client $k$, data chunks are disjoint across FL rounds (i.e., $\mathcal{D}_k^t \cap \mathcal{D}_k^{t'} = \emptyset$ for all $t \neq t'$), yet class sets allow for arbitrary overlap (i.e., $0 \leq |\mathcal{C}_k^t \cap \mathcal{C}_k^{<t}| \leq \min(|\mathcal{C}_k^t|, |\mathcal{C}_k^{<t}|)$), where $\mathcal{C}_k^{<t} = \bigcup_{\tau=1}^{t-1} \mathcal{C}_k^{\tau}$ includes all classes encountered up to FL round $t-1$.

At the global level, let $\mathcal{C}_g = \bigcup_{k = 1}^K \bigcup_{t = 1}^T \mathcal{C}_k^{t}$ denote the global class set with size $|\mathcal{C}_g| = C_{\max}$, we set each client $k \in \mathcal{K}$ to encounter all $C_{\max}$ classes across all $T$ FL rounds with heterogeneous class trajectories, resulting in arbitrary overlap among the local cumulative class sets $\mathcal{C}_k^{\leq t} = \bigcup_{\tau = 1}^t \mathcal{C}_k^{\tau}$ for any intermediate FL round $t < T$ (i.e., $0 \leq |\mathcal{C}_k^{\leq t} \cap \mathcal{C}_{k'}^{\leq t}| \leq \min(|\mathcal{C}_k^{\leq t}|, |\mathcal{C}_{k'}^{\leq t}|)$ for all $k \neq k'$), thereby introducing complementary knowledge among clients. 

Formally, we define the model $\theta = \{\phi, h\}$, where $\phi$ is the feature extractor and $h$ is the linear layer of fixed output dimension $C_{\max}$. 
In FL round $t$, client $k$ activates the output dimensions of its local model $\theta_k^{t} = \{\phi_k^{t}, h_k^{t}\}$ or the global model $\theta_g^{t} = \{\phi_g^{t}, h_g^{t}\}$ corresponding to $\mathcal{C}_{k}^{\leq t}$ through masking. 
The objective for each client $k \in \mathcal{K}$ in the $t$-th FL round is to optimize the inference model $\tilde{\theta}_k^{t} = \{\tilde{\phi}_k^{t}, \tilde{h}_k^{t}\}$ (i.e., $\theta_k^{t}$ or $\theta_g^{t}$) for the local cumulative class set $\mathcal{C}_k^{\leq t}$, formalized as:
\begin{equation}
    \min_{\tilde{\theta}_k^{t}} \mathcal{F}(\tilde{\theta}_k^{t}) =  \sum_{c \in \mathcal{C}_{k}^{\leq t}}
    \underset{x \sim \mathcal{P}_{g,c}}{\mathbb{E}} \left[ l\left( f_{\tilde{\theta}_k^{t}}(x), c \right) \right] + \mathcal{R}_k^{t}(\tilde{\theta}_k^{t})
\end{equation}
where $f_{\tilde{\theta}_k^{t}}(x) = \tilde{h}_k^{t}(\tilde{\phi}_k^{t}(x))$ is the prediction function, $\mathcal{P}_{g,c}$ is the global class-conditional data distribution for class $c$, $l$ is the loss function, and $\mathcal{R}_k^{t}(\cdot)$ is a federated regularization term applied to $\tilde{\theta}_k^{t}$ to enhance global generalization.

\section{Methodology}
\subsection{Adaptive Inference Model Switching Mechanism}
\label{Section 4.1}
In FL, local and global models exhibit distinct performance trajectories on local data: local models outperform the global model in early FL rounds but overfit local data (or local buffer) more rapidly while the global model achieves better generalization via global regularization over time.
Inspired by \cite{18}, we propose that each client $k \in \mathcal{K}$ defaults to the local model for inference and then adaptively switches to the global model, based on the rate of change in the discrepancy between the global model's average accuracy $\mathrm{ACC}_{k,\mathrm{BF}}^{t}$ and the average prediction probability of the ground-truth class $\mathrm{PROB}_{k,\mathrm{BF}}^{t}$, both computed on the representative local buffer $\mathcal{M}_k^{t}$ after client $k$ receives the global model $\theta_g^{t}$ from the server at the end of FL round $t$:
\begin{equation}
  \mathrm{ACC}_{k,\mathrm{BF}}^{t} = \frac{1}{\left| \mathcal{M}_k^{t} \right|} \sum_{(x,y) \in \mathcal{M}_k^{t}} \mathbb{I}\left( \hat{y}_g^{t}(x) = y \right);
  \quad
   \mathrm{PROB}_{k,\mathrm{BF}}^{t} = \frac{1}{\left| \mathcal{M}_k^{t} \right|} \sum_{(x,y) \in \mathcal{M}_k^{t}} z_g^{t}(x)[y]
\end{equation}
where $z_g^{t}(x)$ and $\hat{y}_g^{t}(x)$ denote the normalized logits and the predicted label for input $x$ generated by $\theta_g^{t}$, respectively. 
Then, we define the generalization gap $\mathrm{GAP}_k^{t}$ between $\mathrm{ACC}_{k,\mathrm{BF}}^{t}$ and $\mathrm{PROB}_{k,\mathrm{BF}}^{t}$ and its rate of change $\Delta \mathrm{GAP}_k^{t}$:
\begin{equation}
  \mathrm{GAP}_k^{t} = \max(0, \mathrm{ACC}_{k,\mathrm{BF}}^{t} - \mathrm{PROB}_{k,\mathrm{BF}}^{t});
  \quad
   \Delta\mathrm{GAP}_k^{t} = \mathrm{GAP}_k^{t} - \mathrm{GAP}_k^{t-1}, \quad \forall t > 1
\end{equation}
When $\Delta\mathrm{GAP}_k^{t}<0$, the gap between $\mathrm{ACC}_{k,\mathrm{BF}}^{t}$ and $\mathrm{PROB}_{k,\mathrm{BF}}^{t}$ is decreasing on the local buffer, indicating that the global model has converged sufficiently to adapt to client $k$'s local data distribution. 
To ensure robustness, client $k$ switchs its inference model to the global model after this condition holds for two consecutive FL rounds, with the switching FL round $t_{k}^{\mathrm{sw}}$ defined as:
\begin{equation}
  t_{k}^{\mathrm{sw}} = \mathrm{min}\left\{t \mid \Delta\mathrm{GAP}_k^{t} < 0 \land \Delta\mathrm{GAP}_k^{t-1} < 0 \right\}
\end{equation}
When $t < t_{k}^{\mathrm{sw}}$, client $k$ uses the local model $\theta_k^{t}$ for inference, while for $t \geq t_{k}^{\mathrm{sw}}$, it uses the global model $\theta_g^{t}$ for inference and stops computing $\mathrm{ACC}_{k,\mathrm{BF}}^{t}$ and $\mathrm{PROB}_{k,\mathrm{BF}}^{t}$ to reduce computational cost.
The subsequent Theorem 3 proves the necessity of switching the inference model.

\subsection{Responsive Gradient-Balanced Replay Scheme (RGBRS)}
\label{Section 4.2}
In streaming FCL, heterogeneous data distributions across clients lead to divergent parameter update directions and magnitudes, rendering fixed knowledge retention strategies unable to adapt to varying client learning dynamics.
Due to the lack of task identifiers, we propose that each client treats samples in its representative local buffer as a unified historical task.
In FL round $t$, client $k$ receives a new data chunk $\mathcal{D}_k^{t}$ with class set $\mathcal{C}_k^t$ and accesses its previous local buffer $\mathcal{M}_k^{t-1}$ with historical class set $\mathcal{C}_{k}^{<t} = \bigcup_{\tau=1}^{t-1} \mathcal{C}_k^{\tau}$, exhibiting arbitrary class overlap such that $0 \leq |\mathcal{C}_k^t \cap \mathcal{C}_k^{<t}| \leq \min(|\mathcal{C}_k^t|, |\mathcal{C}_k^{<t}|)$.
For each epoch $j \in [1,J]$, given the local model $\theta_k^{t,j-1}$ to be updated and the cross-entropy loss function $l(\cdot; \cdot)$, the task loss $L_{k,\text{task}}^{t}(\theta_k^{t,j-1})$ on $\mathcal{D}_k^{t}$ and the replay loss $L_{k,\text{rep}}^{t}(\theta_k^{t,j-1})$ on $\mathcal{M}_k^{t-1}$ are: 
\begin{equation}
  L_{k,\text{task}}^{t}(\theta_k^{t,j-1}) = \sum_{(x,y) \in \mathcal{D}_k^{t}} l\left(f_{\theta_k^{t,j-1}}(x)[\mathcal{C}_k^{t}]; y \right)
\end{equation}
\begin{equation}
  L_{k,\text{rep}}^{t}(\theta_k^{t,j-1}) = \sum_{(x,y) \in \mathcal{M}_k^{t-1}} l\left(f_{\theta_k^{t,j-1}}(x)[\mathcal{C}_{k}^{<t}]; y \right)
\end{equation}
where $[\mathcal{C}_k^{t}]$ and $[\mathcal{C}_{k}^{<t}]$ denote the output masks that restrict predictions to the respective classes. Given an adaptive replay loss weight $\lambda_k^{t,j}$, the total training loss in epoch $j$ is:
\begin{equation}
  L_{k, \text{total}}^{t}(\theta_k^{t,j - 1}) = L_{k,\text{task}}^{t}(\theta_k^{t,j - 1}) + \lambda_k^{t,j} \cdot L_{k,\text{rep}}^{t}(\theta_k^{t,j - 1})
\end{equation}
Then, client $k$ update $\theta_k^{t,j} = \theta_k^{t,j-1} - \eta \cdot \nabla_{\theta}L_{k, \text{total}}^{t}(\theta_k^{t,j - 1})$ with learning rate $\eta$, completing the $j$-th epoch.
For the adaptive replay loss weight $\lambda$, given that the squared L2 norms of gradients quantify parameter change magnitudes \cite{09}, we propose that client $k$ computes the ideal weight $\tilde{\lambda}_k^{t,j+1}$ in epoch $j+1$ as the ratio of the squared L2 norms of the replay and task loss gradients, evaluated at $\theta_k^{t,j-1}$ during epoch $j$: 
\begin{equation}
  \tilde{\lambda}_k^{t,j+1} = \| \nabla_{\theta} L_{k,\text{rep}}^{t}(\theta_k^{t,j - 1})\|_2^2 / \|\nabla_{\theta} L_{k,\text{task}}^{t}(\theta_k^{t,j - 1})\|_2^2 \!
\end{equation}
This weight naturally increases to strengthen constraints for knowledge retention when buffer gradients dominate, and decreases to relax these constraints when new data gradients prevail.
However, computing the full-model gradients for both replay and task losses incurs two full backward passes. Following \cite{10}, we use the output layer $h$ as a proxy, and define the actual weight $\lambda_k^{t,j+1}$ for epoch $j+1$ as:
\begin{equation}
  \lambda_k^{t,j+1} = \|\nabla_{h} L_{k,\text{rep}}^{t}(\theta_k^{t,j - 1})\|_2^2 / \|\nabla_{h} L_{k,\text{task}}^{t}(\theta_k^{t,j - 1})\|_2^2 \!
\end{equation}
This approximation replaces full backward passes with output-layer computations to reduce overhead, with gradient norms computed from running averages of batch gradients within each epoch.
In the first epoch, each client $k \in \mathcal{K}$ initializes its local model $\theta_k^{t,0} = \theta_g^{t-1}$ and sets $\lambda_k^{t,1} = 1$.

Building on the Responsive Gradient-Balanced Replay Scheme, we formally state the \hyperref[Theorem 1]{Theorem 1}:

\setcounter{theorem}{0}   
\begin{theorem}[Local Saddle-Point Convergence of RGBRS]
   \label{Theorem 1}
   Under standard assumptions (see \hyperref[Appendix B.1]{Appendix B.1}), let $\theta_k^{t,0}$ be the initial model for client $k$ in FL round $t$. After $J$ local epochs via the Responsive Gradient-Balanced Replay Scheme, $\theta_k^{t,J}$ converges to the ideal saddle point $(\theta_k^{t,*}, \lambda_k^{t,*})$ within a truncation error bound:
   \begin{equation}
   \mathbb{E}[L_{k,\text{total}}^{t}(\theta_k^{t,J}|\lambda_k^{t,J})] 
   \leq L_{k,\text{total}}^{t}(\theta_k^{t,*}|\lambda_k^{t,*})
   + \mathcal{E}_{k, \text{bound}}^{t}(\{\eta_k^{t,j}\}_{j=0}^{J-1}, \|\theta_k^{t,0} - \theta_k^{t,*}\|_2^2)  
   \end{equation}
   where the ideal saddle point $(\theta_k^{t,*}, \lambda_k^{t,*})$ satisfies the minimax equilibrium property for any $\theta \in \Omega_\theta$ and $\lambda \in \Omega_\lambda$:
   \begin{equation}
      L_{k,\text{total}}^{t}(\theta_k^{t,*}| \lambda) \leq L_{k,\text{total}}^{t}(\theta_k^{t,*}| \lambda_k^{t,*}) \leq L_{k,\text{total}}^{t}(\theta| \lambda_k^{t,*})
   \end{equation}
   $\mathcal{E}_{k, \text{bound}}^{t}\left(\{\eta_k^{t,j}\}_{j=0}^{J-1}, \|\theta_k^{t,0} - \theta_k^{t,*}\|_2^2\right)$ is an error bound functional detailed in \hyperref[Appendix B.3]{Appendix B.3}.
\end{theorem}
The proof of \hyperref[Theorem 1]{Theorem 1} is in \hyperref[Appendix B.3]{Appendix B.3}.

\subsection{Holistic Buffer Maintenance Strategy (HBMS)}
\label{Section 4.3}
In streaming FCL without task identifiers, client buffer constraints pose particular challenges in the presence of class overlap, impeding the balanced representation of identical classes across data chunks.
Guided by kernel spectral boundary theory and information value assessment, we design a two-stage buffer maintenance strategy to maximize boundary-critical knowledge retention.

In FL round $t$, client $k$ updates the global model $\theta_g^{t-1}$ using the new data chunk $\mathcal{D}_k^t$ and its buffer $\mathcal{M}_k^{t-1}$ (with capacity $M$) to obtain the local model $\theta_k^{t,J}$, and then constructs its new buffer $\mathcal{M}_k^t$ from $\mathcal{M}_k^{t-1} \cup \mathcal{D}_k^t$ using $\theta_k^{t,J}$ for the class set $\mathcal{C}_{k}^{\leq t} = \mathcal{C}_{k}^{<t} \cup \mathcal{C}_k^t$: First, client $k$ computes the normalized logit vector $\hat{g}_k^t(x) = f_{\theta_k^{t,J}}(x) / \|f_{\theta_k^{t,J}}(x)\|_2$ for each sample $x \in \mathcal{M}_k^{t-1} \cup \mathcal{D}_k^t$, and next: 

\textbf{Computing Kernel Diversity and Class Metrics}: 
Maximizing the minimum distance of candidate samples to the current buffer enhances feature dispersion and prioritizes boundary samples to ensure significance and informativeness, reducing the kernel matrix condition number and hypothesis space complexity.
To achieve this, we first define a Gaussian kernel function $K(x, x_i)$ that captures the spatial distribution of $\hat{g}_k^t(x)$ via its smoothness and local sensitivity:
\begin{equation}  
  K(x, x_i) = \exp\left(-\beta \|\hat{g}_k^t(x) - \hat{g}_k^t(x_i)\|_2^2 \right)  
\end{equation}  
where $\beta$ controls the kernel decay rate, defined as $|\mathcal{M}_k^{t-1}|^{2/d}$ for effective class dimension $d = |\mathcal{C}_{k}^{\leq t}|$ if $\mathcal{M}_k^{t-1} \neq \emptyset$, and $1$ otherwise. 
Then, client $k$ computes the kernel diversity score (DS) for each $x \in \mathcal{M}_k^{t-1} \cup \mathcal{D}_k^{t}$ with respect to $\mathcal{M}_k^{t-1}$:
\begin{equation}  
  \begin{split}
    \text{DS}(x) 
    & \underset{\text{objective}}{\triangleq} - \log \left( \max_{x_i \in \mathcal{M}_k^{t-1}} K(x, x_i) \right)
    \\
    &\underset{\text{computed}}{\propto} \min_{x_i \in \mathcal{M}_k^{t-1}} \|\hat{g}_k^t(x) - \hat{g}_k^t(x_i)\|_2^2 
  \end{split}
\end{equation}  
If $\mathcal{M}_k^{t-1} = \emptyset$, $\text{DS}(x) = 0$ due to the absence of historical samples. 
Then, for each sample $x \in \mathcal{M}_k^{t-1} \cup \mathcal{D}_k^t$ with hard label $c_x$, if $c_x \in \mathcal{C}_{k}^{<t}$, client $k$ uses $K(x, x_i)$ to compute the conditional predictive distributions before and after its inclusion in the buffer, by reusing the pre-computed pairwise distances $\|\hat{g}_k^t(x) - \hat{g}_k^t(x_i)\|_2^2$ obtained when computing DS: 
\begin{equation}
  \bar{p}_{\mathcal{M}_k^{t-1}}(c_x \mid x) = A(x, c_x) / B(x);
  \quad
  \! \bar{p}_{\mathcal{M}_k^{t-1} \cup \{x\}}(c_x \mid x) = \frac{A(x, c_x) + K(x,x) \cdot p(c_x \mid x)}{B(x) + K(x,x)} \!
\end{equation}
where $A(x, c_x) = \sum_{x_i \in \mathcal{M}_k^{t-1}} K(x, x_i) \cdot p(c_x \mid x_i)$ with $p(c_x \mid x)$ being the predicted probability for class $c_x$ under $\theta_k^{t,J}$, and $B(x) = \sum_{x_i \in \mathcal{M}_k^{t-1}} K(x, x_i)$. 
If $\mathcal{M}_k^{t-1} = \emptyset$ or new classes $c_x \in \mathcal{C}_k^t \setminus \mathcal{C}_{k}^{<t}$, both $\bar{p}_{\mathcal{M}_k^{t-1}}(c_x \mid x)$ and $\bar{p}_{\mathcal{M}_k^{t-1} \cup \{x\}}(c_x \mid x)$ are set to $p(c_x \mid x)$ due to the lack of prior information. 
Accordingly, client $k$ computes the Information-Diversity Value (IDV) and Consistency-Diversity Value (CDV) for each sample $x \in \mathcal{M}_k^{t-1} \cup \mathcal{D}_k^t$. When $c_x \in \mathcal{C}_{k}^{<t}$, IDV and CDV are defined as: 
\begin{equation}
  \mathrm{IDV}(x \mid c_x) = -\log \bar{p}_{\mathcal{M}_k^{t-1}}(c_x \mid x) + \lambda_1 \cdot \mathrm{DS}(x)
\end{equation}
where the adaptive factors $\lambda_1 = \log |\mathcal{M}_k^{t-1}|/(|\mathcal{M}_k^{t-1}|)^{1/2}$ and $\lambda_2 = 1/(|\mathcal{M}_k^{t-1}|)^{1/2}$ prioritize diversity for small buffers and informativeness for large buffers, with $\lambda_1 = \lambda_2 = 0$ if $\mathcal{M}_k^{t-1} = \emptyset$. 
If $\mathcal{M}_k^{t-1} = \emptyset$ or new classes $c_x \in \mathcal{C}_k^t \setminus \mathcal{C}_{k}^{<t}$, we set $\mathrm{IDV}(x \mid c_x) = -\log p(c_x \mid x) + \lambda_1 \cdot \mathrm{DS}(x)$ and $\mathrm{CDV}(x \mid c_x) = 0$ due to the lack of prior information.

\textbf{Maintaining Buffer via Two-Stage}: 
In our method, buffer capacity $M$ is evenly distributed to form subsets $\mathcal{M}_{k,c}^{t}$ for each class $c \in \mathcal{C}_{k}^{\leq t}$, each of size $\lfloor M / {|\mathcal{C}_{k}^{\leq t}|}\rfloor$, where samples are sequentially selected from the candidate set $X_c = \{x \in \mathcal{M}_k^{t-1} \cup \mathcal{D}_k^t \mid y_x = c\}$ of class $c$ to retain high-uncertainty samples using the IDV metric, followed by boundary-critical ones guided by the CDV metric. 
For old classes $c \in \mathcal{C}_{k}^{<t}$, client $k$ selects the top-$\min\{2|\mathcal{M}_{k,c}^{t}|, |X_c|\}$ samples from $X_c$ ranked by $\mathrm{IDV}(x|c)$ to form a candidate set $\tilde{\mathcal{M}}_{k,c}^{t}$, and then retains the top-$|\mathcal{M}_{k,c}^{t}|$ samples ranked by $\mathrm{CDV}(x|c)$ as the buffer $\mathcal{M}_{k,c}^{t}$.
For new classes $c \in \mathcal{C}_k^t \setminus \mathcal{C}_{k}^{<t}$ where $\mathrm{CDV}(x \mid c) = 0$ for all $x \in X_c$, client $k$ forms $\mathcal{M}_{k,c}^{t}$ by selecting $|\mathcal{M}_{k,c}^{t}|$ samples via information-weighted sampling without replacement, with probability $P(x) = \exp(\mathrm{IDV}(x|c))/\sum_{x' \in X_c}\exp(\mathrm{IDV}(x'|c))$.
Finally, client $k$ constructs the complete buffer $\mathcal{M}_k^{t} = \bigcup_{c \in \mathcal{C}_{k}^{\leq t}} \mathcal{M}_{k,c}^{t}$.
Building on the Holistic Buffer Maintenance Strategy, we formally state the \hyperref[Theorem 2]{Theorem 2}:

\setcounter{theorem}{1}   
\begin{theorem}[Spectral Efficiency of HBMS]
   \label{Theorem 2}
   Given $M > 2C_{\max}$ and standard assumptions (see \hyperref[Appendix B.1]{Appendix B.1}), 
   let $C_{\mathrm{rand}} \triangleq 1$ be the spectral efficiency constant of class-wise random sampling, and let $C_\kappa \triangleq \inf_{\theta \in \Omega_\theta, \hat{g} \in \mathcal{G}} \exp(-\beta(\theta, \hat{g}) \Delta \rho(\theta, \hat{g}))$ (depending only on $M$ and $C_{\max}$) be that of our Holistic Buffer Maintenance Strategy, where $\Delta \rho(\theta, \hat{g}) \triangleq \rho_{\mathrm{rand}} - \rho_{\mathrm{HBMS}}$ is the spectral radius gap between the two methods.
   Then, $0 < C_\kappa < C_{\mathrm{rand}}$ holds for all $k \in \mathcal{K}$ and $t \in [T]$.
\end{theorem}
The proof of \hyperref[Theorem 2]{Theorem 2} is in \hyperref[Appendix B.4]{Appendix B.4}.

\begin{figure}[tbp]
  \begin{center}
    \centering
    \includegraphics[width=\linewidth]{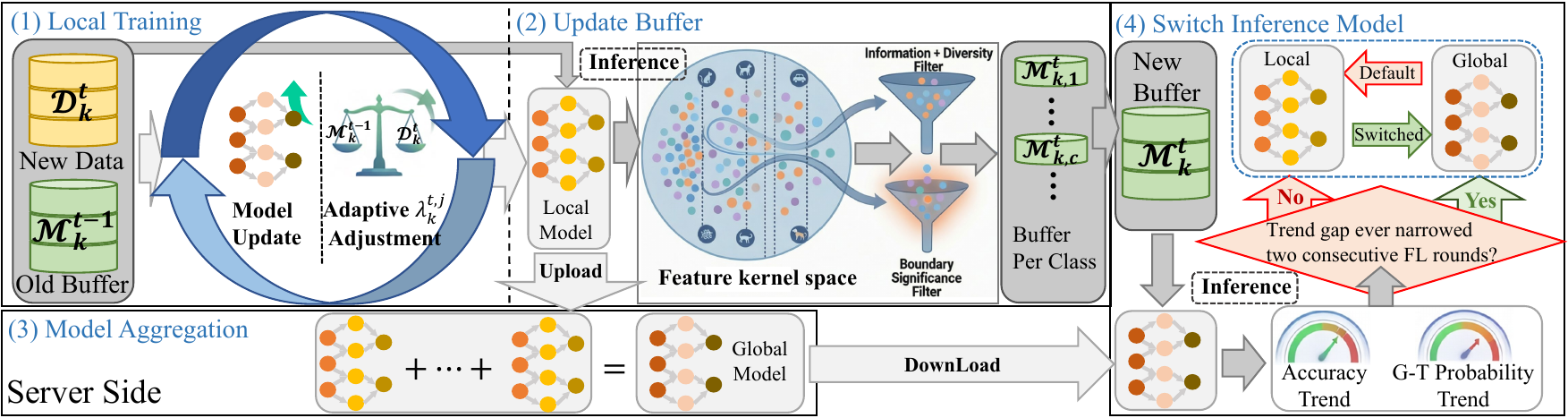}
    \label{Figure 1}
    \vskip -0.1in
    \caption{
      Illustration of the FedKACE framework. 
      In FL round $t$, client $k$ trains locally via the \textbf{Responsive Gradient-Balanced Replay Scheme} (see \hyperref[Section 4.2]{Sec. 4.2}), 
      then updates its buffer via the \textbf{Holistic Buffer Maintenance Strategy} (see \hyperref[Section 4.3]{Sec. 4.3}), 
      subsequently uploads the local model for aggregation,  
      and finally evaluates the new global model via the \textbf{Adaptive Inference Model Switching Mechanism} (see \hyperref[Section 4.1]{Sec. 4.1}).
    }
  \end{center}
  \vskip -0.2in
\end{figure}

\subsection{FedKACE framework}
Based on the above, we propose the FedKACE framework: 
In each FL round $t$, each client $k \in \mathcal{K}$ updates the previous global model $\theta_g^{t-1}$ on its buffer $\mathcal{M}_k^{t-1}$ and new data chunk $\mathcal{D}_k^t$ via the Responsive Gradient-Balanced Replay Scheme and obtains $\theta_k^{t,J}$. 
Then, client $k$ updates its buffer to $\mathcal{M}_k^t$ from $\mathcal{D}_k^t \cup \mathcal{M}_k^{t-1}$ via the Holistic Buffer Maintenance Strategy. 
Next, client $k$ uploads $\theta_k^{t,J}$ for server aggregation, yielding the global model  $\theta_g^t = \frac{1}{K}\sum_{k=1}^K \theta_k^{t,J}$.
Finally, using the Adaptive Inference Model Switching Mechanism, client $k$ evaluates $\theta_g^t$ on $\mathcal{M}_k^t$ to decide whether to switch its inference model from $\theta_k^{t,J}$ to $\theta_g^t$, thereby concluding the $t$-th FL round. 
Building on the FedKACE framework, we formally state the \hyperref[Theorem 3]{Theorem 3}:

\setcounter{theorem}{2} 
\begin{theorem}[Regret Gap between Global and Local Models in FedKACE]
   \label{Theorem 3}
   Under our problem formulation and standard assumptions (see \hyperref[Appendix B.1]{Appendix B.1}), 
   let the optimal model for client $k$ under the local cumulative data distribution $p_k^{\leq t}$ be $f^{t,*}_{k,\mathrm{cum}} \triangleq \arg\min_{\theta \in \Omega_\theta} \mathbb{E}_{p_k^{\leq t}}[\mathcal{R}(f_\theta)]$, 
   and define $\mathrm{Regret}_k^{t}(\theta) \triangleq \mathbb{E}_{p_k^{\leq t}}\left[\mathcal{R}(f_{\theta}) - \mathcal{R}(f^{t,*}_{k,\mathrm{cum}})\right]$ as the regret of model $\theta$ on client $k$ at FL round $t$,  
   the regret gap between the local model $\theta_k^{t,J}$ and the global model $\theta_g^t$ satisfies:
   \begin{equation}
      \begin{split}
         \underset{k \in \mathcal{K}}{\mathbb{E}}
         \left[ \mathrm{Regret}_k^{t}(\theta_g^t) - \mathrm{Regret}_k^{t}(\theta_k^{t,J}) \right]
         \leq \frac{4(K-1)}{K} \Delta^{\mathrm{TV}}(t, \nu) - \frac{\mu}{2K}\sum_{k=1}^K \|\theta_k^{t,J} - \theta_g^t\|_2^2
      \end{split}
   \end{equation}
   where $\Delta^{\mathrm{TV}}(t, \nu)$ globally bounds the cross-client TV distance $\|p_k^{\leq t} - p_{k'}^{\leq t}\|_{\mathrm{TV}}$ for all $k, k' \in \mathcal{K}$ at FL round $t$, decaying to a constant as $t$ increases, with definition detailed in \hyperref[Appendix B.7]{Appendix B.7}. $\nu = n_{\min}/n_{\max}$ is the ratio of the minimum to the maximum per-round data size, defined as $n_{\min} \triangleq \min_{k \in \mathcal{K}, \tau \leq T}(n_k^\tau)$ and $n_{\max} \triangleq \max_{k \in \mathcal{K}, \tau \leq T}(n_k^\tau)$.
   For small $t$, $\Delta^{\mathrm{TV}}(t, \nu)$ dominates, indicating that the local model $\theta_k^{t,J}$ is superior; for large $t$, the negative model discrepancy term dominates, favoring the global model $\theta_g^t$.
\end{theorem}
The proof of \hyperref[Theorem 3]{Theorem 3} is in \hyperref[Appendix B.7]{Appendix B.7}.

\section{Experiments}
\subsection{Experiment Settings}
\textbf{Datasets}:
On CIFAR100 \cite{19} and ImageNet100 (first 100 classes of ILSVRC2012 \cite{20}), we conduct experiments with $C_{\max}=100$ global classes, total $K=10$ clients and total $T=100$ FL rounds.
Before training, each client independently and randomly permutes all 100 classes into a cyclic list, and pre-defines its per-round class sets via a sliding window of size 5 with stride $5-O$, where $O \in \{5,4,2,0\}$ (i.e., \textbf{O}verlap) denotes the class overlap between adjacent FL rounds.
When $O=5$, the window remains fixed for 5 consecutive FL rounds and then advances by 5 classes, simulating static class sets in OCL. 
For $O\in\{4,2,0\}$, the window strides by $1,3,5$ classes per FL round along the cyclic list with wrap-around, thereby modeling varying class overlap across consecutive FL rounds.
In each FL round, every client trains for $J=20$ epochs on 5 window-determined classes using $100$ samples (i.e., $(x,y)$ pairs) per class without explicit task identifiers. 
The local data are non-overlapping for each client across all its FL rounds, but the finite dataset size causes cross-client sample overlap.

\textbf{Evaluation Metric}:
As the forgetting rate is inapplicable without task identifiers, we adopt accuracy (ACC) and regret (REG) from online learning as metrics. 
After each FL round $t$, each client computes its accuracy on the standard test set of its classes encountered thus far, and the $ACC^t$ is averaged across all clients.
To compute regret, we establish a Centralized baseline as an upper bound, where clients independently train newly initialized models for $J=20$ epochs on all samples from FL rounds $1$ to $t$ without global aggregation. The regret for an evaluated method is then defined as $\text{REG}_{\text{method}}^{t} = \text{ACC}_{\text{Centralized}}^{t} - \text{ACC}_{\text{method}}^{t}$, denoting the accuracy gap between the method and the Centralized baseline.
The overall evaluation metrics for a method are Average Accuracy (AA), defined as $\text{AA}_{\text{method}} = \frac{1}{T}\sum_{t=1}^{T}\text{ACC}_{\text{method}}^{t}$, and Average Regret (AR), defined as $\text{AR}_{\text{method}} = \frac{1}{T}\sum_{t=1}^{T}\text{REG}_{\text{method}}^{t}$.

\textbf{Baseline Methods}:
We evaluate seven baselines: the classical FL method FedAVG \cite{21}; five FCL methods, including the parameter isolation-based TFCL \cite{22}, the dynamic collaboration-based DCFCL \cite{23}, and three sample replay-based methods, namely Re-Fed \cite{24}, OFCL \cite{25} and FedCBDR \cite{26}; and the Centralized method as an upper bound, with Re-Fed, OFCL, FedCBDR and our FedKACE employing a local buffer capacity of $M=1000$.
All methods use ResNet-18 \cite{46} as model.
We exclude generative replay and prompt-based FCL methods, as they underperform sample replay-based methods \cite{27}. The remaining FCL baselines use the FL round index as the explicit task identifier.

More details are provided in \hyperref[Appendix A]{Appendix A}.

\subsection{Baseline Experiments}

\setcounter{table}{0}   
\begin{table}[tbp]
   \centering                           
   \caption{Experiment Results of Baseline Methods}
   \label{Table 1}
   \resizebox{\textwidth}{!}{%
      {\footnotesize                      
         \setlength{\tabcolsep}{4pt}
         \begin{tabular}{l||c@{\hspace{4pt}}c|c@{\hspace{4pt}}c|c@{\hspace{4pt}}c|c@{\hspace{4pt}}c||c@{\hspace{4pt}}c|c@{\hspace{4pt}}c|c@{\hspace{4pt}}c|c@{\hspace{4pt}}c}
         \specialrule{1.2pt}{0pt}{0pt}
         \multirow{3}{*}{Method} & \multicolumn{8}{c||}{CIFAR100} & \multicolumn{8}{c}{ImageNet100} \\
         \cline{2-17}
         & \multicolumn{2}{c|}{O=5} & \multicolumn{2}{c|}{O=4} & \multicolumn{2}{c|}{O=2} & \multicolumn{2}{c||}{O=0} & 
         \multicolumn{2}{c|}{O=5} & \multicolumn{2}{c|}{O=4} & \multicolumn{2}{c|}{O=2} & \multicolumn{2}{c}{O=0} \\
         \cline{2-17}
         & AA$\uparrow$ & AR$\downarrow$ & AA$\uparrow$ & AR$\downarrow$ & AA$\uparrow$ & AR$\downarrow$ & AA$\uparrow$ & AR$\downarrow$ & AA$\uparrow$ & AR$\downarrow$ & AA$\uparrow$ & AR$\downarrow$ & AA$\uparrow$ & AR$\downarrow$ & AA$\uparrow$ & AR$\downarrow$ \\
         \specialrule{0.8pt}{0pt}{0pt}
         FedAVG      & 5.99  & 45.55 & 4.36 & 44.10 & 4.18 & 32.27 & 4.39 & 30.80 & 5.34 & 48.91 & 4.39 & 46.88 & 3.58 & 36.00 & 4.31 & 33.01 \\
         \hline
         TFCL        & 3.58  & 47.95 & 3.12 & 45.35 & 1.63 & 34.83 & 1.36 & 33.83 & 3.51 & 50.74 & 3.18 & 48.08 & 1.78 & 37.80 & 1.53 & 35.79 \\
         
         DCFCL       & 4.10  & 47.44 & 3.59 & 44.88 & 1.85 & 34.61 & 1.54 & 33.65 & 4.01 & 50.24 & 3.67 & 47.59 & 2.03 & 37.55 & 1.74 & 35.58 \\
         \hline
         Re-Fed      & 8.61  & 42.92 & 7.29 & 41.08 & 8.61 & 27.84 & 10.12 & 25.07 & 6.74 & 47.50 & 5.59 & 45.67 & 7.28 & 32.30 & 9.65 & 27.67 \\
         
         OFCL        & 16.91 & 34.63 & 16.41 & 32.05 & 12.44 & 24.01 & 13.40 & 21.79 & 15.98 & 38.26 & 15.90 & 35.37 & 16.54 & 23.04 & 16.59 & 20.73 \\
         
         FedCBDR     & 21.91 & 29.62 & 20.73 & 27.73 & 18.44 & 18.01 & 19.09 & 16.10 & 19.01 & 35.23 & 18.90 & 32.37 & 18.53 & 21.05 & 18.25 & 19.07 \\
         \hline
         FedKACE     & \fontseries{b}\selectfont 26.59 & \fontseries{b}\selectfont 24.95 & \fontseries{b}\selectfont 26.74 & \fontseries{b}\selectfont 21.73 & \fontseries{b}\selectfont 22.03 & \fontseries{b}\selectfont 14.43 & \fontseries{b}\selectfont 20.96 & \fontseries{b}\selectfont 14.23 & \fontseries{b}\selectfont 22.82 & \fontseries{b}\selectfont 31.43 & \fontseries{b}\selectfont 24.39 & \fontseries{b}\selectfont 26.87 & \fontseries{b}\selectfont 22.28 & \fontseries{b}\selectfont 17.30 & \fontseries{b}\selectfont 21.65 & \fontseries{b}\selectfont 15.67 \\
         \hline
         Centralized & 51.53 & 0.00  & 48.47 & 0.00  & 36.45 & 0.00  & 35.19 & 0.00  & 54.25 & 0.00  & 51.27 & 0.00  & 39.58 & 0.00  & 37.32 & 0.00  \\
         \specialrule{1.2pt}{0pt}{0pt}
         \end{tabular}%
      }%
   }%
\end{table}

\begin{figure*}[t]
    \vskip -0.15in
    \centering
    \begin{subfigure}[b]{0.24\linewidth}
        \centering
        \caption*{\scriptsize\textbf{CIFAR100 \& O=5}}
        \includegraphics[width=\linewidth]{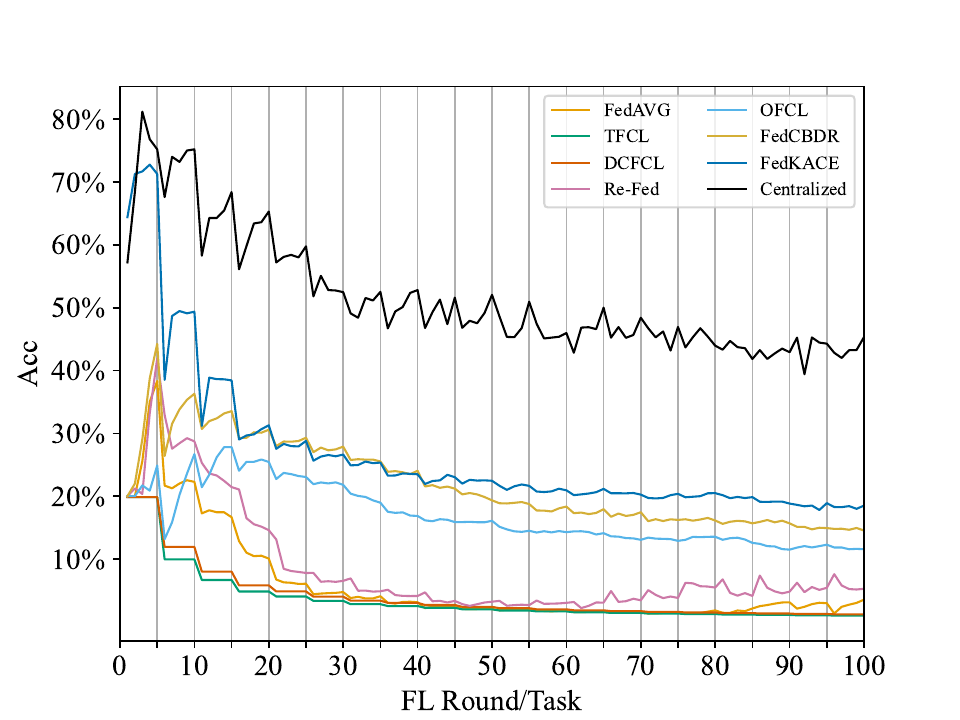}
    \end{subfigure}
    \begin{subfigure}[b]{0.24\linewidth}
        \centering
        \caption*{\scriptsize\textbf{CIFAR100 \& O=4}}
        \includegraphics[width=\linewidth]{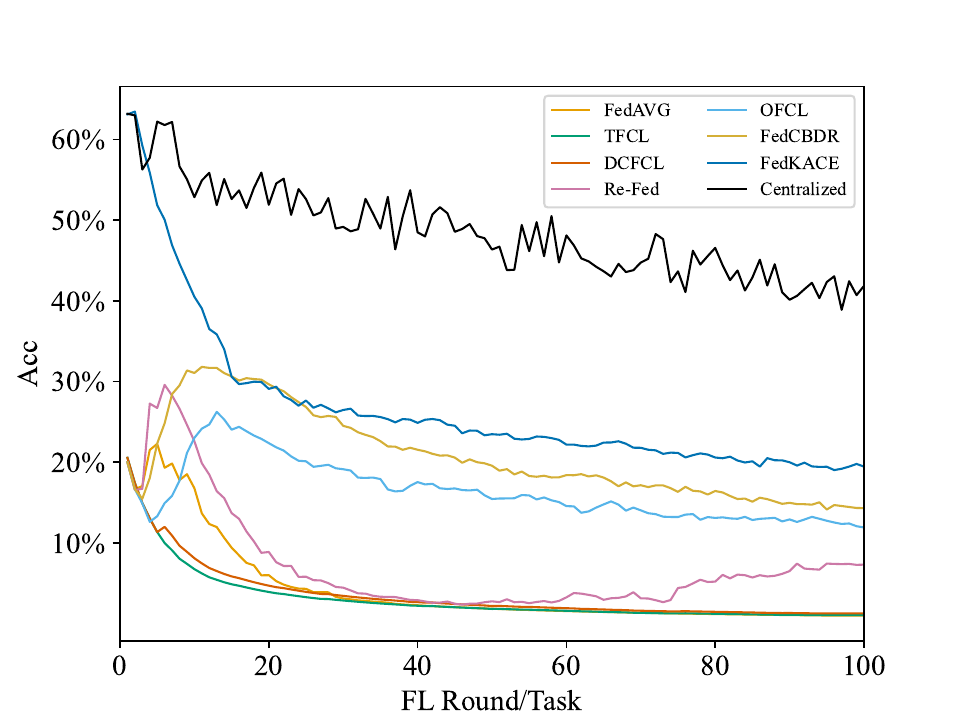}
    \end{subfigure}
    \begin{subfigure}[b]{0.24\linewidth}
        \centering
        \caption*{\scriptsize\textbf{CIFAR100 \& O=2}}
        \includegraphics[width=\linewidth]{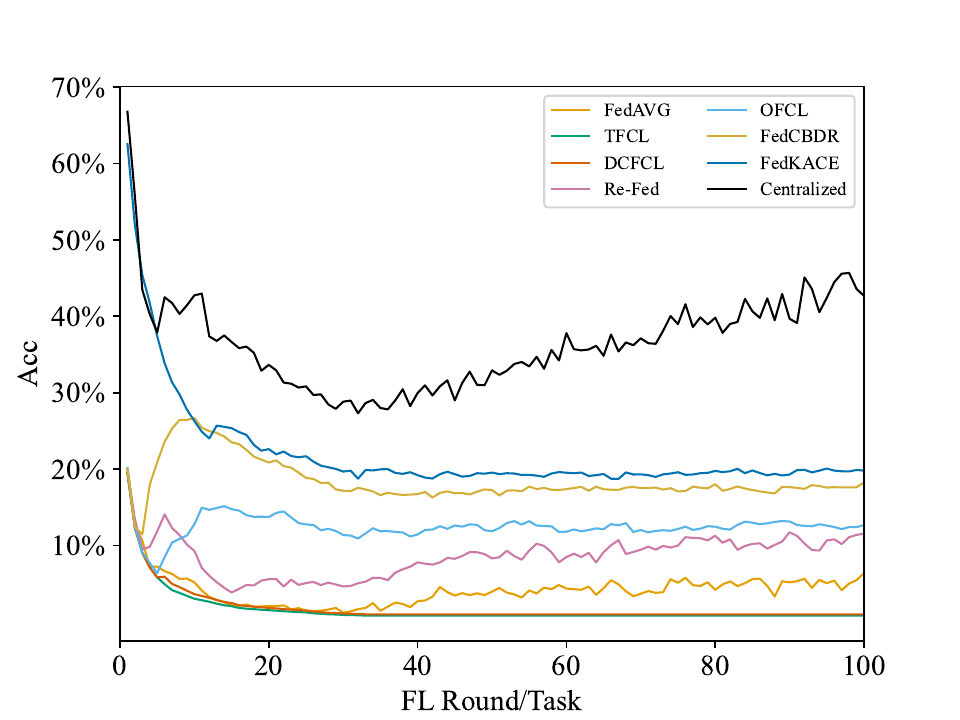}
    \end{subfigure}
    \begin{subfigure}[b]{0.24\linewidth}
        \centering
        \caption*{\scriptsize\textbf{CIFAR100 \& O=0}}
        \includegraphics[width=\linewidth]{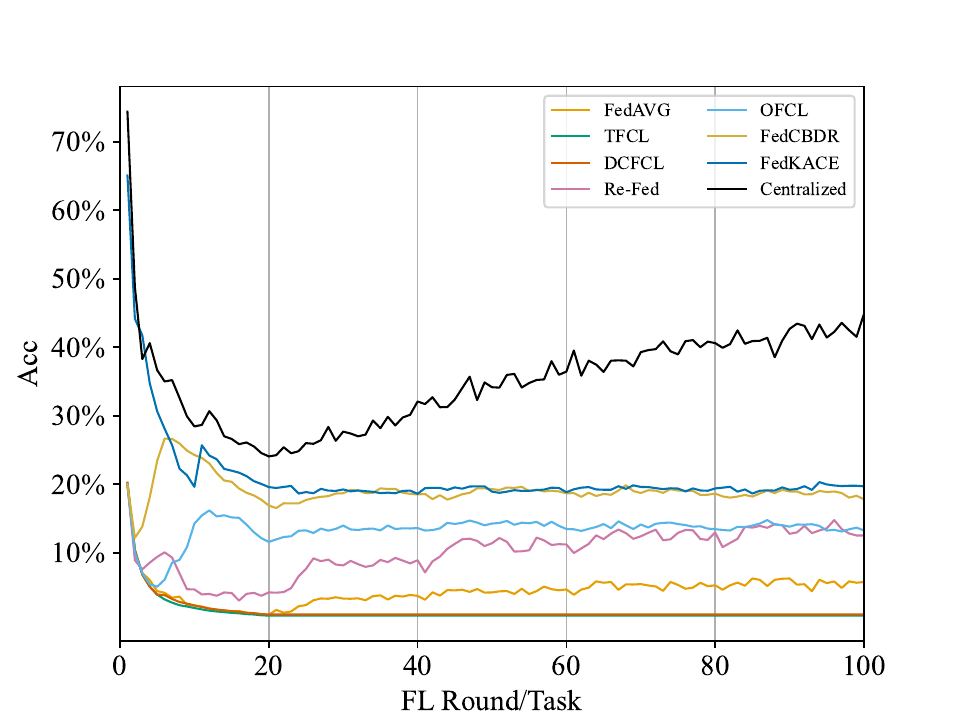}
    \end{subfigure}

    \vskip -0.05in
    \begin{subfigure}[b]{0.24\linewidth}
        \centering
        \includegraphics[width=\linewidth]{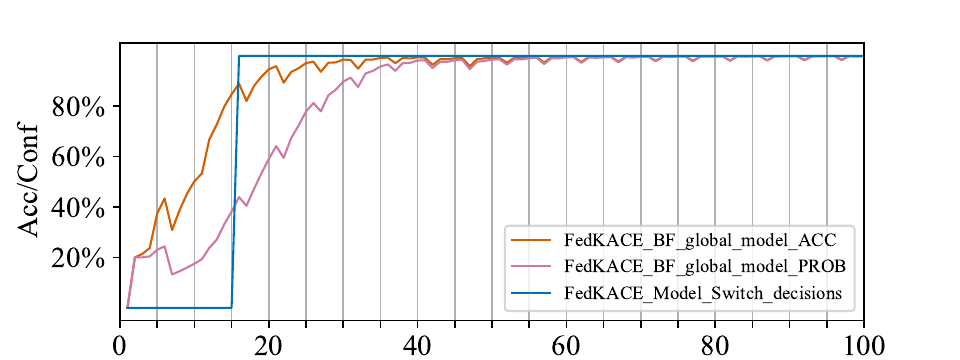}
    \end{subfigure}
    \begin{subfigure}[b]{0.24\linewidth}
        \centering
        \includegraphics[width=\linewidth]{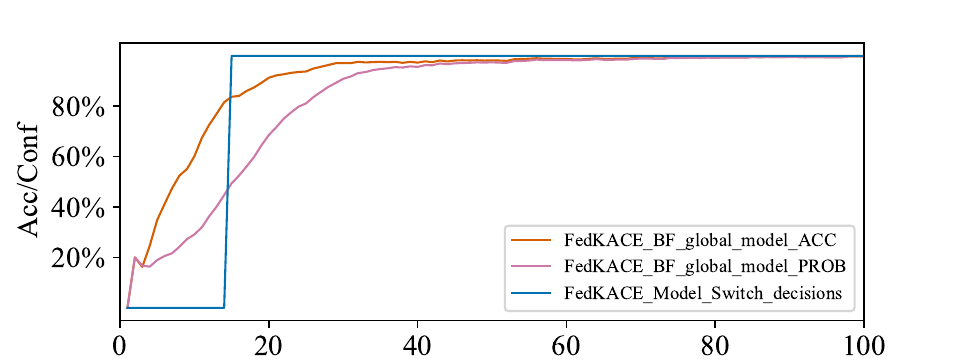}
    \end{subfigure}
    \begin{subfigure}[b]{0.24\linewidth}
        \centering
        \includegraphics[width=\linewidth]{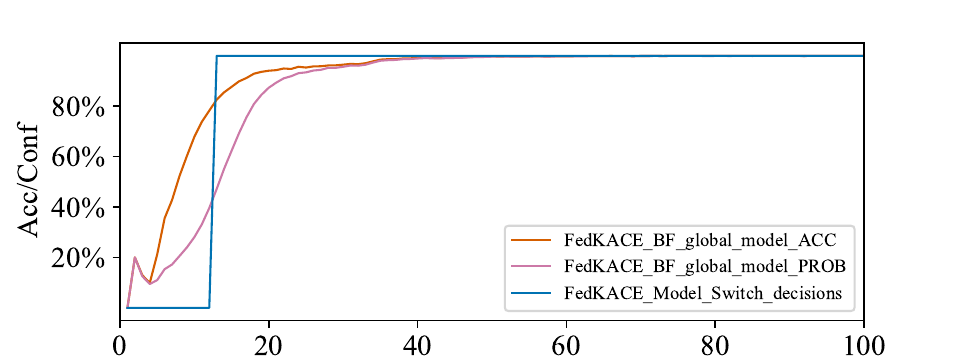}
    \end{subfigure}
    \begin{subfigure}[b]{0.24\linewidth}
        \centering
        \includegraphics[width=\linewidth]{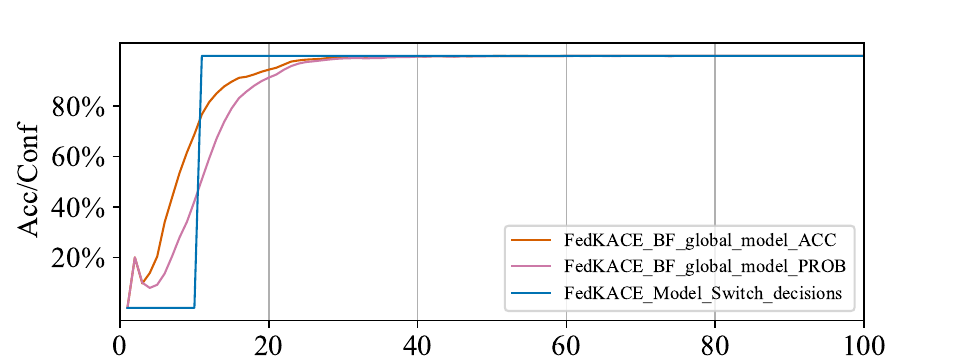}
    \end{subfigure}

    \hfill

    \vskip -0.15in
    \begin{subfigure}[b]{0.24\linewidth}
        \centering
        \caption*{\scriptsize\textbf{ImageNet100 \& O=5}}
        \includegraphics[width=\linewidth]{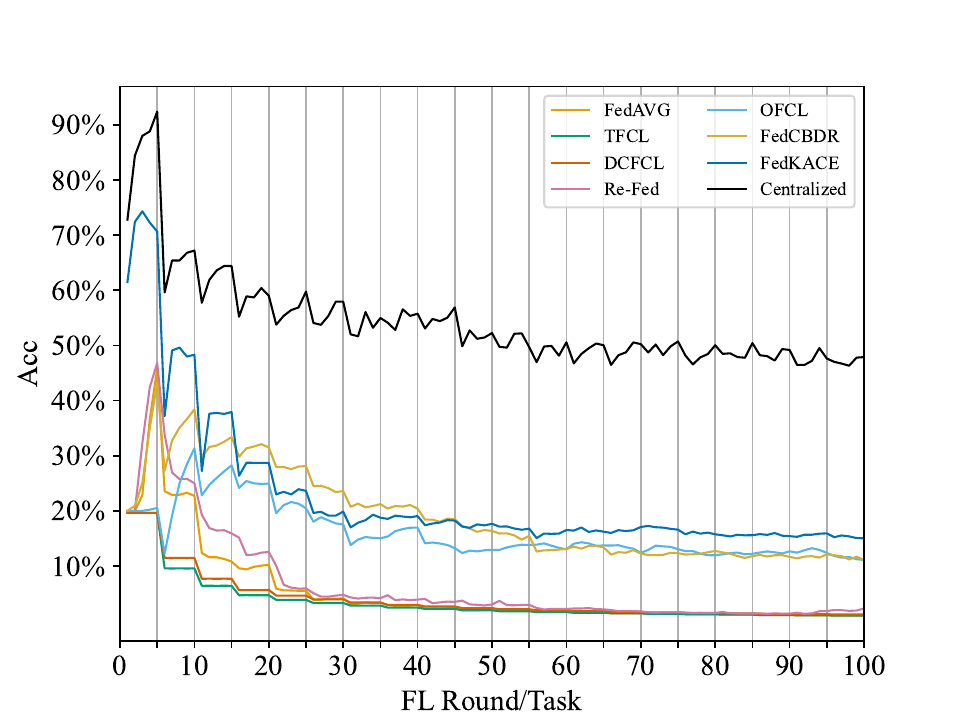}
    \end{subfigure}
    \begin{subfigure}[b]{0.24\linewidth}
        \centering
        \caption*{\scriptsize\textbf{ImageNet100 \& O=4}}
        \includegraphics[width=\linewidth]{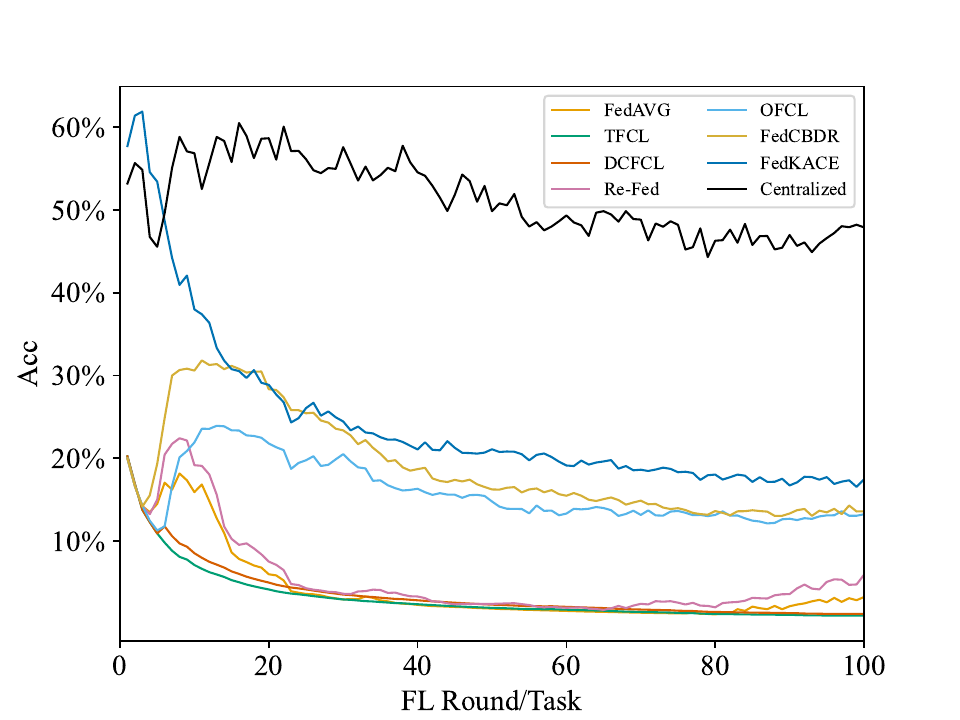}
    \end{subfigure}
    \begin{subfigure}[b]{0.24\linewidth}
        \centering
        \caption*{\scriptsize\textbf{ImageNet100 \& O=2}}
        \includegraphics[width=\linewidth]{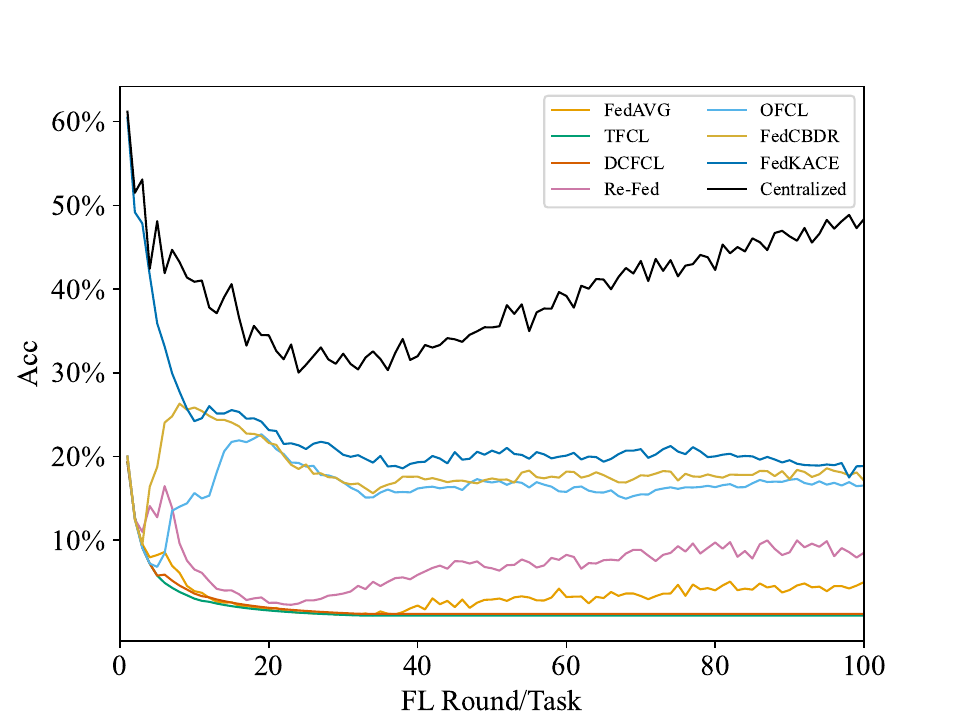}
    \end{subfigure}
    \begin{subfigure}[b]{0.24\linewidth}
        \centering
        \caption*{\scriptsize\textbf{ImageNet100 \& O=0}}
        \includegraphics[width=\linewidth]{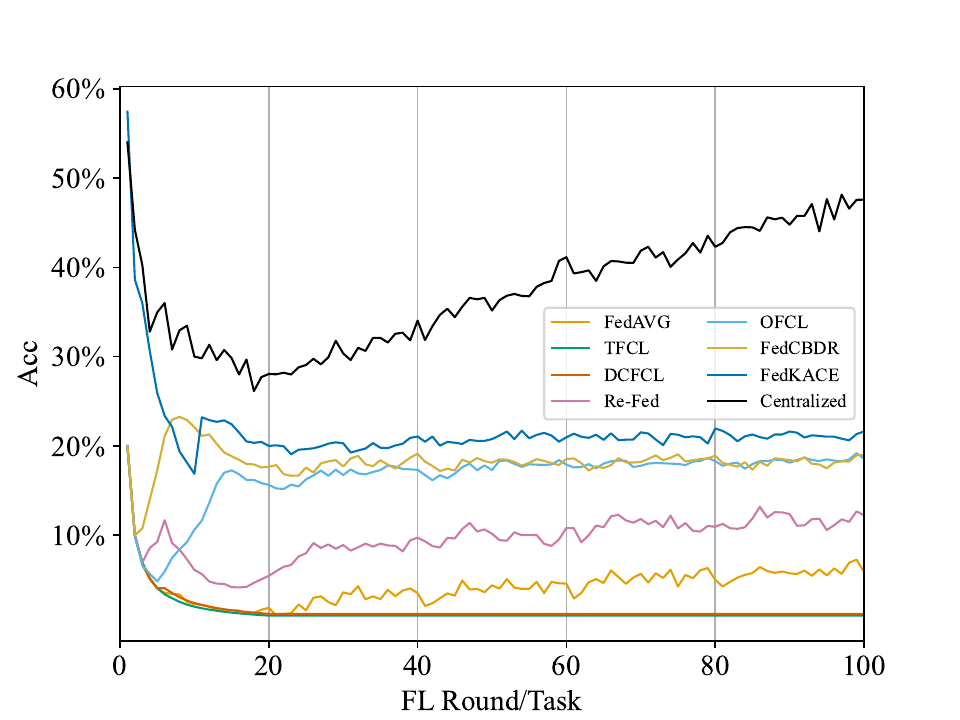}
    \end{subfigure}

    \vskip -0.05in
    \begin{subfigure}[b]{0.24\linewidth}
        \centering
        \includegraphics[width=\linewidth]{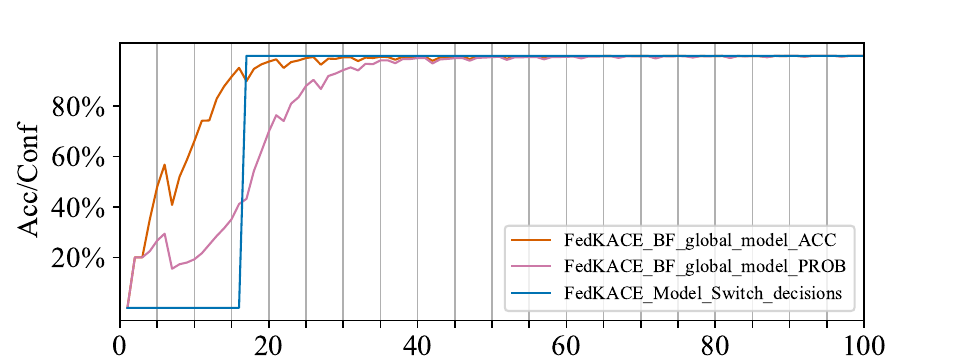}
    \end{subfigure}
    \begin{subfigure}[b]{0.24\linewidth}
        \centering
        \includegraphics[width=\linewidth]{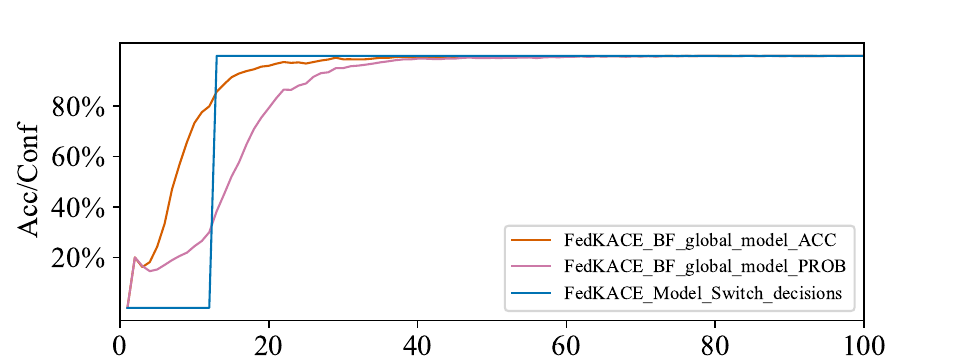}
    \end{subfigure}
    \begin{subfigure}[b]{0.24\linewidth}
        \centering
        \includegraphics[width=\linewidth]{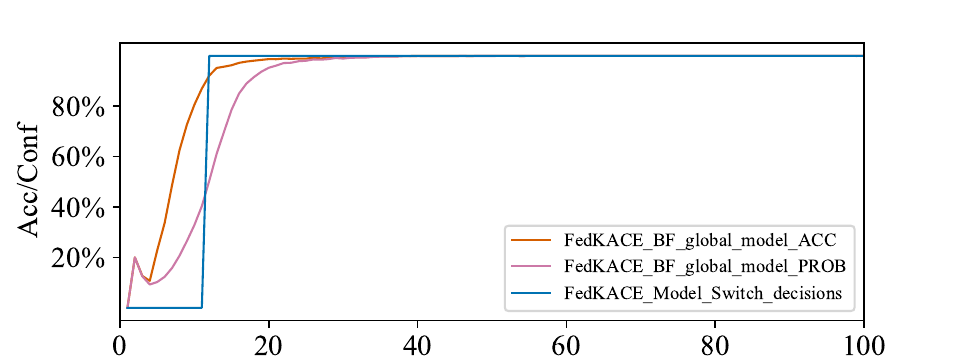}
    \end{subfigure}
    \begin{subfigure}[b]{0.24\linewidth}
        \centering
        \includegraphics[width=\linewidth]{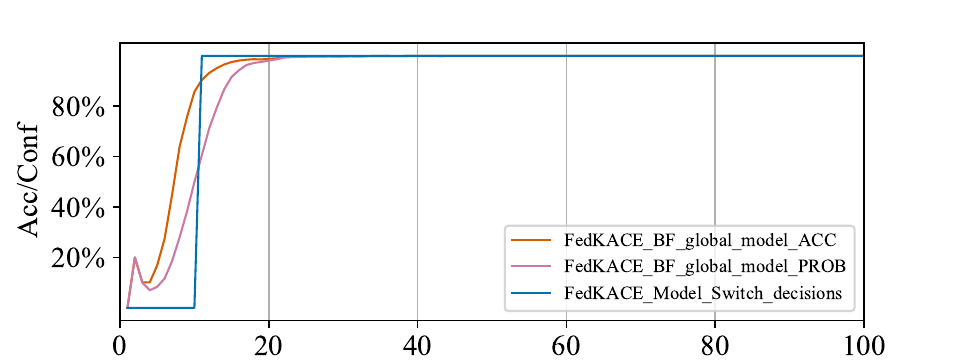}
    \end{subfigure}

    \label{Figure 2}
    \caption{Trends in Method Accuracy and FedKACE Buffer Metrics on Baseline Experiments.}
\end{figure*}

Under the setup above, \hyperref[Table 1]{Table 1} shows results for baseline methods and FedKACE, while \hyperref[Figure 2]{Figure 2} illustrates accuracy trends of all methods across all $T=100$ FL rounds.
All baseline methods exhibited performance limitations in streaming FCL setting:
FedAVG lacks mechanisms to handle dynamic class evolution, consequently failing to retain historical knowledge; 
TFCL's parameter isolation mechanism fails to integrate inter-class knowledge across FL rounds, fragmenting model knowledge;
DCFCL's coalitional affinity game fails to reach stable equilibria due to incomplete class coverage, undermining knowledge retention;
Re-Fed's static sample importance estimation mechanism largely fails to adapt to continually shifting distributions, reducing historical sample representativeness;
OFCL's Bregman Information-based uncertainty estimation partially fails under persistent distribution drift, compromising the selection of per-class representative samples;
FedCBDR's feature encryption-reconstruction mechanism via matrix decomposition slightly fails to maintain unbiased representations, hindering intra-class feature discrimination.
In contrast, FedKACE eliminates dependency on explicit task boundaries through its three synergistic mechanisms, thereby achieving the highest average accuracy and the lowest average regret.

\subsection{Ablation Studies}

\setcounter{table}{1}   
\begin{wraptable}{r}{0.5\textwidth}
  \centering 
  \vskip -0.2in 
  \caption{Experiment Results of Ablation Study}
  \vskip -0.1in 
  \label{Table 2}
  \resizebox{\linewidth}{!}{%
    {\footnotesize 
        \setlength{\tabcolsep}{3pt}
        \begin{tabular}{l||c@{\hspace{3pt}}c|c@{\hspace{3pt}}c||c@{\hspace{3pt}}c|c@{\hspace{3pt}}c}
          \specialrule{1.2pt}{0pt}{0pt}
          Me & \multicolumn{4}{c||}{Cifar100} & \multicolumn{4}{c}{ImageNet100} \\
          \cline{2-9}
          th & \multicolumn{2}{c|}{O=5} & \multicolumn{2}{c||}{O=0} & \multicolumn{2}{c|}{O=5} & \multicolumn{2}{c}{O=0} \\
          \cline{2-9}
          od & AA$\uparrow$ & AR$\downarrow$ & AA$\uparrow$ & AR$\downarrow$ & AA$\uparrow$ & AR$\downarrow$ & AA$\uparrow$ & AR$\downarrow$ \\
          \specialrule{0.8pt}{0pt}{0pt}
          AS1 & 25.42 & 26.12 & \fontseries{b}\selectfont 21.03 & \fontseries{b}\selectfont 14.17 & 22.05 & 32.20 & \fontseries{b}\selectfont 21.72 & \fontseries{b}\selectfont 15.60 \\
          
          AS2 & 23.20 & 28.34 & 19.82 & 15.37 & 20.21 & 34.04 & 20.79 & 16.53 \\
          
          AS3 & 24.23 & 27.30 & 16.81 & 18.38 & 22.03 & 32.22 & 16.17 & 21.14 \\
          \hline
          AS4 & 25.83 & 25.70 & 20.32 & 14.87 & 21.40 & 32.85 & 20.24 & 17.08 \\

          AS5 & \fontseries{b}\selectfont 26.63 & \fontseries{b}\selectfont 24.90 & 21.00 & 14.19 & \fontseries{b}\selectfont 22.89 & 31.36 & \fontseries{b}\selectfont 21.67 & 15.65 \\
          \hline
          AS6 & 26.54 & 25.00 & 20.24 & 14.95 & 22.23 & 32.02 & 20.98 & 16.33 \\
          
          AS7 & 26.47 & 25.06 & 20.21 & 14.99 & 22.34 & 31.91 & 20.18 & 17.14 \\
          \hline
          AS8 & 24.72 & 26.81 & 17.75 & 17.44 & 19.17 & 35.08 & 15.52 & 21.80 \\
          
          LKC & 18.78 & 32.75 & 13.17 & 22.02 & 18.23 & 36.02 & 11.47 & 25.85 \\
          \hline
          FKC & 26.59 & 24.95 & 20.96 & 14.23 &  22.82 & 31.43 & 21.65 & 15.67 \\
          \hline
          CTR & 51.53 & 0.00 & 35.19 & 0.00 & 54.25 & 0.00 & 37.32 & 0.00 \\
          \specialrule{1.2pt}{0pt}{0pt}
      \end{tabular}%
    }%
  }%
 
  \vskip 0.2in 
  \caption{AS Variants' Average local Buffer Kernel Condition Numbers in Final 75 FL Rounds}
  \vskip -0.1in 
  \label{Table 3}
  \resizebox{0.85\linewidth}{!}{%
    {\footnotesize 
        \setlength{\tabcolsep}{4pt} 
        \begin{tabular}{l|c|c|c|c}
          \specialrule{1.2pt}{0pt}{0pt}
          & \multicolumn{2}{c|}{Cifar100} & \multicolumn{2}{c}{ImageNet100} \\
          \cline{2-5}
          \multirow{-2}{*}{Method} & O=5 & O=0 & O=5 & O=0 \\
          \specialrule{0.8pt}{0pt}{0pt}
          FedKACE\_AS5 & 1575.6 & 1055.4 & 1638.6 & 396.8 \\
          FedKACE\_AS6 & 1637.3 & 1126.8 & 1885.4 & 478.4 \\
          FedKACE & 1593.4 & 1085.5 & 1711.6 & 421.2 \\
          \specialrule{1.2pt}{0pt}{0pt}
        \end{tabular}%
    }%
  }%
  \vskip -0.1in 
\end{wraptable}
We conduct ablation studies under the $O=5$ and $O=0$ settings. 
For the Adaptive Inference Model Switching Mechanism, we evaluate: FedKACE\_AS1(AS1), which triggers switching when $\Delta(\mathrm{gap}_k^{t})<0$ is first satisfied; FedKACE\_AS2(AS2) and FedKACE\_AS3(AS3), which exclusively use the global and local models, respectively.
For the Responsive Gradient-Balanced Replay Scheme, we examine: FedKACE\_AS4(AS4), which fixes $\lambda_k^{t,j} = 1$; and FedKACE\_AS5(AS5), which uses the full model gradient to compute $\lambda_k^{t,j}$.
For the Holistic Buffer Maintenance Strategy, we analyze: FedKACE\_AS6(AS6), which removes CDV and uses only IDV-weighted sampling; and FedKACE\_AS7(AS7), which replaces HBMS with class-balanced random sampling.
For the complete FedKACE framework, we test: FedKACE\_AS8(AS8), which preserving adaptive inference model switching but fixing $\lambda_k^{t,j} = 1$ and using class-balanced random sampling; and LocalKACE(LKC), which incorporates all FedKACE components except federated aggregation, relying solely on local training.

\begin{figure*}[t]
    \vskip -0.1in
    \centering
    \begin{subfigure}[b]{0.24\linewidth}
        \centering
        \caption*{\scriptsize\textbf{CIFAR100 \& O=5}}
        \includegraphics[width=\linewidth]{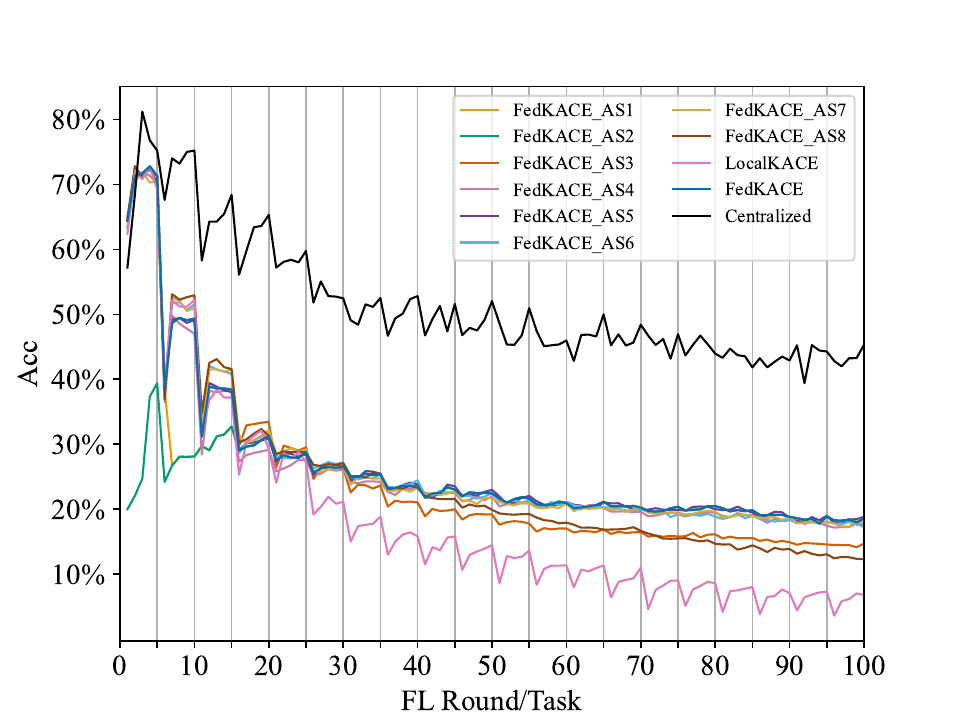}
    \end{subfigure}
    \begin{subfigure}[b]{0.24\linewidth}
        \centering
        \caption*{\scriptsize\textbf{CIFAR100 \& O=0}}
        \includegraphics[width=\linewidth]{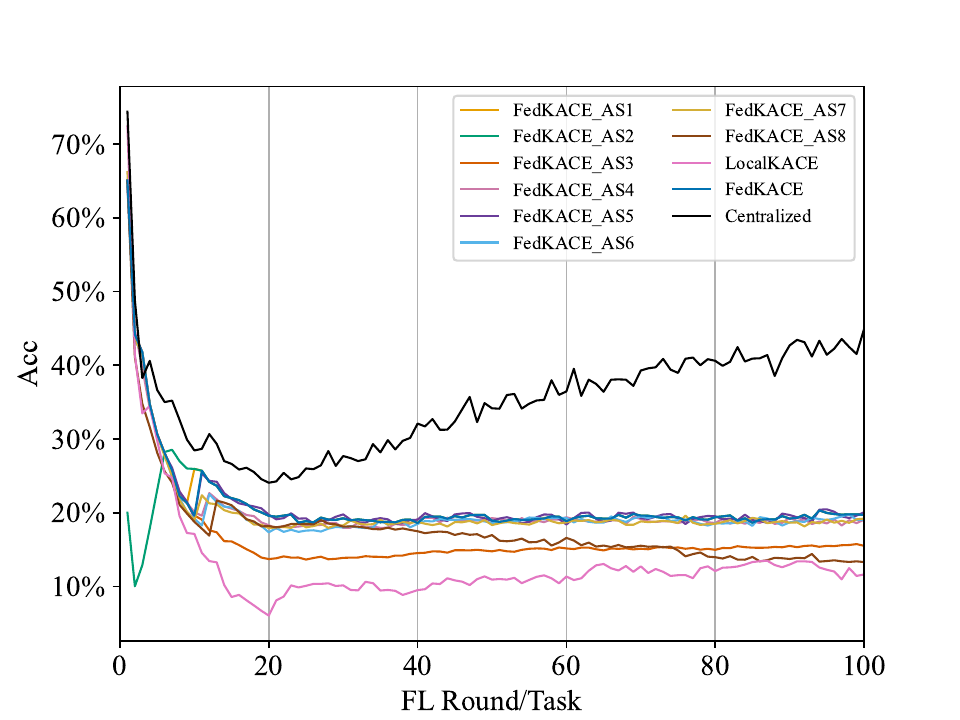}
    \end{subfigure}
    \begin{subfigure}[b]{0.24\linewidth}
        \centering
        \caption*{\scriptsize\textbf{ImageNet100 \& O=5}}
        \includegraphics[width=\linewidth]{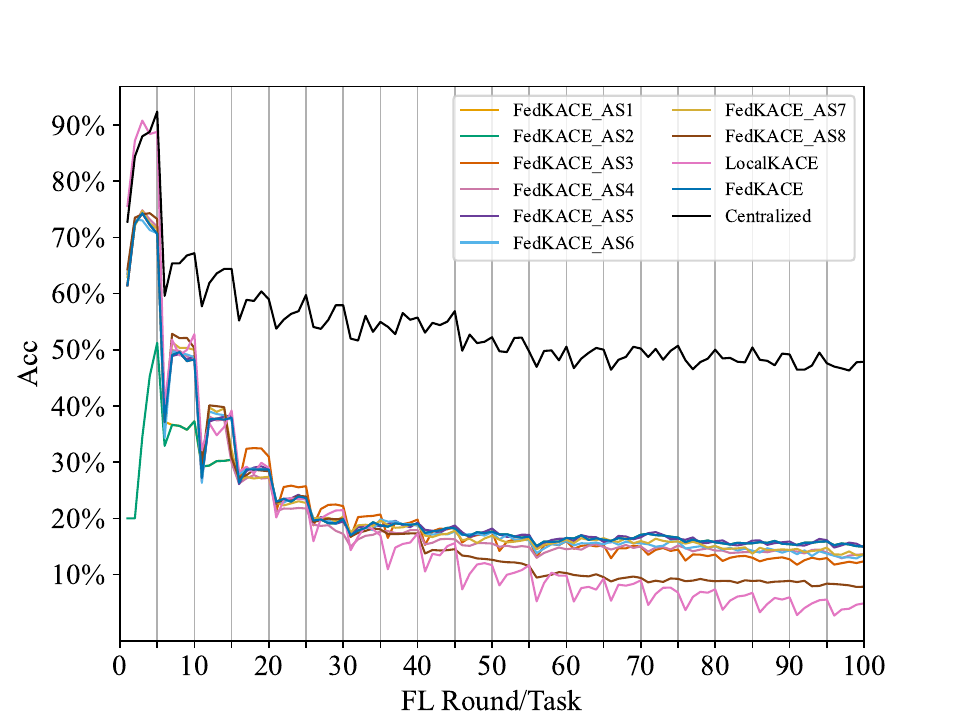}
    \end{subfigure}
    \begin{subfigure}[b]{0.24\linewidth}
        \centering
        \caption*{\scriptsize\textbf{ImageNet100 \& O=0}}
        \includegraphics[width=\linewidth]{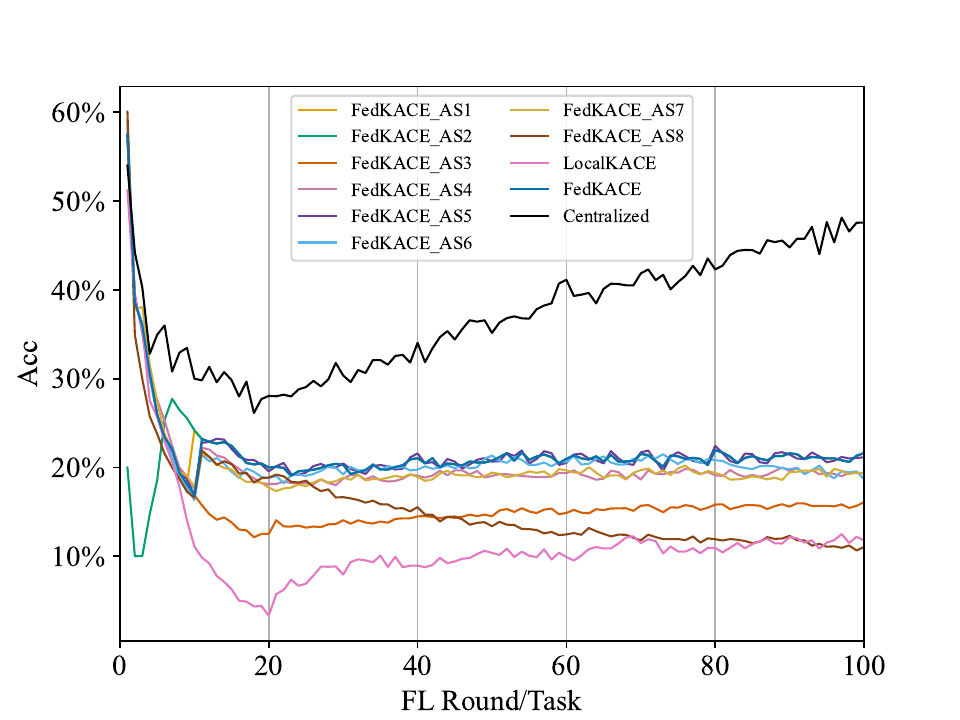}
    \end{subfigure}
    \label{Figure 3}
    \caption{Trends in Method Accuracy on Ablation Studies.}
    \vskip -0.1in
\end{figure*}

\hyperref[Table 2]{Table 2} shows ablation results (FedKACE as FKC, Centralized as CTR), while \hyperref[Figure 3]{Figure 3} illustrates accuracy trends of all variants, and \hyperref[Table 3]{Table 3} reports the average buffer kernel condition numbers for AS5, AS6, and FedKACE over the last 75 FL rounds. 
AS1 achieves marginal improvement in $O=0$ but suffers significant degradation in $O=5$ due to premature switching before model stabilization. 
AS2 and AS3 underperform by only using global or local models for inference, respectively, forfeiting adaptive model selection benefits.
AS4, which fixes $\lambda_k^{t,j} = 1$, cannot accommodate dynamically evolving knowledge and degrades performance.
AS5 computes $\lambda_k^{t,j}$ using full model gradients, yielding minimal gains while increasing gradient FLOPs per sample from 12.8K to 35.68M ($\sim 2780 \times$ on CIFAR) and 51.2K to 1823.57M ($\sim 35600 \times$ on ImageNet).
AS6 only uses IDV-weighted sampling, slightly reducing buffer kernel condition number but failing to identify decision boundary-critical samples, consequently compromising performance. 
AS7 uses class-balanced random sampling, yielding the highest kernel condition number and poor accuracy due to inefficient buffer management.
Finally, AS8 and LKC both underperform, demonstrating the necessity of integrating the components of FedKACE.

\setcounter{table}{3}   
\begin{table}[tp]
  \caption{Experimental Results of FedKACE with Varying Buffer Capacity $M$ and Client Count $K$}
  \label{Table 4}
  \resizebox{\textwidth}{!}{%
    {\footnotesize 
        \setlength{\tabcolsep}{4pt}    
        \begin{tabular}{l||c@{\hspace{4pt}}c|c@{\hspace{4pt}}c|c@{\hspace{4pt}}c|c@{\hspace{4pt}}c||c@{\hspace{4pt}}c|c@{\hspace{4pt}}c|c@{\hspace{4pt}}c|c@{\hspace{4pt}}c}
        \specialrule{1.2pt}{0pt}{0pt} 
        \multirow{3}{*}{Method} & \multicolumn{8}{c||}{CIFAR100} & \multicolumn{8}{c}{ImageNet100} \\
        \cline{2-17}
        & \multicolumn{2}{c|}{O=5} & \multicolumn{2}{c|}{O=4} & \multicolumn{2}{c|}{O=2} & \multicolumn{2}{c||}{O=0} & 
        \multicolumn{2}{c|}{O=5} & \multicolumn{2}{c|}{O=4} & \multicolumn{2}{c|}{O=2} & \multicolumn{2}{c}{O=0} \\
        \cline{2-17}
        & AA$\uparrow$ & AR$\downarrow$ & AA$\uparrow$ & AR$\downarrow$ & AA$\uparrow$ & AR$\downarrow$ & AA$\uparrow$ & AR$\downarrow$ & AA$\uparrow$ & AR$\downarrow$ & AA$\uparrow$ & AR$\downarrow$ & AA$\uparrow$ & AR$\downarrow$ & AA$\uparrow$ & AR$\downarrow$ \\
        \specialrule{0.8pt}{0pt}{0pt} 
        K10\_M500 & 24.60 & 26.94 & 25.19 & 23.27 & 19.53 & 16.93 & 18.99 & 16.20 & 21.21 & 33.03 & 23.26 & 28.00 & 21.67 & 17.91 & 20.11 & 17.21 \\
        \hline
        K10\_M1000 & 26.59 & 24.95 & 26.74 & 21.73 & 22.03 & 14.43 & 20.96 & 14.23 & 22.82 & 31.43 & 24.39 & 26.87 & 22.28 & 17.30 & 21.65 & 15.67 \\
        \hline
        K10\_M2000 & 27.72 & 23.82 & 28.21 & 20.26 & 23.15 & 13.31 & 23.82 & 11.37 & 23.98 & 30.27 & 25.43 & 25.84 & 24.46 & 15.12 & 24.19 & 11.13 \\
        \hline
        K20\_M1000 & 27.07 & 24.46 & 28.23 & 20.24 & 23.05 & 13.41 & 21.88 & 13.32 & 23.74 & 30.51 & 24.74 & 26.53 & 23.30 & 16.28 & 22.69 & 14.62 \\
        \hline
        Centralized & 51.53 & 0.00 & 48.47 & 0.00 & 36.45 & 0.00 & 35.19 & 0.00 & 54.25 & 0.00 & 51.27 & 0.00 & 39.58 & 0.00 & 37.32 & 0.00 \\
        \specialrule{1.2pt}{0pt}{0pt} 
    \end{tabular}
    }%
  }%
  \vskip -0.1in 
\end{table}

\subsection{Extend Research}
We analyze FedKACE under varying buffer capacities $M$ (default 1000) and client counts $K$ (default 10).
Experimental results in \hyperref[Table 4]{Table 4} confirm that while increasing either $M$ or $K$ improves performance, expanding $M$ yields greater gains than adding $K$ (e.g., $M=2000, K=10$ outperforms $M=1000, K=20$ for an identical $K\times M$ product), indicating more substantial gains from buffer expansion than client addition.

\section{Limitations}
With buffer size $M$ and $D$ new samples per round, HBMS incurs $O(M(M+D))$ complexity. While acceptable for small $M$, this becomes prohibitive as $M$ grows. For large $M$ where the need for explicit representativeness optimization diminishes, we suggest reducing the complexity to $O(r(M+D))$ by randomly sampling $r < M$ buffer landmarks, trading theoretical guarantees for efficiency.

\section{Conclusion}
In this paper, we extend FCL to streaming scenarios lacking task identifiers and featuring arbitrary class overlap across data chunks, posing three core challenges: knowledge confusion between old and new data knowledge, unknown and dynamically evolving data heterogeneity among clients, and the need for real-time inference on all encountered classes.  
To address these challenges, we introduce FedKACE with three components:
First, an adaptive mechanism monitors global model performance on the local buffer to determine when to switch the inference model from the local to the global one, improving client-side inference performance.
Second, a responsive gradient-balanced replay scheme utilizes the ratio of the squared L2 gradient norms, achieving client-specific knowledge balance between acquisition and retention.
Finally, a holistic buffer maintenance strategy employs two-stage screening process to jointly optimize buffer samples' informativeness, decision boundary significance, and feature space dispersion, enhancing knowledge retention under class overlap.
Experiments across diverse scenarios and theoretical analysis demonstrate FedKACE's effectiveness.

\bibliographystyle{unsrt}  
\bibliography{main}

\newpage
\appendix
\section{Datasets, Baseline Methods and Experimental Setup}
\label{Appendix A}
\subsection{Details of Datasets}
\label{Appendix A.1}
We conduct experiments on public datasets. 
Given the finite dataset size, we allow identical samples to be accessed by different clients across different FL rounds. 
Specifically, across all $T$ rounds, each sample is accessed by at most $K$ distinct clients, limiting its total system entries to no more than $K$. 
In contrast, traditional FCL methods employ static data partitioning where a single client exclusively holds each sample, but since each task spans far more than $K$ FL rounds, that client repeatedly trains on the same sample, yielding a usage count far exceeding $K$. 
Thus, although our setting relaxes data exclusivity, it still imposes a stricter constraint than the traditional FCL setting by strictly limiting each sample's overall reuse frequency.

\textbf{CIFAR100}: The CIFAR100 dataset \cite{19} is a real-world image classification dataset consisting of RGB images organized into 20 superclasses with 5 classes each, totaling 100 classes, where each class contains 500 training images and 100 test images.
In our experimental setup, for each class assigned to a client in an FL round, the client randomly selects 100 unaccessed training images from the 500 available samples per class, with no sample repetition across its FL rounds. 
After each FL round, each client evaluates its accuracy using the standard test set (all 100 images per class) from all its classes encountered thus far.

\textbf{ImageNet100}: The ImageNet100 dataset consists of the first 100 classes from the ILSVRC2012 dataset \cite{20}, with 1300 samples per class, where we designate the first 1000 samples as training images and the remaining 300 as test images. Prior to training, each client independently and randomly selects a fixed set of 100 samples from the 300 test samples per class to construct its unique standard test set. In our experimental setup, for each class assigned to a client in an FL round, the client randomly selects 100 unaccessed training images from the 1000 available samples per class, with no sample repetition across its FL rounds. After each FL round, each client evaluates its accuracy using the standard test set (all 100 images per class) from all classes encountered thus far.

\subsection{Details of Baseline Methods}
\label{Appendix A.2}
We select various FL and FCL methods for comparison, where the FL round index is provided as an explicit task identifier for each FCL method to adapt them to our setting.

\textbf{FedAVG}: FedAVG \cite{21} is a representative federated learning method. In each FL round, the server first broadcasts the current global model to all clients; each client then performs local training initialized from this model and uploads its updated parameters to the server. Subsequently, the server aggregates all local parameters through weighted averaging to obtain the updated global model, where each weight equals the ratio of the client's local dataset size to the total dataset size across all clients.

\textbf{TFCL}: TFCL \cite{22} is a parameter-isolation-based federated continual learning approach that addresses the limitations of conventional FCL methods that fail to account for task recurrence. At the onset of each task, TFCL decomposes the entire model into tagged sub-models through its Traceable Task Learning (TTL) module, thereby enabling the system to precisely locate and refine corresponding features from previous tasks when recurring tasks emerge; additionally, its Group-wise Knowledge Aggregation (GKA) module selectively federates recurring tasks by grouping similar tasks and aggregating their knowledge on the server side via client-server knowledge distillation across multiple FL rounds.

\textbf{DCFCL}: DCFCL \cite{23} is a decentralized federated continual learning method driven by dynamic cooperation that addresses catastrophic forgetting across temporal and spatial dimensions.
In each FL round, clients form non-overlapping coalitions through a coalitional affinity game that quantifies the benefits derived from cooperation by evaluating gradient coherence and model similarity; subsequently, a cooperative equilibrium, where no alternative coalition yields greater benefits for any participant, is achieved via a merge-blocking algorithm integrated with dynamic cooperative evolution, thereby enabling personalized models to effectively balance new knowledge acquisition and prior knowledge retention under heterogeneous data distributions.

\textbf{Re-Fed}: Re-Fed \cite{24} is a sample replay-based federated continual learning method. When a new task arrives, each client first updates a Personalized Informative Model (PIM) that incorporates both global and local information to compute sample importance. Clients quantify sample importance by calculating weighted accumulated gradient norms during PIM updates, and then cache the most valuable samples within their storage limits. Subsequently, clients train their local models using both the cached historical samples and the new task samples. During each FL round, the server aggregates the updated local models to form a new global model. When the next task arrives, clients again evaluate and store important samples from all previous tasks using the same PIM mechanism for replay in subsequent tasks.

\textbf{OFCL}: OFCL \cite{25} is a sample replay-based federated online continual learning method , where all clients process streaming data restricted to the same task-specific classes within each task. When a new task arrives, clients process streaming mini-batches of data and employ Bregman Information to estimate epistemic uncertainty at the sample level. Based on these estimates, clients selectively store representative samples with low epistemic uncertainty for each class into local buffers, and then train on both current and buffered historical data. Upon task transitions, clients refresh their buffers using the same uncertainty-based selection mechanism to retain knowledge for subsequent tasks.

\textbf{FedCBDR}: FedCBDR \cite{26} is a sample replay-based federated continual learning method. During each FL round of every task, clients employ a task-aware temperature scaling module to adaptively adjust the temperature of logits and reweight the learning objective at both the class and instance levels, thereby jointly training local models with historical samples and new task data; the server then aggregates these updated local models to construct a new global model. Upon task completion, clients obfuscate local features via ISVD and upload them to the server, which aggregates these matrices, performs SVD to reconstruct global representations of historical tasks while preserving privacy, and then conducts class-aware and importance-sensitive balanced sampling based on leverage scores to determine the indices of historical samples for retention, enabling clients to update their replay buffers accordingly.

\subsection{Details of Experimental Setup}
\label{Appendix A.3}
The following are the specific details of the experimental setup.

\textbf{Model Configuration}: 
To accommodate dataset-specific characteristics, randomly initialized ResNet-18 models \cite{46} are configured with an initial channel width of 16 for CIFAR-100 and 64 for ImageNet-100. All models employ an output layer with a fixed output dimension of 100, adapted to streaming classes via a dynamic masking mechanism: at FL round $t$, client $k$ activates only the output neurons corresponding to its accumulated class set $\mathcal{C}_{k}^{\leq t} = \bigcup_{\tau=1}^{t}\mathcal{C}_k^{\tau}$, while masking predictions and setting gradients to zero for all other neurons.

\textbf{Local Training Protocol}:  
For all methods, during the local training phase of each FL round, clients employ the AdamW optimizer (weight decay = 0.001) with a batch size of 32 and a fixed number of $J=20$ training epochs; the learning rate decays from 0.01 to 0 via a cosine annealing schedule where the decay period is synchronized with the local training epochs. For FedKACE, the adaptive inference model switching mechanism does not interfere with local training. Regarding hyperparameter configuration, all baseline FCL methods strictly adhere to the fixed hyperparameters specified in their original papers, whereas FedKACE adaptively adjusts its hyperparameters through data-driven mechanisms, thereby eliminating the need for manual tuning.

\newpage
\section{Theorem Proof}
This proof contains three theorems and three lemmas, where Lemma 1 supports the proof of Theorem 1, and Lemmas 2 and 3 support the proof of Theorem 3.

\subsection{Assumptions}
\label{Appendix B.1}

\textbf{Assumption 1} (Bounded Gradients of Local Loss Functions): For all clients $k = 1, \ldots, K$, all FL rounds $t = 1, \ldots, T $, and all $\theta \in \Omega_\theta$, the local loss functions $L_{k,\text{task}}^{t}(\theta)$ and $L_{k,\text{rep}}^{t}(\theta)$ are continuously differentiable, and there exist constants $L_1, L_2 > 0 $ such that the Euclidean norms of their gradients are bounded by:
\begin{equation}
  \|\nabla_{\theta} L_{k,\text{task}}^{t}(\theta)\|_2 \leq L_1, \quad \|\nabla_{\theta} L_{k,\text{rep}}^{t}(\theta)\|_2 \leq L_2
\end{equation}

\textbf{Assumption 2} (Compactness of Parameter Spaces): Both the model parameter space $\Omega_\theta$ and the replay weight space $\Omega_\lambda$ are compact sets.

\textbf{Assumption 3} (Robbins-Monro Condition for Local Learning Rates): The sequence of local learning rates $\{\eta_k^{t,j}\}$ satisfies $\sum_{j=1}^\infty \eta_k^{t,j} = \infty$ and $\sum_{j=1}^\infty (\eta_k^{t,j})^2 < \infty$. In practical implementations, a sufficiently large finite number of local iterations $J$ is used to approximate this asymptotic condition, ensuring approximate convergence of the local model.

\textbf{Assumption 4} (Compactness of the Range of the Normalized Logit Vector): The range of the normalized logit vector $\hat{g}(x)$ constitutes a compact set $\mathcal{G} \subseteq \mathbb{R}^d$, where the effective dimension $d = |\mathcal{C}_{k}^{\leq t}|$ is defined as the size of the class set.

\textbf{Assumption 5} (Lipschitz Continuity and Boundedness of the Loss Function): The loss function $\ell(f(x), y)$ is $L_{\mathrm{Lip}}$-Lipschitz continuous and is bounded within $[0,1]$.

\textbf{Assumption 6} (Uniformity of Samples per Class in the Buffer): In the $t$-th FL round, for any client $k$, the number of samples in the buffer for each class $c \in \mathcal{C}_{k}^{\leq t}$ is given by $|\mathcal{M}_{k,c}^{t}| = \Theta(M/|\mathcal{C}_{k}^{\leq t}|)$.

\textbf{Assumption 7} (Finiteness of the Number of Global Classes): The total number of global classes is finite, as formalized by $\left| \bigcup_{k=1}^K \bigcup_{\tau=1}^\infty \mathcal{C}_k^{\tau} \right| = C_{\max} < \infty$.

\textbf{Assumption 8} (Client Distribution Shift Constraint): For any client $k$, the data distributions $p_k^t$ and $p_k^{t-1}$ in consecutive FL rounds $t$ and $t-1$ satisfy $D_{\mathrm{TV}}(p_k^t , p_k^{t-1}) \leq C_{\max} / t^\alpha$, where $D_{\mathrm{TV}}$ is the total variation distance and $\alpha > 0$ denotes the decay rate of the distribution shift.

\textbf{Assumption 9} (Static Generalization Bound for Buffer Sampling): For any client $k \in \mathcal{K}$ and any FL round $t$, assume the buffer $\mathcal{M}_k^t$ contains $M$ samples. For any model $f \in \mathcal{F}$, the difference between the empirical risk on the buffer $\hat{\mathcal{R}}_{\mathcal{M}_k^t}(f)$ and the expected risk under the buffer distribution $\mathcal{R}_{p_{\mathcal{M}_k^t}}(f)$ satisfies the following explicit bound:
\begin{equation}
    \left| \hat{\mathcal{R}}_{\mathcal{M}_k^t}(f) - \mathcal{R}_{p_{\mathcal{M}_k^t}}(f) \right| \leq \frac{C_{\text{rep}}}{\sqrt{M}}
\end{equation}
where the constant $C_{\text{rep}} = \sqrt{2}\left(4 L_{\mathrm{Lip}} B + \sqrt{8\log(2/\delta)}\right)$, $L_{\mathrm{Lip}}$ is the Lipschitz constant of the loss function, $B$ is the RKHS norm bound, and $\delta$ is the confidence parameter. This bound solely captures the statistical fluctuations induced by finite sampling.

\textbf{Assumption 10} (L-smoothness of the local loss function): Let $L_{k, \text{total}}^{t}$ abstractly denote the local loss function for any client $k \in \mathcal{K}$ at FL round $t$, for any model parameters $\theta_a$ and $\theta_b$, the following inequality holds:
\begin{equation}  
  L_{k, \text{total}}^{t}(\theta_a) \leq L_{k, \text{total}}^{t}(\theta_b) + \langle \nabla_\theta L_{k, \text{total}}^{t}(\theta_b), \theta_a - \theta_b \rangle + \frac{L}{2}\|\theta_a - \theta_b\|_2^2  
\end{equation}

\textbf{Assumption 11} ($\mu$-strong convexity of the local loss function): Let $L_{k, \text{total}}^{t}$ abstractly denote the local loss function for any client $k \in \mathcal{K}$ at FL round $t$, for any model parameters $\theta_a$ and $\theta_b$, the following inequality holds:
\begin{equation}
  L_{k, \text{total}}^{t}(\theta_a) \geq L_{k, \text{total}}^{t}(\theta_b) + \langle \nabla_\theta L_{k, \text{total}}^{t}(\theta_b), \theta_a - \theta_b \rangle + \frac{\mu}{2}\|\theta_a - \theta_b\|_2^2
\end{equation}

\textbf{Assumption 12} (Bounded Local Gradient Variance): Let $L_{k, \text{total}}^{t}$ abstractly denote the local loss function for any client $k \in \mathcal{K}$ at FL round $t$, given that $\xi_k^{t,j}$ is uniformly sampled from the local data of client $k$ at epoch $j$, the variance of the stochastic gradient satisfies the following bound:  
\begin{equation}
  \mathbb{E}\left[\|\nabla_\theta L_{k, \text{total}}^{t}(\theta_k^{t,j}, \xi_k^{t,j}) - \nabla_\theta L_{k, \text{total}}^{t}(\theta_k^{t,j})\|_2^2\right] \leq \sigma_k^2
\end{equation}

\textbf{Assumption 13} (Bounded Local Gradient Norm): Let $L_{k, \text{total}}^{t}$ abstractly denote the local loss function for any client $k \in \mathcal{K}$ at FL round $t$, given that $\xi_k^{t,j}$ is uniformly sampled from the local data of client $k$ at epoch $j$, the expected squared $\ell_2$-norm of the stochastic gradient satisfies the following bound:  
\begin{equation}
  \mathbb{E}\left[\|\nabla_\theta L_{k, \text{total}}^{t}(\theta_k^{t,j}, \xi_k^{t,j})\|_2^2\right] \leq G^2
\end{equation}

\newpage
\subsection{Proof of Lemma 1}
\label{Appendix B.2}
\setcounter{theorem}{0}   
\begin{lemma}[Existence and Asymptotic Stability of the Ideal Saddle Point in Local Training]
  Under standard assumptions (see \hyperref[Appendix B.1]{Appendix B.1}), let $L_{k,\text{total}}^{t}(\theta_k^{t,j}, \lambda_k^{t,j}) = L_{k,\text{task}}^{t}(\theta_k^{t,j}) + \lambda_k^{t,j} L_{k,\text{rep}}^{t}(\theta_k^{t,j})$ be the total loss function of client $k$ at the $j$-th iteration step of the $t$-th FL round, where the replay loss weight $\lambda_k^{t,j+1}$ is updated via full-model gradients as $\lambda_k^{t,j+1} = \|\nabla_{\theta} L_{k,\text{rep}}^{t}(\theta_k^{t,j})\|_2^2 / \|\nabla_{\theta} L_{k,\text{task}}^{t}(\theta_k^{t,j})\|_2^2$.
  Under these conditions, there exists an ideal saddle point $(\theta_k^{t,*}, \lambda_k^{t,*}) \in \Omega_\theta \times \Omega_\lambda$ that exactly satisfies the minimax equilibrium condition:
\begin{equation}
    L_{k,\text{total}}^{t}(\theta_k^{t,*}, \lambda) \leq L_{k,\text{total}}^{t}(\theta_k^{t,*}, \lambda_k^{t,*}) \leq L_{k,\text{total}}^{t}(\theta, \lambda_k^{t,*}), \quad \forall \theta \in \Omega_\theta, \lambda \in \Omega_\lambda
\end{equation}
  Furthermore, as the number of local training steps $J$ approaches infinity, the sequence of iterates generated by the aforementioned update rule asymptotically converges to this saddle point within its neighborhood.
\end{lemma}

\textbf{Proof}:

We establish the existence of the ideal saddle point by leveraging the theoretical framework in \cite{02}. According to Assumption 11, $L_{k,\text{total}}^t(\theta, \cdot)$ is convex with respect to $\lambda$. Furthermore, based on the specific construction of $L_{k,\text{total}}^t$, its second-order partial derivative with respect to $\lambda$ is zero (i.e., $\partial^2_{\lambda\lambda}L_{k,\text{total}}^t(\theta, \lambda)=0$), which implies that $L_{k,\text{total}}^t(\theta, \cdot)$ is actually an affine function with respect to $\lambda$. Additionally, Assumption 2 ensures that the domain $\Omega_\theta \times \Omega_\lambda$ is a compact set. These properties satisfy the prerequisites of Theorem 1 in \cite{02}, guaranteeing the existence of a saddle point $(\Delta x^*,\theta^*)$ such that the value function $H(\cdot,\cdot)$ satisfies:
\begin{equation}
    H(\Delta x,\theta^*) \leq H(\Delta x^*,\theta^*) \leq H(\Delta x^*,\theta), \quad \forall \theta \in \Omega_\theta, \Delta x \in \Omega_\lambda
\end{equation}

In the context of this paper, by mapping the value function $H$ to $L_{k,\text{total}}^t$, the variable $\Delta x$ to $\lambda$, and keeping $\theta$ as $\theta$, we can conclude that there exists $(\theta_k^{t,*}, \lambda_k^{t,*}) \in \Omega_\theta \times \Omega_\lambda$ satisfying:
\begin{equation}
    L_{k,\text{total}}^t(\theta_k^{t,*}, \lambda) \leq L_{k,\text{total}}^t(\theta_k^{t,*}, \lambda_k^{t,*}) \leq L_{k,\text{total}}^t(\theta, \lambda_k^{t,*}), \quad \forall \theta \in \Omega_\theta, \lambda \in \Omega_\lambda
\end{equation}

Next, we prove the asymptotic stability of the saddle point $(\theta_k^{t,*}, \lambda_k^{t,*})$. By definition, the total loss function for local training is given by $L_{k,\text{total}}^{t}(\theta_k^{t,j}, \lambda_k^{t,j}) = L_{k,\text{task}}^{t}(\theta_k^{t,j}) + \lambda_k^{t,j} L_{k,\text{rep}}^{t}(\theta_k^{t,j})$. Utilizing the first-order necessary condition at the saddle point $(\theta_k^{t,*}, \lambda_k^{t,*})$, i.e., $\nabla_\theta L_{k,\text{total}}^t(\theta_k^{t,*}, \lambda_k^{t,*}) = \mathbf{0}$, we obtain:
\begin{equation}
    \nabla_\theta L_{k,\text{task}}^t(\theta_k^{t,*}) + \lambda_k^{t,*} \nabla_\theta L_{k,\text{rep}}^t(\theta_k^{t,*}) = \mathbf{0}
\end{equation}

Furthermore, since $\nabla_\theta L_{k,\text{total}}^t(\theta_k^{t,*}, \lambda_k^{t,*})=\mathbf{0}$, its inner product with any vector is zero, which implies that $\langle \nabla_\theta L_{k,\text{task}}^t(\theta_k^{t,*}), \nabla_\theta L_{k,\text{total}}^t(\theta_k^{t,*}, \lambda_k^{t,*}) \rangle = 0$ and $\langle \nabla_\theta L_{k,\text{rep}}^t(\theta_k^{t,*}), \nabla_\theta L_{k,\text{total}}^t(\theta_k^{t,*}, \lambda_k^{t,*}) \rangle = 0$. By substituting the expression of the total gradient and expanding the aforementioned inner products, we obtain:
\begin{equation}
    \begin{cases} 
        \|\nabla_\theta L_{k,\text{task}}^t(\theta_k^{t,*})\|_2^2 + \lambda_k^{t,*} \langle \nabla_\theta L_{k,\text{task}}^t(\theta_k^{t,*}), \nabla_\theta L_{k,\text{rep}}^t(\theta_k^{t,*}) \rangle = 0 
        \\ 
        \langle \nabla_\theta L_{k,\text{rep}}^t(\theta_k^{t,*}), \nabla_\theta L_{k,\text{task}}^t(\theta_k^{t,*}) \rangle + \lambda_k^{t,*} \|\nabla_\theta L_{k,\text{rep}}^t(\theta_k^{t,*})\|_2^2 = 0 
    \end{cases}
\end{equation}

By eliminating the cross inner product term $\langle \nabla_\theta L_{k,\text{rep}}^t(\theta_k^{t,*}), \nabla_\theta L_{k,\text{task}}^t(\theta_k^{t,*}) \rangle$, we obtain the gradient energy balance property at the saddle point:
\begin{equation}
    \|\nabla_\theta L_{k,\text{task}}^t(\theta_k^{t,*})\|_2^2 - (\lambda_k^{t,*})^2 \|\nabla_\theta L_{k,\text{rep}}^t(\theta_k^{t,*})\|_2^2 = 0
\end{equation}

Substituting the optimal replay loss weight at the saddle point, $\lambda_k^{t,*} = \|\nabla_\theta L_{k,\text{rep}}^t(\theta_k^{t,*})\|_2^2 / \|\nabla_\theta L_{k,\text{task}}^t(\theta_k^{t,*})\|_2^2$, into the above equation yields:
\begin{equation}
    \|\nabla_\theta L_{k,\text{task}}^t(\theta_k^{t,*})\|_2^6 = \|\nabla_\theta L_{k,\text{rep}}^t(\theta_k^{t,*})\|_2^6
\end{equation}

Combined with the non-negativity of the norm, the above equation implies that $\|\nabla_\theta L_{k,\text{task}}^t(\theta_k^{t,*})\|_2^2 = \|\nabla_\theta L_{k,\text{rep}}^t(\theta_k^{t,*})\|_2^2$, which yields a replay loss weight of $\lambda_k^{t,*} = 1$ at the saddle point. This indicates that in this dual-task setting, the saddle point is given by $(\theta_k^{t,*}, 1)$.

Next, considering that the replay loss weight $\lambda$ is a function of $\theta$, we define the quotient mapping $\mathcal{Q}(\theta) = \|\nabla_\theta L_{k,\text{task}}^t(\theta)\|_2^2 / \|\nabla_\theta L_{k,\text{rep}}^t(\theta)\|_2^2$. Since $\|\nabla_\theta L_{k,\text{task}}^t(\theta_k^{t,*})\|_2^2 > 0$ at the saddle point, we define $\mathcal{B}_\delta \triangleq \{ \theta : \|\theta - \theta_k^{t,*}\|_2 \leq \delta \}$ as the $\delta$-neighborhood of $\theta_k^{t,*}$. By the continuity of the gradients and the non-negativity of the norm, it follows that the $\|\nabla_\theta L_{k,\text{task}}^t(\theta)\|_2$ has a strictly positive lower bound within this neighborhood:
\begin{equation}
    \inf_{\theta \in \mathcal{B}_\delta} \|\nabla_\theta L_{k,\text{task}}^t(\theta)\|_2 \triangleq \epsilon_{\text{min,task}} \geq 0
\end{equation}

Given that $\nabla_\theta\|\nabla_\theta L_{k,\text{rep}}^t(\theta)\|_2^2 = 2 \nabla_\theta^2 L_{k,\text{rep}}^t(\theta) \nabla_\theta L_{k,\text{rep}}^t(\theta)$ and $\nabla_\theta\|\nabla_\theta L_{k,\text{task}}^t(\theta)\|_2^2 = 2 \nabla_\theta^2 L_{k,\text{task}}^t(\theta) \nabla_\theta L_{k,\text{task}}^t(\theta)$, applying the quotient rule to compute the gradient of the quotient mapping $\mathcal{Q}(\theta)$ yields:
\begin{equation}
    \nabla_\theta \mathcal{Q}(\theta) = 
    \frac{
        2\|\nabla_\theta L_{k,\text{task}}^t(\theta)\|_2^2 \nabla_\theta^2 L_{k,\text{rep}}^t(\theta) \nabla_\theta L_{k,\text{rep}}^t(\theta) 
        - 2\|\nabla_\theta L_{k,\text{rep}}^t(\theta)\|_2^2 \nabla_\theta^2 L_{k,\text{task}}^t(\theta) \nabla_\theta L_{k,\text{task}}^t(\theta)
    }
    {\|\nabla_\theta L_{k,\text{task}}^t(\theta)\|_2^4}
\end{equation}

Taking the $\ell_2$-norm of $\nabla_\theta \mathcal{Q}(\theta)$ and bounding the terms yields:
\begin{equation}
    \|\nabla_\theta \mathcal{Q}(\theta)\|_2 \leq 
    \frac{
        2\|\nabla_\theta L_{k,\text{task}}^t(\theta)\|_2^2 \|\nabla_\theta^2 L_{k,\text{rep}}^t(\theta)\|_2 \|\nabla_\theta L_{k,\text{rep}}^t(\theta) \|_2
        + 2\|\nabla_\theta L_{k,\text{rep}}^t(\theta)\|_2^2 \|\nabla_\theta^2 L_{k,\text{task}}^t(\theta)\|_2 \|\nabla_\theta L_{k,\text{task}}^t(\theta)\|_2
    }
    {\|\nabla_\theta L_{k,\text{task}}^t(\theta)\|_2^4}
\end{equation}

According to Assumption 1, we have $\|\nabla_\theta L_{k,\text{task}}^t(\theta)\|_2 \leq L_1$ and $\|\nabla_\theta L_{k,\text{rep}}^t(\theta)\|_2 \leq L_2$. Furthermore, under Assumption 10, it follows that $\|\nabla_\theta^2 L_{k,\text{task}}^t(\theta)\|_2 \leq L$ and $\|\nabla_\theta^2 L_{k,\text{rep}}^t(\theta)\|_2 \leq L$. Consequently, the norm of $\nabla_\theta \mathcal{Q}(\theta)$ is bounded. Denoting this upper bound by $L_{\mathcal{Q}}$, we obtain:
\begin{equation}
    \|\nabla_\theta \mathcal{Q}(\theta)\|_2 \leq \frac{2L L_1 L_2 (L_1 + L_2)}{(\epsilon_{\text{min,task}})^4} \triangleq L_{\mathcal{Q}}, 
    \quad \forall \theta \in \mathcal{B}_\delta
\end{equation}

That is, the quotient mapping $\mathcal{Q}(\theta)$ is $L_{\mathcal{Q}}$-Lipschitz continuous. Leveraging this quotient mapping, we define the composite objective function $\tilde{L}(\theta) = L_{k,\text{task}}^t(\theta) + \mathcal{Q}(\theta)L_{k,\text{rep}}^t(\theta)$. Computing its Hessian matrix $\nabla_\theta^2\tilde{L}(\theta)$ with respect to $\theta$ yields:
\begin{equation}
    \begin{split}
        \nabla_\theta^2\tilde{L}(\theta) 
        =& \nabla_\theta(\nabla_\theta L_{k,\text{task}}^t(\theta) + \nabla_\theta \mathcal{Q}(\theta) \cdot L_{k,\text{rep}}^t(\theta) + \mathcal{Q}(\theta) \cdot \nabla_\theta L_{k,\text{rep}}^t(\theta))
        \\
        =& \nabla_\theta^2 L_{k,\text{task}}^t(\theta) + \left[ \nabla_\theta \mathcal{Q}(\theta) \cdot (\nabla_\theta L_{k,\text{rep}}^t(\theta))^\top + L_{k,\text{rep}}^t(\theta) \cdot \nabla_\theta^2 \mathcal{Q}(\theta) \right] 
        \\
        &+ \left[ \nabla_\theta L_{k,\text{rep}}^t(\theta) \cdot (\nabla_\theta \mathcal{Q}(\theta))^\top + \mathcal{Q}(\theta) \cdot \nabla_\theta^2 L_{k,\text{rep}}^t(\theta) \right]
    \end{split}
\end{equation}

Since the replay loss weight at the saddle point is $\lambda_k^{t,*} = 1$, the total loss function evaluated at $(\theta, \lambda_k^{t,*})$ simplifies to $L_{k,\text{total}}^t(\theta, \lambda_k^{t,*}) = L_{k,\text{task}}^t(\theta) + \lambda_k^{t,*} L_{k,\text{rep}}^t(\theta) = L_{k,\text{task}}^t(\theta) + L_{k,\text{rep}}^t(\theta)$, and its Hessian matrix is given by $\nabla_\theta^2 L_{k,\text{total}}^t(\theta, \lambda_k^{t,*}) = \nabla_\theta^2 L_{k,\text{task}}^t(\theta) + \nabla_\theta^2 L_{k,\text{rep}}^t(\theta)$. Consequently, we have:
\begin{equation}
    \begin{split}
        \nabla_\theta^2\tilde{L}(\theta) - \nabla_\theta^2 L_{k,\text{total}}^t(\theta, \lambda_k^{t,*})
        =& \nabla_\theta \mathcal{Q}(\theta) \cdot (\nabla_\theta L_{k,\text{rep}}^t(\theta))^\top + L_{k,\text{rep}}^t(\theta) \cdot \nabla_\theta^2 \mathcal{Q}(\theta)
        \\
        &+ \nabla_\theta L_{k,\text{rep}}^t(\theta) \cdot (\nabla_\theta \mathcal{Q}(\theta))^\top + (\mathcal{Q}(\theta) - \lambda_k^{t,*}) \cdot \nabla_\theta^2 L_{k,\text{rep}}^t(\theta)
    \end{split}
\end{equation}

Taking the $\ell_2$-norm of the above expression. According to Assumption 1, we have $\|\nabla_\theta L_{k,\text{rep}}^t(\theta)\|_2 \leq L_2$. Furthermore, under Assumption 10, it follows that $\|\nabla_\theta^2 L_{k,\text{rep}}^t(\theta)\|_2 \leq L$. Consequently, we obtain:
\begin{equation}
    \|\nabla_\theta^2 \tilde{L}(\theta) - \nabla_\theta^2 L_{k,\text{total}}^t(\theta, \lambda_k^{t,*})\|_2 
    \leq \sup_{\theta \in \Omega_\theta}(|\mathcal{Q}(\theta)-\lambda_k^{t,*}|)\cdot L + 2L_{\mathcal{Q}} L_2 + 
    \sup_{\theta \in \Omega_\theta}(L_{k,\text{rep}}^{t}(\theta)) \cdot \|\nabla^2\mathcal{Q}(\theta)\|_2
\end{equation}

Since $\lambda_k^{t,*} = \mathcal{Q}(\theta_k^{t,*})$, for any $\theta \in \mathcal{B}_\delta(\theta_k^{t,*})$, we have:
\begin{equation}
    |\mathcal{Q}(\theta)-\lambda_k^{t,*}| = |\mathcal{Q}(\theta)-\mathcal{Q}(\theta_k^{t,*})| 
    = \|\mathcal{Q}(\theta)-\mathcal{Q}(\theta_k^{t,*})\|_2 \leq L_{\mathcal{Q}}\|\theta-\theta_k^{t,*}\|_2 \leq L_{\mathcal{Q}}\delta
\end{equation}

Consequently, we obtain $\sup_{\theta \in \mathcal{B}_\delta(\theta_k^{t,*})} |\mathcal{Q}(\theta)-\lambda_k^{t,*}| \leq L_{\mathcal{Q}}\delta$. Furthermore, under Assumption 5, it follows that $\sup_{\theta \in \Omega_\theta} L_{k,\text{rep}}^{t}(\theta) = 1$. Additionally, according to Assumption 1, the Hessian matrix mapping $\theta \mapsto \nabla_\theta^2 \mathcal{Q}(\theta)$ is continuous; combined with Assumption 2, the neighborhood $\mathcal{B}_\delta(\theta_k^{t,*}) \subset \Omega_\theta$ is a compact set. By applying the Weierstrass Extreme Value Theorem, there exists a finite supremum $H_{\mathcal{Q}} \triangleq \sup_{\theta \in \mathcal{B}_\delta} \|\nabla_\theta^2 \mathcal{Q}(\theta)\|_2 < \infty$. Thus, within the neighborhood of the saddle point, the spectral norm of the difference between the Hessian matrix of the composite objective $\tilde{L}(\theta)$ and the Hessian matrix at the saddle point is bounded:
\begin{equation}
    \|\nabla_\theta^2 \tilde{L}(\theta) - \nabla_\theta^2 L_{k,\text{total}}^t(\theta, \lambda_k^{t,*})\|_2 
    \leq \delta L_{\mathcal{Q}}L_{\mathrm{lip}} + 2L_{\mathcal{Q}} L_2 + H_{\mathcal{Q}} < \mu, 
    \forall \theta \in \mathcal{B}_\delta
\end{equation}

Next, we prove the strong convexity of the composite objective $\tilde{L}(\theta)$ within the neighborhood of the saddle point. By decomposing the Hessian as $\nabla_\theta^2 \tilde{L}(\theta) = \nabla_\theta^2 L_{k,\text{total}}^t(\theta, \lambda_k^{t,*}) + (\nabla_\theta^2 \tilde{L}(\theta) - \nabla_\theta^2 L_{k,\text{total}}^t(\theta, \lambda_k^{t,*}))$, we define the upper bound function for the spectral norm of the error term as $\rho(\delta) \triangleq \sup_{\theta \in \mathcal{B}_\delta} \|\nabla_\theta^2 \tilde{L}(\theta) - \nabla_\theta^2 L_{k,\text{total}}^t(\theta, \lambda_k^{t,*})\|_2$. Since $\lambda_k^{t,*} = \mathcal{Q}(\theta_k^{t,*}) = 1$ and the relevant mappings are continuous, it follows that as $\theta \to \theta_k^{t,*}$, the terms $\mathcal{Q}(\theta) - \lambda_k^{t,*}$, $\nabla_\theta \mathcal{Q}(\theta)$, and $\nabla_\theta^2 \mathcal{Q}(\theta)$ all approach zero. Consequently, $\lim_{\delta \to 0^+} \rho(\delta) = 0$. Furthermore, based on the continuity of $\rho(\delta)$ and its limit at zero, for the strong convexity parameter $\mu > 0$ given in Assumption 11, we define the neighborhood radius:
\begin{equation}
    \delta' = \rho^{-1}(\mu)
\end{equation}

such that $0 \leq \rho(\delta) < \mu, \quad \forall \delta \in (0, \delta')$. According to Assumption 11, we have $\nabla_\theta^2 L_{k,\text{total}}^t(\theta, \lambda_k^{t,*}) = \nabla_\theta^2 L_{k,\text{total}}^t(\theta, 1) \succeq \mu \mathbf{I}, \quad \forall \theta \in \mathcal{B}_{\delta'}$, which is equivalent to its minimum eigenvalue satisfying $\lambda_{\min}(\nabla_\theta^2 L_{k,\text{total}}^t(\theta, \lambda_k^{t,*})) \geq \mu, \quad \forall \theta \in \mathcal{B}_{\delta'}$. Finally, by Weyl's perturbation theorem, we obtain:
\begin{equation}
    \begin{split}
        \lambda_{\min}(\nabla_\theta^2 \tilde{L}(\theta)) 
        \geq& \lambda_{\min}(\nabla_\theta^2 L_{k,\text{total}}^t(\theta, \lambda_k^{t,*})) - \|\nabla_\theta^2 \tilde{L}(\theta) - \nabla_\theta^2 L_{k,\text{total}}^t(\theta, \lambda_k^{t,*})\|_2
        \\
        \geq& \mu - \rho(\delta) > \mu - (\delta L_{\mathcal{Q}}L_{\mathrm{lip}} + 2L_{\mathcal{Q}} L_2 + H_{\mathcal{Q}})
    \end{split}
\end{equation}

Consequently, we have:
\begin{equation}
    \nabla_\theta^2 \tilde{L}(\theta) \succeq (\mu - \rho(\delta))\mathbf{I} \succeq (\mu - (\delta L_{\mathcal{Q}}L_{\mathrm{lip}} + 2L_{\mathcal{Q}} L_2 + H_{\mathcal{Q}}))\mathbf{I} \succ \mathbf{0}, \quad \forall \theta \in \mathcal{B}_{\delta}
\end{equation}

Consequently, the composite objective $\tilde{L}(\theta)$ is $(\mu - (\delta L_{\mathcal{Q}}L_{\mathrm{lip}} + 2L_{\mathcal{Q}} L_2 + H_{\mathcal{Q}}))$-strongly convex within the saddle point neighborhood $\mathcal{B}_{\delta}$. For any $\theta \in \mathcal{B}_\delta$, consider the parameterized line segment $\phi(\tau) = \theta_k^{t,*} + \tau(\theta - \theta_k^{t,*})$ for $\tau \in [0,1]$. Since $\mathcal{B}_\delta$ is a convex set, it follows that $\phi(\tau) \in \mathcal{B}_\delta$. Therefore, the gradient difference $\nabla_\theta \tilde{L}(\theta) - \nabla_\theta \tilde{L}(\theta_k^{t,*})$ can be expressed as the integral of the Hessian matrix along this segment:
\begin{equation}
    \nabla_\theta \tilde{L}(\theta) - \nabla_\theta \tilde{L}(\theta_k^{t,*}) = \int_0^1 \frac{d}{d\tau} \nabla_\theta \tilde{L}(\phi(\tau)) \, d\tau = \int_0^1 \nabla_\theta^2 \tilde{L}(\phi(\tau)) (\theta - \theta_k^{t,*}) \, d\tau 
\end{equation}

Consequently, we have:
\begin{equation}
    \begin{split}
        \langle \nabla_\theta \tilde{L}(\theta) - \nabla_\theta \tilde{L}(\theta_k^{t,*}), \theta - \theta_k^{t,*} \rangle 
        &= \left\langle \int_0^1 \nabla_\theta^2 \tilde{L}(\phi(\tau)) (\theta - \theta_k^{t,*}) \, d\tau, \ \theta - \theta_k^{t,*} \right\rangle 
        \\
        &= \int_0^1 (\theta - \theta_k^{t,*})^\top \nabla_\theta^2 \tilde{L}(\phi(\tau)) (\theta - \theta_k^{t,*}) \, d\tau
        \\
        &\leq (\mu - (\delta L_{\mathcal{Q}}L_{\mathrm{lip}} + 2L_{\mathcal{Q}} L_2 + H_{\mathcal{Q}})) \int_0^1 (\theta - \theta_k^{t,*})^\top (\theta - \theta_k^{t,*}) \, d\tau
        \\
        &= (\mu - (\delta L_{\mathcal{Q}}L_{\mathrm{lip}} + 2L_{\mathcal{Q}} L_2 + H_{\mathcal{Q}})) \|\theta - \theta_k^{t,*}\|_2^2
    \end{split}
\end{equation}

Next, we prove the local asymptotic stability of the optimization dynamics at the saddle point. Consider the following manipulation of the gradient of the composite objective function $\nabla_\theta\tilde{L}(\theta)$:
\begin{equation}
    \begin{split}
        \nabla_\theta\tilde{L}(\theta) 
        =& \nabla_\theta L_{k,\text{task}}^t(\theta) + \nabla_\theta \mathcal{Q}(\theta) \cdot L_{k,\text{rep}}^t(\theta) + \mathcal{Q}(\theta) \cdot \nabla_\theta L_{k,\text{rep}}^t(\theta)
        \\
        =& [\nabla_\theta L_{k,\text{task}}^t(\theta) + \mathcal{Q}(\theta) \cdot \nabla_\theta L_{k,\text{rep}}^t(\theta)] + \nabla_\theta \mathcal{Q}(\theta) \cdot L_{k,\text{rep}}^t(\theta) 
        \\
        =& \left.\frac{\partial L_{k,\text{total}}^t(\theta, \lambda)}{\partial \theta} \right|_{\lambda=\mathcal{Q}(\theta)} + \nabla_\theta \mathcal{Q}(\theta) \cdot L_{k,\text{rep}}^t(\theta)
    \end{split}
\end{equation}

Consequently, we obtain the expression for the gradient of the total loss function with respect to $\theta$ evaluated at $\lambda = \mathcal{Q}(\theta)$:
\begin{equation}
    \nabla_\theta L_{k,\text{total}}^t(\theta, \mathcal{Q}(\theta)) \triangleq \left.\frac{\partial L_{k,\text{total}}^t(\theta, \lambda)}{\partial \theta} \right|_{\lambda=\mathcal{Q}(\theta)}
    = \nabla_\theta\tilde{L}(\theta) - \nabla_\theta \mathcal{Q}(\theta) \cdot L_{k,\text{rep}}^t(\theta)
\end{equation}

Considering the $j+1$-th parameter update $\theta_k^{t,j+1} = \theta_k^{t,j} - \eta_k^{t,j} \nabla_\theta L_{k, \text{total}}^t(\theta_k^{t,j}, \mathcal{Q}(\theta_k^{t,j}))$, we have:
\begin{equation}
    \begin{split}
        \|\theta_k^{t,j+1} - \theta_k^{t,*}\|_2^2 
        =& \|(\theta_k^{t,j} - \theta_k^{t,*}) - \nabla_\theta L_{k, \text{total}}^t(\theta_k^{t,j}, \mathcal{Q}(\theta_k^{t,j}))\|_2^2
        \\
        =& \|\theta_k^{t,j} - \theta_k^{t,*}\|_2^2 
        - 2\eta_k^{t,j}\langle \nabla_\theta L_{k, \text{total}}^t(\theta_k^{t,j}, \mathcal{Q}(\theta_k^{t,j})), \theta_k^{t,j} - \theta_k^{t,*} \rangle 
        + (\eta_k^{t,j})^2\|\nabla_\theta L_{k, \text{total}}^t(\theta_k^{t,j}, \mathcal{Q}(\theta_k^{t,j}))\|_2^2
    \end{split}
\end{equation}

For the inner product term $\langle \nabla_\theta L_{k, \text{total}}^t(\theta_k^{t,j}, \mathcal{Q}(\theta_k^{t,j})), \theta_k^{t,j} - \theta_k^{t,*} \rangle$, substituting $\nabla_\theta L_{k,\text{total}}^t(\theta_k^{t,j}, \mathcal{Q}(\theta_k^{t,j})) = \nabla_\theta\tilde{L}(\theta_k^{t,j}) - \nabla_\theta \mathcal{Q}(\theta_k^{t,j}) L_{k,\text{rep}}^t(\theta_k^{t,j})$ yields:
\begin{equation}
    \begin{split}
        \langle \nabla_\theta L_{k, \text{total}}^t(\theta_k^{t,j}, \mathcal{Q}(\theta_k^{t,j})), \theta_k^{t,j} - \theta_k^{t,*} \rangle
        =& \langle \nabla_\theta\tilde{L}(\theta_k^{t,j}), \theta_k^{t,j} - \theta_k^{t,*} \rangle - \langle \nabla_\theta \mathcal{Q}(\theta_k^{t,j}) \cdot L_{k,\text{rep}}^t(\theta_k^{t,j}), \theta_k^{t,j} - \theta_k^{t,*} \rangle
        \\
        =& \langle \nabla_\theta\tilde{L}(\theta_k^{t,j}) - \nabla_\theta\tilde{L}(\theta_k^{t,*}) , \theta_k^{t,j} - \theta_k^{t,*} \rangle + \langle\nabla_\theta\tilde{L}(\theta_k^{t,*}) , \theta_k^{t,j} - \theta_k^{t,*} \rangle
        \\
        &- \langle \nabla_\theta \mathcal{Q}(\theta_k^{t,j}) \cdot L_{k,\text{rep}}^t(\theta_k^{t,j}), \theta_k^{t,j} - \theta_k^{t,*} \rangle
    \end{split}
\end{equation}

Furthermore, utilizing the inequality $\langle \nabla_\theta \tilde{L}(\theta_k^{t,j}) - \nabla_\theta \tilde{L}(\theta_k^{t,*}), \theta_k^{t,j} - \theta_k^{t,*} \rangle \leq (\mu - (\delta L_{\mathcal{Q}}L_{\mathrm{lip}} + 2L_{\mathcal{Q}} L_2 + H_{\mathcal{Q}})) \|\theta_k^{t,j} - \theta_k^{t,*}\|_2^2$, we obtain:
\begin{equation}
    \begin{split}
        \langle \nabla_\theta L_{k, \text{total}}^t(\theta_k^{t,j}, \mathcal{Q}(\theta_k^{t,j})), \theta_k^{t,j} - \theta_k^{t,*} \rangle
        \leq& (\mu - (\delta L_{\mathcal{Q}}L_{\mathrm{lip}} + 2L_{\mathcal{Q}} L_2 + H_{\mathcal{Q}})) \|\theta_k^{t,j} - \theta_k^{t,*}\|_2^2 + \langle\nabla_\theta\tilde{L}(\theta_k^{t,*}) , \theta_k^{t,j} - \theta_k^{t,*} \rangle
        \\
        &- \langle \nabla_\theta \mathcal{Q}(\theta_k^{t,j}) \cdot L_{k,\text{rep}}^t(\theta_k^{t,j}), \theta_k^{t,j} - \theta_k^{t,*} \rangle
    \end{split}
\end{equation}

Consequently, we have:
\begin{equation}
    \begin{split}
        \|\theta_k^{t,j+1} - \theta_k^{t,*}\|_2^2 
        \leq& (1-2\eta_k^{t,j}(\mu - (\delta L_{\mathcal{Q}}L_{\mathrm{lip}} + 2L_{\mathcal{Q}} L_2 + H_{\mathcal{Q}})))\|\theta_k^{t,j} - \theta_k^{t,*}\|_2^2 
        - 2\eta_k^{t,j}\langle\nabla_\theta\tilde{L}(\theta_k^{t,*}) , \theta_k^{t,j} - \theta_k^{t,*} \rangle
        \\
        &+ 2\eta_k^{t,j}\langle \nabla_\theta \mathcal{Q}(\theta_k^{t,j}) \cdot L_{k,\text{rep}}^t(\theta_k^{t,j}), \theta_k^{t,j} - \theta_k^{t,*} \rangle
        + (\eta_k^{t,j})^2\|\nabla_\theta L_{k, \text{total}}^t(\theta_k^{t,j}, \mathcal{Q}(\theta_k^{t,j}))\|_2^2
        \\
        =& (1-2\eta_k^{t,j}(\mu - (\delta L_{\mathcal{Q}}L_{\mathrm{lip}} + 2L_{\mathcal{Q}} L_2 + H_{\mathcal{Q}})))\|\theta_k^{t,j} - \theta_k^{t,*}\|_2^2 + (\eta_k^{t,j})^2\|\nabla_\theta L_{k, \text{total}}^t(\theta_k^{t,j}, \mathcal{Q}(\theta_k^{t,j}))\|_2^2
        \\
        &+ 2\eta_k^{t,j}\langle \nabla_\theta \mathcal{Q}(\theta_k^{t,j}) \cdot L_{k,\text{rep}}^t(\theta_k^{t,j}) - \nabla_\theta\tilde{L}(\theta_k^{t,*}), \theta_k^{t,j} - \theta_k^{t,*} \rangle
    \end{split}
\end{equation}

For the last term $\langle \nabla_\theta \mathcal{Q}(\theta_k^{t,j}) L_{k,\text{rep}}^t(\theta_k^{t,j}) - \nabla_\theta\tilde{L}(\theta_k^{t,*}), \theta_k^{t,j} - \theta_k^{t,*} \rangle$, since the first-order necessary condition $\nabla_\theta L_{k, \text{total}}^t(\theta_k^{t,*}, \mathcal{Q}(\theta_k^{t,*})) = \mathbf{0}$ is satisfied at the saddle point, we have $\nabla_\theta\tilde{L}(\theta_k^{t,*}) = \nabla_\theta L_{k, \text{total}}^t(\theta_k^{t,*}, \mathcal{Q}(\theta_k^{t,*})) + \nabla_\theta \mathcal{Q}(\theta_k^{t,*}) L_{k,\text{rep}}^t(\theta_k^{t,*}) = \nabla_\theta \mathcal{Q}(\theta_k^{t,*}) L_{k,\text{rep}}^t(\theta_k^{t,*})$. Consequently, we obtain:
\begin{equation}
    \begin{split}
        & \langle \nabla_\theta \mathcal{Q}(\theta_k^{t,j}) \cdot L_{k,\text{rep}}^t(\theta_k^{t,j}) - \nabla_\theta\tilde{L}(\theta_k^{t,*}), \theta_k^{t,j} - \theta_k^{t,*} \rangle
        \\
        =& 
        \langle \nabla_\theta \mathcal{Q}(\theta_k^{t,j}) \cdot L_{k,\text{rep}}^t(\theta_k^{t,j}) - \nabla_\theta \mathcal{Q}(\theta_k^{t,*}) \cdot L_{k,\text{rep}}^t(\theta_k^{t,*}), \theta_k^{t,j} - \theta_k^{t,*} \rangle
        + \langle \nabla_\theta \mathcal{Q}(\theta_k^{t,*}) \cdot L_{k,\text{rep}}^t(\theta_k^{t,*}) - \nabla_\theta\tilde{L}(\theta_k^{t,*}), \theta_k^{t,j} - \theta_k^{t,*} \rangle
        \\
        =& \langle \nabla_\theta \mathcal{Q}(\theta_k^{t,j}) \cdot L_{k,\text{rep}}^t(\theta_k^{t,j}) - \nabla_\theta \mathcal{Q}(\theta_k^{t,*}) \cdot L_{k,\text{rep}}^t(\theta_k^{t,*}), \theta_k^{t,j} - \theta_k^{t,*} \rangle
        + \langle \mathbf{0}, \theta_k^{t,j} - \theta_k^{t,*} \rangle 
        \\
        =& \langle \nabla_\theta \mathcal{Q}(\theta_k^{t,j}) \cdot L_{k,\text{rep}}^t(\theta_k^{t,j}) - \nabla_\theta \mathcal{Q}(\theta_k^{t,*}) \cdot L_{k,\text{rep}}^t(\theta_k^{t,*}), \theta_k^{t,j} - \theta_k^{t,*} \rangle 
        \\
        =& \langle \nabla_\theta \mathcal{Q}(\theta_k^{t,j}) \cdot L_{k,\text{rep}}^t(\theta_k^{t,j}) - \nabla_\theta \mathcal{Q}(\theta_k^{t,*}) \cdot L_{k,\text{rep}}^t(\theta_k^{t,j}), \theta_k^{t,j} - \theta_k^{t,*} \rangle
        \\
        &+ \langle \nabla_\theta \mathcal{Q}(\theta_k^{t,*}) \cdot L_{k,\text{rep}}^t(\theta_k^{t,j}) - \nabla_\theta \mathcal{Q}(\theta_k^{t,*}) \cdot L_{k,\text{rep}}^t(\theta_k^{t,*}), \theta_k^{t,j} - \theta_k^{t,*} \rangle
        \\
        =& \langle (\nabla_\theta \mathcal{Q}(\theta_k^{t,j}) - \nabla_\theta \mathcal{Q}(\theta_k^{t,*})) \cdot L_{k,\text{rep}}^t(\theta_k^{t,j}), \theta_k^{t,j} - \theta_k^{t,*} \rangle
        + \langle \nabla_\theta \mathcal{Q}(\theta_k^{t,*}) \cdot (L_{k,\text{rep}}^t(\theta_k^{t,j}) - L_{k,\text{rep}}^t(\theta_k^{t,*})), \theta_k^{t,j} - \theta_k^{t,*} \rangle
    \end{split}
\end{equation}

Furthermore, applying the norm inequality to the above inner product and utilizing $\|\nabla_\theta \mathcal{Q}(\theta_k^{t,j}) - \nabla_\theta \mathcal{Q}(\theta_k^{t,*})\|_2 \leq H_{\mathcal{Q}} \|\theta_k^{t,j} - \theta_k^{t,*}\|_2$ and $\|\nabla_\theta \mathcal{Q}(\theta_k^{t,*})\|_2 \leq L_{\mathcal{Q}}$, along with the $L_2$-Lipschitz continuity which yields $\|L_{k,\text{rep}}^t(\theta_k^{t,j}) - L_{k,\text{rep}}^t(\theta_k^{t,*})\|_2 \leq L_2 \|\theta_k^{t,j} - \theta_k^{t,*}\|_2$, we obtain:
\begin{equation}
    \begin{split}
        &\langle \nabla_\theta \mathcal{Q}(\theta_k^{t,j}) \cdot L_{k,\text{rep}}^t(\theta_k^{t,j}) - \nabla_\theta\tilde{L}(\theta_k^{t,*}), \theta_k^{t,j} - \theta_k^{t,*} \rangle
        \\
        \leq& 
        (\|\nabla_\theta \mathcal{Q}(\theta_k^{t,j}) - \nabla_\theta \mathcal{Q}(\theta_k^{t,*})\|_2 \|L_{k,\text{rep}}^t(\theta_k^{t,j})\|_2
        +
        \|\nabla_\theta \mathcal{Q}(\theta_k^{t,*})\|_2 \|L_{k,\text{rep}}^t(\theta_k^{t,j}) - L_{k,\text{rep}}^t(\theta_k^{t,*})\|_2
        ) \|\theta_k^{t,j} - \theta_k^{t,*}\|_2
        \\
        \leq& (H_{\mathcal{Q}} \cdot \sup_{\theta \in \mathcal{B}_\delta}(L_{k,\text{rep}}^{t}(\theta)) + L_{\mathcal{Q}} L_2) \|\theta_k^{t,j} - \theta_k^{t,*}\|_2
    \end{split}
\end{equation}

Substituting this back into the original inequality, we obtain:
\begin{equation}
    \begin{split}
        \|\theta_k^{t,j+1} - \theta_k^{t,*}\|_2^2 
        \leq& (1 - 2\eta_k^{t,j}(\mu - (\delta L_{\mathcal{Q}}L_{\mathrm{lip}} + 2L_{\mathcal{Q}} L_2 + H_{\mathcal{Q}})))\|\theta_k^{t,j} - \theta_k^{t,*}\|_2^2 + (\eta_k^{t,j})^2\|\nabla_\theta L_{k, \text{total}}^t(\theta_k^{t,j}, \mathcal{Q}(\theta_k^{t,j}))\|_2^2
        \\
        &+ 2\eta_k^{t,j} (H_{\mathcal{Q}} \cdot \sup_{\theta \in \mathcal{B}_\delta}(L_{k,\text{rep}}^{t}(\theta)) + L_{\mathcal{Q}} L_2) \|\theta_k^{t,j} - \theta_k^{t,*}\|_2
        \\
        =& (1 - 2\eta_k^{t,j}(\mu - (\delta L_{\mathcal{Q}}L_{\mathrm{lip}} + 2L_{\mathcal{Q}} L_2 + H_{\mathcal{Q}}) - (H_{\mathcal{Q}} \cdot \sup_{\theta \in \mathcal{B}_\delta}(L_{k,\text{rep}}^{t}(\theta)) + L_{\mathcal{Q}} L_2)) 
        )\|\theta_k^{t,j} - \theta_k^{t,*}\|_2^2 
        \\
        &+ (\eta_k^{t,j})^2\|\nabla_\theta L_{k, \text{total}}^t(\theta_k^{t,j}, \mathcal{Q}(\theta_k^{t,j}))\|_2^2
    \end{split}
\end{equation}

Next, considering the term $\nabla_\theta L_{k, \text{total}}^t(\theta_k^{t,j}, \mathcal{Q}(\theta_k^{t,j}))$ and utilizing the first-order optimality condition at the saddle point $\nabla_\theta L_{k,\text{total}}^t(\theta_k^{t,*}) = \mathbf{0}$, we obtain:
\begin{equation}
    \begin{split}
        \nabla_\theta L_{k,\text{total}}^t(\theta_k^{t,j}, \mathcal{Q}(\theta_k^{t,j})) 
        =& \nabla_\theta\tilde{L}(\theta_k^{t,j}) - \nabla_\theta \mathcal{Q}(\theta_k^{t,j}) \cdot L_{k,\text{rep}}^t(\theta_k^{t,j})
        \\
        =& [\nabla_\theta\tilde{L}(\theta_k^{t,j}) - \nabla_\theta\tilde{L}(\theta_k^{t,*})] + [\nabla_\theta\tilde{L}(\theta_k^{t,*}) - \nabla_\theta \mathcal{Q}(\theta_k^{t,*}) \cdot L_{k,\text{rep}}^t(\theta_k^{t,*})]
        \\
        &+ [\nabla_\theta \mathcal{Q}(\theta_k^{t,*}) \cdot L_{k,\text{rep}}^t(\theta_k^{t,*}) - \nabla_\theta \mathcal{Q}(\theta_k^{t,j}) \cdot L_{k,\text{rep}}^t(\theta_k^{t,j})]
        \\
        =& [\nabla_\theta\tilde{L}(\theta_k^{t,j}) - \nabla_\theta\tilde{L}(\theta_k^{t,*})] + \nabla_\theta L_{k,\text{total}}^t(\theta_k^{t,*})
        \\
        &+ [\nabla_\theta \mathcal{Q}(\theta_k^{t,*}) \cdot L_{k,\text{rep}}^t(\theta_k^{t,*}) - \nabla_\theta \mathcal{Q}(\theta_k^{t,j}) \cdot L_{k,\text{rep}}^t(\theta_k^{t,j})]
        \\
        =& [\nabla_\theta\tilde{L}(\theta_k^{t,j}) - \nabla_\theta\tilde{L}(\theta_k^{t,*})] + [\nabla_\theta \mathcal{Q}(\theta_k^{t,*}) \cdot L_{k,\text{rep}}^t(\theta_k^{t,*}) - \nabla_\theta \mathcal{Q}(\theta_k^{t,j}) \cdot L_{k,\text{rep}}^t(\theta_k^{t,j})]
    \end{split}
\end{equation}

Taking the norm of the above expression, we obtain:
\begin{equation}
    \|\nabla_\theta L_{k,\text{total}}^t(\theta_k^{t,j}, \mathcal{Q}(\theta_k^{t,j}))\|_2 = \|\nabla_\theta\tilde{L}(\theta_k^{t,j}) - \nabla_\theta\tilde{L}(\theta_k^{t,*})\|_2 + \|\nabla_\theta \mathcal{Q}(\theta_k^{t,*}) \cdot L_{k,\text{rep}}^t(\theta_k^{t,*}) - \nabla_\theta \mathcal{Q}(\theta_k^{t,j}) \cdot L_{k,\text{rep}}^t(\theta_k^{t,j})\|_2
\end{equation}

For the first term $\|\nabla_\theta\tilde{L}(\theta_k^{t,j}) - \nabla_\theta\tilde{L}(\theta_k^{t,*})\|_2$, applying the integral form of the mean value theorem yields:
\begin{equation}
    \begin{split}
        \|\nabla_\theta\tilde{L}(\theta_k^{t,j}) - \nabla_\theta\tilde{L}(\theta_k^{t,*})\|_2
        \leq& \sup_{\xi \in \mathcal{B}_\delta}\|\nabla_\theta^2 \tilde{L}(\xi)\|_2 \|\theta_k^{t,j} - \theta_k^{t,*}\|_2
        \\
        \leq& \sup_{\xi \in \mathcal{B}_\delta}(\|\nabla_\theta^2 L_{k, \text{total}}^t(\xi, \lambda_k^{t,*})\|_2 + \|\nabla_\theta^2 \tilde{L}(\xi) - \nabla_\theta^2 L_{k, \text{total}}^t(\xi, \lambda_k^{t,*})\|_2) \|\theta_k^{t,j} - \theta_k^{t,*}\|_2
        \\
        \leq& \sup_{\xi \in \mathcal{B}_\delta}(\|\nabla_\theta^2 L_{k, \text{task}}^t(\xi) + \mathcal{Q} \cdot (\theta_k^{t,*})\nabla_\theta^2 L_{k, \text{rep}}^t(\xi)\|_2) \|\theta_k^{t,j} - \theta_k^{t,*}\|_2
        \\
        &+ \sup_{\xi \in \mathcal{B}_\delta}(\|\nabla_\theta^2 \tilde{L}(\xi) - \nabla_\theta^2 L_{k, \text{total}}^t(\xi, \lambda_k^{t,*})\|_2) \|\theta_k^{t,j} - \theta_k^{t,*}\|_2
        \\
        \leq& (L + \sup_{\xi \in \mathcal{B}_\delta}(\mathcal{Q}(\xi)) \cdot L)\|\theta_k^{t,j} - \theta_k^{t,*}\|_2 + (\delta L_{\mathcal{Q}}L_{\mathrm{lip}} + 2L_{\mathcal{Q}} L_2 + H_{\mathcal{Q}})\|\theta_k^{t,j} - \theta_k^{t,*}\|_2 
        \\
        =& ((1 + \sup_{\xi \in \mathcal{B}_\delta}(\mathcal{Q}(\xi)) + \delta L_{\mathcal{Q}}) L + 2L_{\mathcal{Q}} L_2 + H_{\mathcal{Q}})\|\theta_k^{t,j} - \theta_k^{t,*}\|_2
    \end{split}
\end{equation}

For the second term $\|\nabla_\theta \mathcal{Q}(\theta_k^{t,*}) L_{k,\text{rep}}^t(\theta_k^{t,*}) - \nabla_\theta \mathcal{Q}(\theta_k^{t,j}) L_{k,\text{rep}}^t(\theta_k^{t,j})\|_2$, we have:
\begin{equation}
    \begin{split}
        & \|\nabla_\theta \mathcal{Q}(\theta_k^{t,*}) \cdot L_{k,\text{rep}}^t(\theta_k^{t,*}) - \nabla_\theta \mathcal{Q}(\theta_k^{t,j}) \cdot L_{k,\text{rep}}^t(\theta_k^{t,j})\|_2
        \\
        \leq& \|\nabla_\theta \mathcal{Q}(\theta_k^{t,*}) \cdot L_{k,\text{rep}}^t(\theta_k^{t,*}) - \nabla_\theta \mathcal{Q}(\theta_k^{t,*}) \cdot L_{k,\text{rep}}^t(\theta_k^{t,j})\|_2
        + 
        \|\nabla_\theta \mathcal{Q}(\theta_k^{t,*}) \cdot L_{k,\text{rep}}^t(\theta_k^{t,j}) - \nabla_\theta \mathcal{Q}(\theta_k^{t,j}) \cdot L_{k,\text{rep}}^t(\theta_k^{t,j})\|_2
        \\
        \leq& \|\nabla_\theta \mathcal{Q}(\theta_k^{t,*})\|_2 \|L_{k,\text{rep}}^t(\theta_k^{t,*}) - L_{k,\text{rep}}^t(\theta_k^{t,j})\|_2
        +
        \|\nabla_\theta \mathcal{Q}(\theta_k^{t,*}) - \nabla_\theta \mathcal{Q}(\theta_k^{t,j})\|_2 \|L_{k,\text{rep}}^t(\theta_k^{t,j})\|_2
        \\
        \leq& [L_2 L_{\mathcal{Q}} +  H_{\mathcal{Q}} \cdot \sup_{\theta \in \mathcal{B}_\delta}(L_{k,\text{rep}}^t(\theta))] \|\theta_k^{t,j} - \theta_k^{t,*}\|_2
    \end{split}
\end{equation}

Consequently, we have:
\begin{equation}
    \|\nabla_\theta L_{k,\text{total}}^t(\theta_k^{t,j}, \mathcal{Q}(\theta_k^{t,j}))\|_2 
    \leq ((1 + \sup_{\xi \in \mathcal{B}_\delta}(\mathcal{Q}(\xi)) + \delta L_{\mathcal{Q}}) L + 3 L_{\mathcal{Q}} L_2 + H_{\mathcal{Q}} (1 + \sup_{\theta \in \mathcal{B}_\delta}(L_{k,\text{rep}}^t(\theta))))\|\theta_k^{t,j} - \theta_k^{t,*}\|_2
\end{equation}

Substituting this back into the original expression and applying the bound $\sup_{\theta \in \mathcal{B}_\delta} L_{k,\text{rep}}^{t}(\theta) \leq 1$ given by Assumption 5, we obtain:
\begin{equation}
    \begin{split}
        \|\theta_k^{t,j+1} - \theta_k^{t,*}\|_2^2 
        \leq& (1 - 2\eta_k^{t,j}(\mu - (\delta L_{\mathcal{Q}}L_{\mathrm{lip}} + 2L_{\mathcal{Q}} L_2 + H_{\mathcal{Q}}) - (H_{\mathcal{Q}} \cdot \sup_{\theta \in \mathcal{B}_\delta}(L_{k,\text{rep}}^{t}(\theta)) + L_{\mathcal{Q}} L_2)) 
        )\|\theta_k^{t,j} - \theta_k^{t,*}\|_2^2 
        \\
        &+ (\eta_k^{t,j})^2((1 + \sup_{\xi \in \mathcal{B}_\delta}(\mathcal{Q}(\xi)) + \delta L_{\mathcal{Q}}) L + 3 L_{\mathcal{Q}} L_2 + H_{\mathcal{Q}} (1 + \sup_{\theta \in \mathcal{B}_\delta}(L_{k,\text{rep}}^t(\theta))))^2 \|\theta_k^{t,j} - \theta_k^{t,*}\|_2^2
        \\
        \leq& (1 - 2\eta_k^{t,j}(\mu - (\delta L_{\mathcal{Q}}L_{\mathrm{lip}} + 3L_{\mathcal{Q}} L_2 + 2H_{\mathcal{Q}})))\|\theta_k^{t,j} - \theta_k^{t,*}\|_2^2 
        \\
        &+ (\eta_k^{t,j})^2((1 + \sup_{\xi \in \mathcal{B}_\delta}(\mathcal{Q}(\xi)) + \delta L_{\mathcal{Q}}) L + 3 L_{\mathcal{Q}} L_2 + 2H_{\mathcal{Q}})^2 \|\theta_k^{t,j} - \theta_k^{t,*}\|_2^2 
    \end{split}
\end{equation}

In the preceding proof, we have established that the composite objective $\tilde{L}(\theta) = L_{k,\text{task}}^t(\theta) + \mathcal{Q}(\theta)L_{k,\text{rep}}^t(\theta)$ is $(\mu - (\delta L_{\mathcal{Q}}L_{\mathrm{lip}} + 2L_{\mathcal{Q}} L_2 + H_{\mathcal{Q}}))$-strongly convex within the saddle point neighborhood $\mathcal{B}_{\delta}$. To simplify the notation, we define:
\begin{equation}
    \begin{split}
        \mu_1(\delta) =& \mu - (\delta L_{\mathcal{Q}}L_{\mathrm{lip}} + 2L_{\mathcal{Q}} L_2 + H_{\mathcal{Q}})
        \\
        L_{\mathrm{smooth}}(\delta) =& (1 + \sup_{\xi \in \mathcal{B}_\delta}(\mathcal{Q}(\xi)) + \delta L_{\mathcal{Q}}) L + 3 L_{\mathcal{Q}} L_2 + 2H_{\mathcal{Q}}
    \end{split}
\end{equation}

Consequently, the above inequality simplifies to $\|\theta_k^{t,j+1} - \theta_k^{t,*}\|_2^2 \leq (1 - 2\eta_k^{t,j} (\mu_1(\delta) - (L_{\mathcal{Q}} L_2 + H_{\mathcal{Q}})) + (\eta_k^{t,j})^2(L_{\mathrm{smooth}}(\delta))^2) \|\theta_k^{t,j} - \theta_k^{t,*}\|_2^2$. To ensure that it satisfies the single-step Lyapunov linear contraction form $\|\theta_k^{t,j+1} - \theta_k^{t,*}\|_2^2 \leq (1 - \eta_k^{t,j}\mu_1(\delta))\|\theta_k^{t,j} - \theta_k^{t,*}\|_2^2$, a constraint on the learning rate is required. This yields:
\begin{equation}
    - 2\eta_k^{t,j} \cdot (\mu_1(\delta) - (L_{\mathcal{Q}} L_2 + H_{\mathcal{Q}})) + (\eta_k^{t,j})^2(L_{\mathrm{smooth}}(\delta))^2 \leq -\eta_k^{t,j}\cdot\mu_1(\delta)
\end{equation}

Consequently, we have:
\begin{equation}
    2\eta_k^{t,j}(L_{\mathcal{Q}} L_2 + H_{\mathcal{Q}}) + (\eta_k^{t,j})^2(L_{\mathrm{smooth}}(\delta))^2 \leq \eta_k^{t,j}\cdot\mu_1(\delta)
\end{equation}

By allocating the target contraction equally to the first- and second-order terms, i.e., bounding each by $\eta_k^{t,j}\mu_1(\delta) / 2$, we obtain:
\begin{equation}
    \begin{cases} 
        (\eta_k^{t,j})^2(L_{\mathrm{smooth}}(\delta))^2 \leq \dfrac{\eta_k^{t,j}\mu_1(\delta)}{2}
        \\ 
        2\eta_k^{t,j}(L_{\mathcal{Q}} L_2 + H_{\mathcal{Q}}) \leq \dfrac{\eta_k^{t,j}\mu_1(\delta)}{2}
    \end{cases}
\end{equation}

To satisfy the first inequality, it is required that $\eta_k^{t,j} \leq \frac{\mu_1(\delta)}{2(L_{\mathrm{smooth}}(\delta))^2}$. For the second inequality, the condition $(L_{\mathcal{Q}} L_2 + H_{\mathcal{Q}}) \leq \frac{\mu_1(\delta)}{4}$ must hold, which leads to the conservative constraint on the learning rate: $\eta_k^{t,j} \leq \frac{\mu_1(\delta)}{4(L_{\mathcal{Q}} L_2 + H_{\mathcal{Q}})}$. Combining these conditions, the upper bound for the learning rate is obtained as:
\begin{equation}
    \eta_k^{t,j} \leq \min\left(\frac{\mu_1(\delta)}{2(L_{\mathrm{smooth}}(\delta))^2}, \frac{\mu_1(\delta)}{4 (L_{\mathcal{Q}} L_2 + H_{\mathcal{Q}})}\right)
\end{equation}

According to Assumption 3, the sum of squared learning rates converges, i.e., $\sum_{j=0}^{\infty}(\eta_k^{t,j})^2 < \infty$. This implies that $\lim_{j \to \infty} \eta_k^{t,j} = 0$, and thus there must exist $J_0 > 0$ such that the learning rate satisfies the aforementioned upper bound condition for all $j > J_0$. Consequently, within the saddle point neighborhood $\mathcal{B}_{\delta}$, the following contraction inequality holds:
\begin{equation}
    \|\theta_k^{t,j+1} - \theta_k^{t,*}\|_2^2 \leq (1 - \eta_k^{t,j}\cdot\mu_1(\delta))\|\theta_k^{t,j} - \theta_k^{t,*}\|_2^2
\end{equation}

Furthermore, for any $j \geq J_0$, applying the inequality $1 - \eta_k^{t,j}\mu_1(\delta) \leq \exp(- \eta_k^{t,j}\mu_1(\delta))$, we obtain:
\begin{equation}
    \begin{split}
        \|\theta_k^{t,j+1} - \theta_k^{t,*}\|_2^2 
        \leq& \|\theta_k^{t,J_0} - \theta_k^{t,*}\|_2^2 \prod_{m=J_0}^{j}(1 - \eta_k^{t,m}\cdot\mu_1(\delta))   
        \\
        \leq& \|\theta_k^{t,J_0} - \theta_k^{t,*}\|_2^2 \cdot \mathrm{exp}(-\mu_1(\delta) \cdot \sum_{m=J_0}^{j}\eta_k^{t,m})
    \end{split}
\end{equation}

Since the initial error $\|\theta_k^{t,J_0} - \theta_k^{t,*}\|_2^2$ is a finite value, and according to Assumption 3 the series $\sum_{m=J_0}^{\infty}\eta_k^{t,m}=\infty$ diverges, the exponential decay term approaches zero as $j \to \infty$. Therefore, we have:
\begin{equation}
    0 \leq \lim_{j\to\infty}\|\theta_k^{t,j+1} - \theta_k^{t,*}\|_2^2 
    \leq \|\theta_k^{t,J_0} - \theta_k^{t,*}\|_2^2 \lim_{j\to\infty} \mathrm{exp}(-\mu_1(\delta) \cdot \sum_{m=J_0}^{j}\eta_k^{t,m}) = 0
\end{equation}

Consequently, we obtain $\lim_{j\to\infty}\|\theta_k^{t,j} - \theta_k^{t,*}\|_2^2 = 0$, which implies:
\begin{equation}
    \lim_{j\to\infty} \theta_k^{t,j} = \theta_k^{t,*}
\end{equation}

By the continuity of $\mathcal{Q}(\theta)$, substituting the limit into the quotient mapping yields:
\begin{equation}
    \lim_{j\to\infty}\mathcal{Q}(\theta_k^{t,j}) = \mathcal{Q}(\lim_{j\to\infty}\theta_k^{t,j}) = \mathcal{Q}(\theta_k^{t,*}) = \lambda_k^{t,*}
\end{equation}

Consequently, within the $\delta$-neighborhood of the saddle point $(\theta_k^{t,*}, \lambda_k^{t,*})$, the sequence of iterates asymptotically converges to this saddle point. This completes the proof of Lemma 1.

\newpage
\subsection{Proof of Theorem 1}
\label{Appendix B.3}
\setcounter{theorem}{0}   
\begin{theorem}[Local Saddle-Point Convergence of the Responsive Gradient-Balanced Replay Scheme (RGBRS)]
    Under standard assumptions (see \hyperref[Appendix B.1]{Appendix B.1}), let $\theta_k^{t,0}$ be the initial model for client $k$ in FL round $t$. After $J$ local epochs via the Responsive Gradient-Balanced Replay Scheme, $\theta_k^{t,J}$ converges to the ideal saddle point $(\theta_k^{t,*}, \lambda_k^{t,*})$ within a truncation error bound:
    \begin{align}
      \mathbb{E}[L_{k,\text{total}}^{t}(\theta_k^{t,J}|\lambda_k^{t,J})] 
      \leq L_{k,\text{total}}^{t}(\theta_k^{t,*}|\lambda_k^{t,*}) + \mathcal{E}_{k, \text{bound}}^{t}(\{\eta_k^{t,j}\}_{j=0}^{J-1}, \|\theta_k^{t,0} - \theta_k^{t,*}\|_2^2)  
    \end{align}
    where the ideal saddle point $(\theta_k^{t,*}, \lambda_k^{t,*})$ satisfies the minimax equilibrium property for any $\theta \in \Omega_\theta$ and $\lambda \in \Omega_\lambda$:
    \begin{equation}
        L_{k,\text{total}}^{t}(\theta_k^{t,*}| \lambda) \leq L_{k,\text{total}}^{t}(\theta_k^{t,*}| \lambda_k^{t,*}) \leq L_{k,\text{total}}^{t}(\theta| \lambda_k^{t,*})
    \end{equation}
    The truncation error upper bound $\mathcal{E}_{k, \text{bound}}^{t}\left(\{\eta_k^{t,j}\}_{j=0}^{J-1}, \|\theta_k^{t,0} - \theta_k^{t,*}\|_2^2\right)$ is a deterministic function depending on the local step-size decay sequence and the initial parameter residual, defined as:
    \begin{equation}
        \begin{split}
            \mathcal{E}_{k, \text{bound}}^{t}\left(\{\eta_k^{t,j}\}_{j=0}^{J-1}, \|\theta_k^{t,0} - \theta_k^{t,*}\|_2^2\right)
            =& \frac{L}{2} \mathrm{exp}\left(- (\mu - 2 L_2 L_{\mathcal{Q}}) \sum_{j=0}^{J-1} \eta_k^{t,j}\right)\mathbb{E}_{\xi}[\|\theta_k^{t,0} - \theta_k^{t,*}\|_2^2]
            \\
            &+ \frac{L}{2} \sum_{j=0}^{J-1} \left(\eta_k^{t,j} \frac{{L_2}^2 {\epsilon_{\text{approx}}}^2}{\mu} + (\eta_k^{t,j})^2 G^2\right)\prod_{i=j+1}^{J-1}(1 - \eta_k^{t,i}(\mu - 2 L_2 L_{\mathcal{Q}}))
            \\
            &+ (L_{\mathcal{Q}}D_\theta + \epsilon_{\text{approx}}) \cdot \sup_{\theta \in \Omega_\theta} L_{k,\text{rep}}^{t}(\theta)
        \end{split}
    \end{equation}
    where $\epsilon_{\text{approx}} \triangleq \sup_{\theta \in \Omega_\theta} \left| \frac{\|\nabla_h L_{k,\text{rep}}^{t}(\theta)\|_2^2}{\|\nabla_h L_{k,\text{task}}^{t}(\theta)\|_2^2} - \frac{\|\nabla_\theta L_{k,\text{rep}}^{t}(\theta)\|_2^2}{\|\nabla_\theta L_{k,\text{task}}^{t}(\theta)\|_2^2} \right|$ denotes the supremum of the surrogate error for the global replay loss weight, and $L_{\mathcal{Q}}$ is the Lipschitz constant of the gradient quotient mapping over the model parameters $\mathcal{Q}(\theta) = \|\nabla_\theta L_{k,\text{task}}^t(\theta)\|_2^2 / \|\nabla_\theta L_{k,\text{rep}}^t(\theta)\|_2^2$, as specifically defined in the proof of Lemma 1.
\end{theorem}

\textbf{Proof}:

The existence and asymptotic stability of the ideal saddle point $(\theta_k^{t,*}, \lambda_k^{t,*})$ have been established in Lemma 1, and thus the proof is omitted here.

Consider the replay loss weight error term $\frac{\|\nabla_h L_{k,\text{rep}}^{t}(\theta)\|_2^2}{\|\nabla_h L_{k,\text{task}}^{t}(\theta)\|_2^2} - \frac{\|\nabla_\theta L_{k,\text{rep}}^{t}(\theta)\|_2^2}{\|\nabla_\theta L_{k,\text{task}}^{t}(\theta)\|_2^2}$. To simplify the notation, let $f_1(\theta) = \|\nabla_h L_{k,\text{rep}}^{t}(\theta)\|_2^2$, $g_1(\theta) = \|\nabla_h L_{k,\text{task}}^{t}(\theta)\|_2^2$, $f_2(\theta) = \|\nabla_\theta L_{k,\text{rep}}^{t}(\theta)\|_2^2$, and $g_2(\theta) = \|\nabla_\theta L_{k,\text{task}}^{t}(\theta)\|_2^2$. Then, we have:
\begin{equation}
    \begin{split}
        \left| \frac{\|\nabla_h L_{k,\text{rep}}^{t}(\theta)\|_2^2}{\|\nabla_h L_{k,\text{task}}^{t}(\theta)\|_2^2} - \frac{\|\nabla_\theta L_{k,\text{rep}}^{t}(\theta)\|_2^2}{\|\nabla_\theta L_{k,\text{task}}^{t}(\theta)\|_2^2} \right| 
        =& \left| \frac{f_1(\theta)}{g_1(\theta)} - \frac{f_2(\theta)}{g_2(\theta)} \right|
        = \frac{\left|f_1(\theta) g_2(\theta) - f_2(\theta) g_1(\theta)\right|}{g_1(\theta) g_2(\theta)} 
        \\
        =& \frac{\left|f_1(\theta) g_2(\theta) - f_1(\theta) g_1(\theta) + f_1(\theta) g_1(\theta) - f_2(\theta) g_1(\theta)\right|}{g_1(\theta) g_2(\theta)} 
        \\
        =& \frac{|f_1(\theta)(g_2(\theta) - g_1(\theta)) + g_1(\theta)(f_1(\theta) - f_2(\theta))|}{g_1(\theta) g_2(\theta)} 
        \\
        \leq& \frac{|f_1(\theta)| |g_1(\theta) - g_2(\theta)| + |g_1(\theta)| |f_1(\theta) - f_2(\theta)|}{g_1(\theta) g_2(\theta)}
    \end{split}
\end{equation}

Under Assumption 1, we have $\|\nabla_\theta L_{k,\text{task}}^{t}(\theta)\|_2 \leq L_2$. According to Proposition 1 in \cite{10}, it holds that $\|\nabla_h L_{k,\text{task}}^{t}(\theta)\|_2 \leq \|\nabla_\theta L_{k,\text{task}}^{t}(\theta)\|_2$ (i.e., $g_1(\theta) \leq f_1(\theta)$). Furthermore, under Assumptions 1 and 2, the mapping $\theta \mapsto \min(\|\nabla_h L_{k,\text{task}}^{t}(\theta)\|_2, \|\nabla_\theta L_{k,\text{task}}^{t}(\theta)\|_2)$ is a continuous positive function on $\Omega_\theta$. Therefore, we define:
\begin{equation}
    \epsilon_{\min} \triangleq \inf_{\theta \in \Omega_\theta} \min(\|\nabla_h L_{k,\text{task}}^{t}(\theta)\|_2, \|\nabla_\theta L_{k,\text{task}}^{t}(\theta)\|_2) > 0
\end{equation}

Substituting these into the original expression yields:
\begin{equation}
    \begin{split}
        \left| \frac{\|\nabla_h L_{k,\text{rep}}^{t}(\theta)\|_2^2}{\|\nabla_h L_{k,\text{task}}^{t}(\theta)\|_2^2} - \frac{\|\nabla_\theta L_{k,\text{rep}}^{t}(\theta)\|_2^2}{\|\nabla_\theta L_{k,\text{task}}^{t}(\theta)\|_2^2} \right| 
        \leq& \frac{|f_1(\theta)| |g_1(\theta) - g_2(\theta)| + |g_1(\theta)| |f_1(\theta) - f_2(\theta)|}{g_1(\theta) g_2(\theta)}
        \\
        \leq& \frac{f_1(\theta)}{g_1(\theta) g_2(\theta)}(|g_1(\theta) - g_2(\theta)| + |f_1(\theta) - f_2(\theta)|)
        \\
        \leq& \frac{{L_2}^2}{{\epsilon_{\min}}^4}(|g_1(\theta) - g_2(\theta)| + |f_1(\theta) - f_2(\theta)|)
        \\
        =& \frac{{L_2}^2}{{\epsilon_{\min}}^4}(|\|\nabla_\theta L_{k,\text{task}}^{t}(\theta)\|_2^2 - \|\nabla_h L_{k,\text{task}}^{t}(\theta)\|_2^2|)
        \\
        &+ \frac{{L_2}^2}{{\epsilon_{\min}}^4}(|\|\nabla_\theta L_{k,\text{rep}}^{t}(\theta)\|_2^2 - \|\nabla_h L_{k,\text{rep}}^{t}(\theta)\|_2^2|)
    \end{split}
\end{equation}

Thus, the surrogate error is bounded. Furthermore, we define the supremum of the global replay loss weight surrogate error as $\epsilon_{\text{approx}}$:
\begin{equation}
    \epsilon_{\text{approx}} \triangleq \sup_{\theta \in \Omega_\theta} \left| \frac{\|\nabla_h L_{k,\text{rep}}^{t}(\theta)\|_2^2}{\|\nabla_h L_{k,\text{task}}^{t}(\theta)\|_2^2} - \frac{\|\nabla_\theta L_{k,\text{rep}}^{t}(\theta)\|_2^2}{\|\nabla_\theta L_{k,\text{task}}^{t}(\theta)\|_2^2} \right|
\end{equation}

Consider the $(j+1)$-th model iteration. The model parameters before the update are denoted as $\theta_k^{t,j}$, and the replay loss weight is given by $\lambda_k^{t,j+1} = \|\nabla_h L_{k,\text{rep}}^{t}(\theta_k^{t,j})\|_2^2 / \|\nabla_h L_{k,\text{task}}^{t}(\theta_k^{t,j})\|_2^2$. The gradient quotient mapping for the replay loss weight with respect to the model parameters is defined as $\mathcal{Q}(\theta_k^{t,j}) = \|\nabla_\theta L_{k,\text{rep}}^{t}(\theta_k^{t,j})\|_2^2 / \|\nabla_\theta L_{k,\text{task}}^{t}(\theta_k^{t,j})\|_2^2$, which has been shown to be $L_{\mathcal{Q}}$-Lipschitz continuous in Lemma 1. By the definition of $\epsilon_{\text{approx}}$, it is evident that $|\lambda_k^{t,j+1} - \mathcal{Q}(\theta_k^{t,j})| \leq \epsilon_{\text{approx}}$. Consequently, regarding the ideal saddle point $(\theta_k^{t,*}, \lambda_k^{t,*})$, we have:
\begin{equation}
    \begin{split}
        |\lambda_k^{t,j+1} - \lambda_k^{t,*}| 
        =& |\lambda_k^{t,j+1} - \mathcal{Q}(\theta_k^{t,j}) + \mathcal{Q}(\theta_k^{t,j}) - \lambda_k^{t,*}|
        \\
        \leq& |\lambda_k^{t,j+1} - \mathcal{Q}(\theta_k^{t,j})| + |\mathcal{Q}(\theta_k^{t,j}) - \mathcal{Q}(\theta_k^{t,*})| 
        \\
        \leq& \epsilon_{\text{approx}} + L_{\mathcal{Q}}\|\theta_k^{t,j} - \theta_k^{t,*}\|_2
    \end{split}
\end{equation}

Consider a random mini-batch $\xi$, with the corresponding stochastic gradient given by $\hat{\nabla}_{\theta} L_{k,\text{total}}^{t}(\theta_k^{t,j}, \xi|\lambda_k^{t,j})$. Expanding the $(j+1)$-th model update yields:
\begin{equation}
    \begin{split}
        \|\theta_k^{t,j+1} - \theta_k^{t,*}\|_2^2 
        =& \|\theta_k^{t,j} - \eta_k^{t,j}\hat{\nabla}_{\theta} L_{k,\text{total}}^{t}(\theta_k^{t,j}, \xi|\lambda_k^{t,j}) - \theta_k^{t,*}\|_2^2
        \\
        =& \|\theta_k^{t,j} - \theta_k^{t,*}\|_2^2 - 2\langle \theta_k^{t,j} - \theta_k^{t,*}, \eta_k^{t,j}\hat{\nabla}_{\theta} L_{k,\text{total}}^{t}(\theta_k^{t,j}, \xi|\lambda_k^{t,j}) \rangle + \|\eta_k^{t,j}\hat{\nabla}_{\theta} L_{k,\text{total}}^{t}(\theta_k^{t,j}, \xi|\lambda_k^{t,j})\|_2^2
        \\
        =& \|\theta_k^{t,j} - \theta_k^{t,*}\|_2^2 - 2\eta_k^{t,j}\langle \theta_k^{t,j} - \theta_k^{t,*}, \hat{\nabla}_{\theta} L_{k,\text{total}}^{t}(\theta_k^{t,j}, \xi|\lambda_k^{t,j}) \rangle + (\eta_k^{t,j})^2\|\hat{\nabla}_{\theta} L_{k,\text{total}}^{t}(\theta_k^{t,j}, \xi|\lambda_k^{t,j})\|_2^2
    \end{split}
\end{equation}

Taking the expectation of the above expression with respect to the random mini-batch $\xi$. By the unbiasedness of stochastic sampling, we have $\mathbb{E}_{\xi}[\hat{\nabla}_{\theta} L_{k,\text{total}}^{t}(\theta_k^{t,j}, \xi|\lambda_k^{t,j})] = \nabla_{\theta} L_{k,\text{total}}^{t}(\theta_k^{t,j}|\lambda_k^{t,j})$; furthermore, under Assumption 13, it holds that $\mathbb{E}_{\xi}[\|\hat{\nabla}_{\theta} L_{k,\text{total}}^{t}(\theta_k^{t,j}, \xi|\lambda_k^{t,j})\|_2^2] \leq G^2$. Combining these results yields:
\begin{equation}
    \begin{split}
        \mathbb{E}_{\xi}[\|\theta_k^{t,j+1} - \theta_k^{t,*}\|_2^2] 
        =& \|\theta_k^{t,j} - \theta_k^{t,*}\|_2^2 - 2\eta_k^{t,j}\langle \theta_k^{t,j} - \theta_k^{t,*}, \mathbb{E}_{\xi}[\hat{\nabla}_{\theta} L_{k,\text{total}}^{t}(\theta_k^{t,j}, \xi|\lambda_k^{t,j})] \rangle 
        \\
        &+ (\eta_k^{t,j})^2 \mathbb{E}_{\xi}[\|\hat{\nabla}_{\theta} L_{k,\text{total}}^{t}(\theta_k^{t,j}, \xi|\lambda_k^{t,j})\|_2^2]
        \\
        \leq& \|\theta_k^{t,j} - \theta_k^{t,*}\|_2^2 - 2\eta_k^{t,j}\langle \theta_k^{t,j} - \theta_k^{t,*}, \nabla_{\theta} L_{k,\text{total}}^{t}(\theta_k^{t,j}|\lambda_k^{t,j}) \rangle + (\eta_k^{t,j})^2 G^2
    \end{split}
\end{equation}

Consider the inner product term $\langle \theta_k^{t,j} - \theta_k^{t,*}, \nabla_{\theta} L_{k,\text{total}}^{t}(\theta_k^{t,j}|\lambda_k^{t,j}) \rangle$. By definition, $L_{k,\text{total}}^{t}(\theta_k^{t,j}|\lambda_k^{t,j}) = L_{k,\text{task}}^{t}(\theta_k^{t,j}) + \lambda_k^{t,j} L_{k,\text{rep}}^{t}(\theta_k^{t,j})$. Since $\lambda_k^{t,j}$ is treated as a constant with respect to $\theta$, we have:
\begin{equation}
    \begin{split}
        \nabla_{\theta} L_{k,\text{total}}^{t}(\theta_k^{t,j}|\lambda_k^{t,j}) 
        =& \nabla_{\theta}L_{k,\text{task}}^{t}(\theta_k^{t,j}) + \lambda_k^{t,j} \cdot \nabla_{\theta}L_{k,\text{rep}}^{t}(\theta_k^{t,j})
        \\
        =& \nabla_{\theta}L_{k,\text{task}}^{t}(\theta_k^{t,j}) + (\lambda_k^{t,*} + (\lambda_k^{t,j} - \lambda_k^{t,*})) \cdot \nabla_{\theta}L_{k,\text{rep}}^{t}(\theta_k^{t,j})
        \\
        =& \nabla_{\theta}L_{k,\text{task}}^{t}(\theta_k^{t,j}) + \lambda_k^{t,*} \cdot \nabla_{\theta}L_{k,\text{rep}}^{t}(\theta_k^{t,j}) + (\lambda_k^{t,j} - \lambda_k^{t,*}) \cdot \nabla_{\theta}L_{k,\text{rep}}^{t}(\theta_k^{t,j})
        \\
        =& \nabla_{\theta} L_{k,\text{total}}^{t}(\theta_k^{t,j}|\lambda_k^{t,*}) + (\lambda_k^{t,j} - \lambda_k^{t,*}) \cdot \nabla_{\theta}L_{k,\text{rep}}^{t}(\theta_k^{t,j})
    \end{split}
\end{equation}

Substituting the above results into the previous expression. Under Assumption 11, we have $\langle \nabla_{\theta} L_{k,\text{total}}^{t}(\theta_k^{t,j}|\lambda_k^{t,*}), \theta_k^{t,j} - \theta_k^{t,*}\rangle \geq \mu \|\theta_k^{t,j} - \theta_k^{t,*}\|_2^2$; under Assumption 1, it holds that $\|\nabla_{\theta}L_{k,\text{rep}}^{t}(\theta_k^{t,j})\|_2 \leq L_2$. Applying the Cauchy-Schwarz inequality yields:
\begin{equation}
    \begin{split}
        \mathbb{E}_{\xi}[\|\theta_k^{t,j+1} - \theta_k^{t,*}\|_2^2] 
        \leq& \|\theta_k^{t,j} - \theta_k^{t,*}\|_2^2 
        - 2\eta_k^{t,j}\langle \nabla_{\theta} L_{k,\text{total}}^{t}(\theta_k^{t,j}|\lambda_k^{t,*}), \theta_k^{t,j} - \theta_k^{t,*}\rangle 
        \\
        &- 2\eta_k^{t,j}(\lambda_k^{t,j} - \lambda_k^{t,*}) \langle \nabla_{\theta}L_{k,\text{rep}}^{t}(\theta_k^{t,j}), \theta_k^{t,j} - \theta_k^{t,*}\rangle 
        + (\eta_k^{t,j})^2 G^2
        \\
        \leq& \|\theta_k^{t,j} - \theta_k^{t,*}\|_2^2 
        - 2\eta_k^{t,j}\mu \|\theta_k^{t,j} - \theta_k^{t,*}\|_2^2
        \\
        &+ 2\eta_k^{t,j}|\lambda_k^{t,j} - \lambda_k^{t,*}| \langle \nabla_{\theta}L_{k,\text{rep}}^{t}(\theta_k^{t,j}), \theta_k^{t,j} - \theta_k^{t,*}\rangle 
        + (\eta_k^{t,j})^2 G^2
        \\
        \leq& (1 - 2\eta_k^{t,j}\mu)\|\theta_k^{t,j} - \theta_k^{t,*}\|_2^2 
        + 2\eta_k^{t,j}|\lambda_k^{t,j} - \lambda_k^{t,*}| \cdot \|\nabla_{\theta}L_{k,\text{rep}}^{t}(\theta_k^{t,j})\|_2 \|\theta_k^{t,j} - \theta_k^{t,*}\|_2
        + (\eta_k^{t,j})^2 G^2
        \\
        \leq& (1 - 2\eta_k^{t,j}\mu)\|\theta_k^{t,j} - \theta_k^{t,*}\|_2^2 
        + 2\eta_k^{t,j}L_2|\lambda_k^{t,j} - \lambda_k^{t,*}| \cdot \|\theta_k^{t,j} - \theta_k^{t,*}\|_2
        + (\eta_k^{t,j})^2 G^2
    \end{split}
\end{equation}

Furthermore, substituting $|\lambda_k^{t,j} - \lambda_k^{t,*}| \leq L_{\mathcal{Q}}\|\theta_k^{t,j} - \theta_k^{t,*}\|_2 + \epsilon_{\text{approx}}$ yields:
\begin{equation}
    \begin{split}
        \mathbb{E}_{\xi}[\|\theta_k^{t,j+1} - \theta_k^{t,*}\|_2^2] 
        \leq& (1 - 2\eta_k^{t,j}\mu)\|\theta_k^{t,j} - \theta_k^{t,*}\|_2^2 
        + 2\eta_k^{t,j}L_2|\lambda_k^{t,j} - \lambda_k^{t,*}| \cdot \|\theta_k^{t,j} - \theta_k^{t,*}\|_2
        + (\eta_k^{t,j})^2 G^2
        \\
        \leq& (1 - 2\eta_k^{t,j}\mu)\|\theta_k^{t,j} - \theta_k^{t,*}\|_2^2 + 2\eta_k^{t,j}L_2
        (L_{\mathcal{Q}}\|\theta_k^{t,j} - \theta_k^{t,*}\|_2 + \epsilon_{\text{approx}})\|\theta_k^{t,j} - \theta_k^{t,*}\|_2 + (\eta_k^{t,j})^2 G^2
        \\
        =& (1 - 2\eta_k^{t,j}\mu + 2\eta_k^{t,j} L_2 L_{\mathcal{Q}}) \|\theta_k^{t,j} - \theta_k^{t,*}\|_2^2 + 2\eta_k^{t,j} L_2 \epsilon_{\text{approx}}\|\theta_k^{t,j} - \theta_k^{t,*}\|_2 + (\eta_k^{t,j})^2 G^2
    \end{split}
\end{equation}

For the second term, by Young's inequality, we have $2L_2 \epsilon_{\text{approx}}\|\theta_k^{t,j} - \theta_k^{t,*}\|_2 \leq \mu\|\theta_k^{t,j} - \theta_k^{t,*}\|_2^2 + \frac{{L_2}^2 {\epsilon_{\text{approx}}}^2}{\mu}$. Substituting this into the above expression yields:
\begin{equation}
    \begin{split}
        \mathbb{E}_{\xi}[\|\theta_k^{t,j+1} - \theta_k^{t,*}\|_2^2] 
        \leq& (1 - 2\eta_k^{t,j}\mu + 2\eta_k^{t,j} L_2 L_{\mathcal{Q}}) \|\theta_k^{t,j} - \theta_k^{t,*}\|_2^2 
        + \eta_k^{t,j} \mu\|\theta_k^{t,j} - \theta_k^{t,*}\|_2^2 
        \\
        &+ \eta_k^{t,j} \frac{{L_2}^2 {\epsilon_{\text{approx}}}^2}{\mu}
        + (\eta_k^{t,j})^2 G^2
        \\
        =& (1 - \eta_k^{t,j}\mu + 2\eta_k^{t,j} L_2 L_{\mathcal{Q}}) \|\theta_k^{t,j} - \theta_k^{t,*}\|_2^2 + \eta_k^{t,j} \frac{{L_2}^2 {\epsilon_{\text{approx}}}^2}{\mu}
        + (\eta_k^{t,j})^2 G^2
        \\
        =& (1 - \eta_k^{t,j}(\mu - 2 L_2 L_{\mathcal{Q}})) \|\theta_k^{t,j} - \theta_k^{t,*}\|_2^2 + \eta_k^{t,j} \frac{{L_2}^2 {\epsilon_{\text{approx}}}^2}{\mu}
        + (\eta_k^{t,j})^2 G^2
        \\
        =& (1 - \eta_k^{t,j}(\mu - 2 L_2 L_{\mathcal{Q}})) \mathbb{E}_{\xi}[\|\theta_k^{t,j} - \theta_k^{t,*}\|_2^2] + \eta_k^{t,j} \frac{{L_2}^2 {\epsilon_{\text{approx}}}^2}{\mu}
        + (\eta_k^{t,j})^2 G^2
    \end{split}
\end{equation}

Taking the expectation of both sides with respect to $\xi$ yields:
\begin{equation}
    \begin{split}
        \mathbb{E}_{\xi}[\|\theta_k^{t,j+1} - \theta_k^{t,*}\|_2^2] \leq (1 - \eta_k^{t,j}(\mu - 2 L_2 L_{\mathcal{Q}})) \mathbb{E}_{\xi}[\|\theta_k^{t,j} - \theta_k^{t,*}\|_2^2] + \eta_k^{t,j} \frac{{L_2}^2 {\epsilon_{\text{approx}}}^2}{\mu}
        + (\eta_k^{t,j})^2 G^2
    \end{split}
\end{equation}

Next, considering the entire $J$-step iteration process, we obtain:
\begin{equation}
    \begin{split}
        \mathbb{E}_{\xi}[\|\theta_k^{t,J} - \theta_k^{t,*}\|_2^2] 
        \leq& (1 - \eta_k^{t,J-1}(\mu - 2 L_2 L_{\mathcal{Q}})) \mathbb{E}_{\xi}[\|\theta_k^{t,J-1} - \theta_k^{t,*}\|_2^2] + \eta_k^{t,J-1} \frac{{L_2}^2 {\epsilon_{\text{approx}}}^2}{\mu}
        + (\eta_k^{t,J-1})^2 G^2
        \\
        \leq& (1 - \eta_k^{t,J-1}(\mu - 2 L_2 L_{\mathcal{Q}})) (1 - \eta_k^{t,J-2}(\mu - 2 L_2 L_{\mathcal{Q}})) \mathbb{E}_{\xi}[\|\theta_k^{t,J-2} - \theta_k^{t,*}\|_2^2] 
        \\
        &+ (1 - \eta_k^{t,J-1}(\mu - 2 L_2 L_{\mathcal{Q}}))\left(\eta_k^{t,J-2} \frac{{L_2}^2 {\epsilon_{\text{approx}}}^2}{\mu}+ (\eta_k^{t,J-2})^2 G^2\right)
        \\
        &+ \left(\eta_k^{t,J-1} \frac{{L_2}^2 {\epsilon_{\text{approx}}}^2}{\mu}
        + (\eta_k^{t,J-1})^2 G^2\right)
        \\
        =& \left[\prod_{j=J-2}^{J-1} \left(1 - \eta_k^{t,j} (\mu - 2 L_2 L_{\mathcal{Q}})\right) \right] \mathbb{E}_{\xi}[\|\theta_k^{t,J-2} - \theta_k^{t,*}\|_2^2]
        \\
        &+ \sum_{j=J-2}^{J-1} \left(\eta_k^{t,j} \frac{{L_2}^2 {\epsilon_{\text{approx}}}^2}{\mu}
        + (\eta_k^{t,j})^2 G^2\right)\prod_{i=j+1}^{J-1}(1 - \eta_k^{t,i}(\mu - 2 L_2 L_{\mathcal{Q}}))
        \\
        \leq& \ldots
        \\
        \leq& \left[\prod_{j=0}^{J-1} \left(1 - \eta_k^{t,j} (\mu - 2 L_2 L_{\mathcal{Q}})\right) \right] \mathbb{E}_{\xi}[\|\theta_k^{t,0} - \theta_k^{t,*}\|_2^2]
        \\
        &+ \sum_{j=0}^{J-1} \left(\eta_k^{t,j} \frac{{L_2}^2 {\epsilon_{\text{approx}}}^2}{\mu}
        + (\eta_k^{t,j})^2 G^2\right)\prod_{i=j+1}^{J-1}(1 - \eta_k^{t,i}(\mu - 2 L_2 L_{\mathcal{Q}}))
    \end{split}
\end{equation}

Applying the inequality $1-x \leq \mathrm{exp}(-x)$ to the product term in the above expression yields:
\begin{equation}
    \begin{split}
        \mathbb{E}_{\xi}[\|\theta_k^{t,J} - \theta_k^{t,*}\|_2^2] 
        \leq& \left[\prod_{j=0}^{J-1} \left(1 - \eta_k^{t,j} (\mu - 2 L_2 L_{\mathcal{Q}})\right) \right] \mathbb{E}_{\xi}[\|\theta_k^{t,0} - \theta_k^{t,*}\|_2^2]
        \\
        &+ \sum_{j=0}^{J-1} \left(\eta_k^{t,j} \frac{{L_2}^2 {\epsilon_{\text{approx}}}^2}{\mu}
        + (\eta_k^{t,j})^2 G^2\right)\prod_{i=j+1}^{J-1}(1 - \eta_k^{t,i}(\mu - 2 L_2 L_{\mathcal{Q}}))
        \\
        \leq& \mathrm{exp}\left(- (\mu - 2 L_2 L_{\mathcal{Q}}) \sum_{j=0}^{J-1} \eta_k^{t,j}\right)\mathbb{E}_{\xi}[\|\theta_k^{t,0} - \theta_k^{t,*}\|_2^2]
        \\
        &+ \sum_{j=0}^{J-1} \left(\eta_k^{t,j} \frac{{L_2}^2 {\epsilon_{\text{approx}}}^2}{\mu}
        + (\eta_k^{t,j})^2 G^2\right) \mathrm{exp}\left(- (\mu - 2 L_2 L_{\mathcal{Q}}) \sum_{i=j+1}^{J-1} \eta_k^{t,i}\right)
    \end{split}
\end{equation}

Consequently, we obtain the optimization error bound after $J$ steps of stochastic gradient descent. This bound mainly comprises two components: the residual contribution of the initial parameter residual after exponential decay, and the accumulated effect of the stochastic gradient noise and the replay weight approximation error over the iteration history.

Next, we expand the loss function after the $J$-th iteration, yielding:
\begin{equation}
    \begin{split}
        L_{k,\text{total}}^{t}(\theta_k^{t,J}|\lambda_k^{t,J}) 
        =& L_{k,\text{task}}^{t}(\theta_k^{t,J}) + \lambda_k^{t,J} \cdot L_{k,\text{rep}}^{t}(\theta_k^{t,J})
        \\
        =& L_{k,\text{task}}^{t}(\theta_k^{t,J}) + (\lambda_k^{t,*} + (\lambda_k^{t,J} - \lambda_k^{t,*})) \cdot L_{k,\text{rep}}^{t}(\theta_k^{t,J})
        \\
        =& L_{k,\text{task}}^{t}(\theta_k^{t,J}) + \lambda_k^{t,*} \cdot L_{k,\text{rep}}^{t}(\theta_k^{t,J}) + (\lambda_k^{t,J} - \lambda_k^{t,*}) \cdot L_{k,\text{rep}}^{t}(\theta_k^{t,J})
        \\
        =& L_{k,\text{total}}^{t}(\theta_k^{t,J}|\lambda_k^{t,*}) + (\lambda_k^{t,J} - \lambda_k^{t,*}) \cdot L_{k,\text{rep}}^{t}(\theta_k^{t,J})
    \end{split}
\end{equation}

For the first term $L_{k,\text{total}}^{t}(\theta_k^{t,J}|\lambda_k^{t,*})$, under the $L$-smoothness in Assumption 10, we apply the second-order Taylor upper bound at $\theta_k^{t,*}$. Substituting the stationary condition $\nabla_{\theta} L_{k,\text{total}}^{t}(\theta_k^{t,*} \mid \lambda_k^{t,*}) = \mathbf{0}$ yields:
\begin{equation}
    \begin{split}
        L_{k,\text{total}}^{t}(\theta_k^{t,J}|\lambda_k^{t,*})
        \leq& L_{k,\text{total}}^{t}(\theta_k^{t,*}|\lambda_k^{t,*}) + \langle \nabla_{\theta} L_{k,\text{total}}^{t}(\theta_k^{t,*} \mid \lambda_k^{t,*}), \theta_k^{t,J} - \theta_k^{t,*} \rangle + \frac{L}{2} \|\theta_k^{t,J} - \theta_k^{t,*}\|_2^2
        \\
        =& L_{k,\text{total}}^{t}(\theta_k^{t,*}|\lambda_k^{t,*}) + 0 + \frac{L}{2} \|\theta_k^{t,J} - \theta_k^{t,*}\|_2^2
    \end{split}
\end{equation}

Taking the expectation of both sides with respect to $\xi$ yields:
\begin{equation}
    \mathbb{E}_{\xi}\left[L_{k,\text{total}}^{t}(\theta_k^{t,J}|\lambda_k^{t,*})\right] 
    \leq L_{k,\text{total}}^{t}(\theta_k^{t,*}|\lambda_k^{t,*}) + \frac{L}{2}\mathbb{E}_{\xi}\left[\|\theta_k^{t,J} - \theta_k^{t,*}\|_2^2\right]
\end{equation}

For the second term $(\lambda_k^{t,J} - \lambda_k^{t,*}) L_{k,\text{rep}}^{t}(\theta_k^{t,J})$, taking its absolute value and then the expectation with respect to $\xi$ yields:
\begin{equation}
    \begin{split}
        \mathbb{E}_{\xi}\left[(\lambda_k^{t,J} - \lambda_k^{t,*}) \cdot L_{k,\text{rep}}^{t}(\theta_k^{t,J})\right] 
        \leq& \mathbb{E}_{\xi}\left[|(\lambda_k^{t,J} - \lambda_k^{t,*}) | \cdot |L_{k,\text{rep}}^{t}(\theta_k^{t,J})|\right]
        \\
        \leq& \mathbb{E}_{\xi}\left[|(\lambda_k^{t,J} - \lambda_k^{t,*})|\right] \cdot \left(\sup_{\theta \in \Omega_\theta} L_{k,\text{rep}}^{t}(\theta)\right)
    \end{split}
\end{equation}

Under Assumption 2, $\Omega_\theta$ is a compact set with diameter $D_\theta \triangleq \sup_{\theta_a,\theta_b \in \Omega_\theta} \|\theta_a-\theta_b\|_2 < \infty$. By the $L_{\mathcal{Q}}$-Lipschitz continuity of the quotient mapping $\mathcal{Q}(\theta)$, we have $|\lambda_k^{t,J} - \lambda_k^{t,*}| \leq L_{\mathcal{Q}}D_\theta + \epsilon_{\text{approx}}$. Substituting this into the previous expression yields:
\begin{equation}
    \mathbb{E}_{\xi}\left[(\lambda_k^{t,J} - \lambda_k^{t,*}) \cdot L_{k,\text{rep}}^{t}(\theta_k^{t,J})\right] 
    \leq (L_{\mathcal{Q}}D_\theta + \epsilon_{\text{approx}}) \cdot \sup_{\theta \in \Omega_\theta} L_{k,\text{rep}}^{t}(\theta)
\end{equation}

Consequently, taking the expectation of the loss function $L_{k,\text{total}}^{t}(\theta_k^{t,J}|\lambda_k^{t,J})$ after the $J$-th iteration with respect to $\xi$ yields:
\begin{equation}
    \begin{split}
        \mathbb{E}_{\xi}\left[L_{k,\text{total}}^{t}(\theta_k^{t,J}|\lambda_k^{t,J})\right] 
        =& \mathbb{E}_{\xi}\left[L_{k,\text{total}}^{t}(\theta_k^{t,J}|\lambda_k^{t,*})\right] + \mathbb{E}_{\xi}\left[(\lambda_k^{t,J} - \lambda_k^{t,*}) \cdot L_{k,\text{rep}}^{t}(\theta_k^{t,J})\right]
        \\
        \leq& L_{k,\text{total}}^{t}(\theta_k^{t,*}|\lambda_k^{t,*}) + \frac{L}{2}\mathbb{E}_{\xi}\left[\|\theta_k^{t,J} - \theta_k^{t,*}\|_2^2\right]
        + (L_{\mathcal{Q}}D_\theta + \epsilon_{\text{approx}}) \cdot \sup_{\theta \in \Omega_\theta} L_{k,\text{rep}}^{t}(\theta)
        \\
        \leq& L_{k,\text{total}}^{t}(\theta_k^{t,*}|\lambda_k^{t,*}) + \frac{L}{2} \mathrm{exp}\left(- (\mu - 2 L_2 L_{\mathcal{Q}}) \sum_{j=0}^{J-1} \eta_k^{t,j}\right)\mathbb{E}_{\xi}[\|\theta_k^{t,0} - \theta_k^{t,*}\|_2^2]
        \\
        &+ \frac{L}{2} \sum_{j=0}^{J-1} \left(\eta_k^{t,j} \frac{{L_2}^2 {\epsilon_{\text{approx}}}^2}{\mu} + (\eta_k^{t,j})^2 G^2\right)\prod_{i=j+1}^{J-1}(1 - \eta_k^{t,i}(\mu - 2 L_2 L_{\mathcal{Q}}))
        \\
        &+ (L_{\mathcal{Q}}D_\theta + \epsilon_{\text{approx}}) \cdot \sup_{\theta \in \Omega_\theta} L_{k,\text{rep}}^{t}(\theta)
    \end{split}
\end{equation}

Consequently, we define the error upper bound function $\mathcal{E}_{k, \text{bound}}^{t}\left(\{\eta_k^{t,j}\}_{j=0}^{J-1}, \|\theta_k^{t,0} - \theta_k^{t,*}\|_2^2\right)$ as:
\begin{equation}
    \begin{split}
        \mathcal{E}_{k, \text{bound}}^{t}\left(\{\eta_k^{t,j}\}_{j=0}^{J-1}, \|\theta_k^{t,0} - \theta_k^{t,*}\|_2^2\right)
        =& \frac{L}{2} \mathrm{exp}\left(- (\mu - 2 L_2 L_{\mathcal{Q}}) \sum_{j=0}^{J-1} \eta_k^{t,j}\right)\mathbb{E}_{\xi}[\|\theta_k^{t,0} - \theta_k^{t,*}\|_2^2]
        \\
        &+ \frac{L}{2} \sum_{j=0}^{J-1} \left(\eta_k^{t,j} \frac{{L_2}^2 {\epsilon_{\text{approx}}}^2}{\mu} + (\eta_k^{t,j})^2 G^2\right)\prod_{i=j+1}^{J-1}(1 - \eta_k^{t,i}(\mu - 2 L_2 L_{\mathcal{Q}}))
        \\
        &+ (L_{\mathcal{Q}}D_\theta + \epsilon_{\text{approx}}) \cdot \sup_{\theta \in \Omega_\theta} L_{k,\text{rep}}^{t}(\theta)
    \end{split}
\end{equation}

Finally, we obtain:
\begin{equation}
    \mathbb{E}\left[L_{k,\text{total}}^{t}(\theta_k^{t,J}|\lambda_k^{t,J})\right] 
    \leq L_{k,\text{total}}^{t}(\theta_k^{t,*}|\lambda_k^{t,*}) + \mathcal{E}_{k, \text{bound}}^{t}\left(\{\eta_k^{t,j}\}_{j=0}^{J-1}, \|\theta_k^{t,0} - \theta_k^{t,*}\|_2^2\right)
\end{equation}

This completes the proof of Theorem 1.

\newpage
\subsection{Proof of Theorem 2}
\label{Appendix B.4}
\setcounter{theorem}{1}   
\begin{theorem}[Spectral Efficiency of the Holistic Buffer Maintenance Strategy (HBMS)]
  Given $M > 2C_{\max}$ and standard assumptions (see \hyperref[Appendix B.1]{Appendix B.1}), 
  let $C_{\mathrm{rand}} \triangleq 1$ be the spectral efficiency constant of class-wise random sampling, and let $C_\kappa \triangleq \inf_{\theta \in \Omega_\theta, \hat{g} \in \mathcal{G}} \exp(-\beta(\theta, \hat{g}) \Delta \rho(\theta, \hat{g}))$ (depending only on $M$ and $C_{\max}$) be that of the Holistic Buffer Maintenance Strategy, where $\Delta \rho(\theta, \hat{g}) \triangleq \rho_{\mathrm{rand}} - \rho_{\mathrm{HBMS}}$ is the spectral radius gap between the two methods.
  Then, $0 < C_\kappa < C_{\mathrm{rand}}$ holds for all $k \in \mathcal{K}$ and $t \in [T]$.
\end{theorem}

\textbf{Proof}:

To establish a baseline for spectral efficiency comparison, we define the normalized spectral efficiency constant of class-wise random sampling as $C_{\mathrm{rand}} \triangleq 1$.

Given that the kernel decay rate scales as $\beta \propto 1/d$, and the dimension $d$ is intrinsically tied to the model parameters $\theta$, we define the kernel decay rate as a continuous function $\beta(\theta, \hat{g})$ over the model parameters $\theta$ and the normalized logits vector $\hat{g}$, thereby satisfying the prerequisites for the Continuous Mapping Theorem and unifying the analysis domain of kernel spectral geometry. By Assumption 2, the model parameter space $\Omega_\theta$ is a compact set; by Assumption 4, the range of the normalized logits vector $\hat{g}$, denoted as $\mathcal{G} \subseteq \mathbb{R}^d$, is also compact. Consequently, their Cartesian product $\Omega_\theta \times \mathcal{G}$ remains compact under the product topology, and $\beta(\theta, \hat{g})$ is a continuous real-valued function defined on this compact set. Furthermore, by the Weierstrass Extreme Value Theorem, the image set $\text{Im}(\beta) \triangleq \{\beta(\theta, \hat{g}) \mid (\theta, \hat{g}) \in \Omega_\theta \times \mathcal{G}\}$ is a bounded closed interval in $\mathbb{R}^+$. Thus, there exist a strictly positive lower bound $\beta_{\min} > 0$ and a finite upper bound $\beta_{\max} < \infty$ such that:
\begin{equation}
    \beta_{\min} \triangleq \min_{(\theta, \hat{g}) \in \Omega_\theta \times \mathcal{G}} \beta(\theta, \hat{g}), \quad \beta_{\max} \triangleq \max_{(\theta, \hat{g}) \in \Omega_\theta \times \mathcal{G}} \beta(\theta, \hat{g}).
\end{equation}

That is, the kernel decay rate $\beta$ satisfies:
\begin{equation}
    \beta \in [\beta_{\min}, \beta_{\max}]
\end{equation}

By Assumption 7, the number of classes satisfies $|\mathcal{C}_k^{\leq t}| \leq C_{\max}$. Under the conditions that $|\mathcal{C}_k^{\leq t}| \geq 2$ and each class contains at least 2 samples, the number of samples $m_{k,c}^t$ for class $c$ in the buffer satisfies:
\begin{equation}
2 \leq m_{k,c}^t = \lfloor M/|\mathcal{C}_k^{\leq t}| \rfloor \leq M/2.
\end{equation}

Since the range $\mathcal{G}$ of the normalized logits vector $\hat{g}$ is a compact set, we define $r_{\mathrm{opt}} \triangleq \sup_{S \subset \mathcal{G}, |S|=m_{k,c}^t} \min_{x_i \neq x_j \in S} \|\hat{g}(x_i) - \hat{g}(x_j)\|_2$ as the theoretical optimal max-min distance when placing $m_{k,c}^t$ points in $\mathcal{G}$. Let the $d$-dimensional Lebesgue measure (volume) of $\mathcal{G}$ be $\mathrm{Vol}(\mathcal{G})$, and the volume of the unit ball in the $d$-dimensional Euclidean space be $V_d = \pi^{d/2}/\Gamma(d/2+1)$. On one hand, since the sum of the volumes of $m_{k,c}^t$ disjoint balls with radius $r_{\mathrm{opt}}/2$ cannot exceed the volume of the container $\mathcal{G}$, we can derive the upper bound $\delta_{\mathrm{ub}}$ for the theoretical optimal max-min distance $r_{\mathrm{opt}}$. On the other hand, when placing points sequentially in the compact set $\mathcal{G}$ based on a greedy strategy, if the total volume of the repulsion balls corresponding to the first $m_{k,c}^t-1$ points does not exceed the volume of $\mathcal{G}$, there must exist a point configuration such that the minimum pairwise distance satisfies the theoretical lower bound $\delta_{\mathrm{lb}}$. That is:
\begin{equation}
    m_{k,c}^t V_d \left(\frac{\delta_{\mathrm{ub}}}{2}\right)^d = \mathrm{Vol}(\mathcal{G}) 
    \quad
    \Rightarrow
    \quad
    \delta_{\mathrm{ub}} = 2 \left( \frac{\mathrm{Vol}(\mathcal{G})}{m_{k,c}^t V_d} \right)^{1/d}
\end{equation}
\begin{equation}
    (m_{k,c}^t - 1) V_d \left(\frac{\delta_{\mathrm{lb}}}{2}\right)^d = \mathrm{Vol}(\mathcal{G}) 
    \quad
    \Rightarrow
    \quad
    \delta_{\mathrm{lb}} = 2 \left( \frac{\mathrm{Vol}(\mathcal{G})}{(m_{k,c}^t - 1) V_d} \right)^{1/d}
\end{equation}

By definition, the theoretical optimal max-min distance $r_{\mathrm{opt}}$ satisfies $\delta_{\mathrm{ub}} \geq r_{\mathrm{opt}} \geq \delta_{\mathrm{lb}}$.

Next, consider the Holistic Buffer Maintenance Strategy proposed in our paper:

In the first stage, for each known class $c$, the algorithm selects the $\min\{2|\mathcal{M}_{k,c}^t|, |X_c|\}$ samples with the highest IDV values from the candidate pool $X_c$ to construct the buffer subset $\tilde{\mathcal{M}}^t_{k,c}$:
\begin{equation}
    \text{IDV}(x | c_x) = -\log \bar{p}_{\mathcal{M}_k^{t-1}}(c_x | x) + \lambda_1 \cdot \min_{x_i \in \mathcal{M}_k^{t-1}} \|\hat{g}(x)-\hat{g}(x_i)\|_2^2
\end{equation}

Since the information term does not contribute to the optimization of the kernel condition number, maximizing the IDV value of a sample $x$ is equivalent to maximizing the distance term $\min_{x_i \in \mathcal{M}_k^{t-1}} \|\hat{g}(x)-\hat{g}(x_i)\|_2^2$. According to the theoretical guarantee of the greedy max-min distance algorithm, in the metric space $(\mathcal{G}, \|\cdot\|_2)$, the minimum distance $\delta_{\min}^{\mathrm{IDV}}$ obtained by this algorithm satisfies $\delta_{\min}^{\mathrm{IDV}} \geq r_{\mathrm{opt}}/2$. Consequently, we define the max-min distance approximation ratio as $\nu_{\mathrm{pack}} \triangleq \delta_{\min}^{\mathrm{IDV}} / r_{\mathrm{opt}}$. It is evident that $\nu_{\mathrm{pack}} \leq 1$, and it satisfies:
\begin{equation}
    \nu_{\mathrm{pack}} \triangleq \frac{\delta_{\min}^{\mathrm{IDV}}}{r_{\mathrm{opt}}} > \frac{r_{\mathrm{opt}} / 2}{r_{\mathrm{opt}}} = \frac{1}{2}
\end{equation}

That is, the max-min distance approximation ratio $\nu_{\mathrm{pack}} \in [0.5, 1]$.

In the second stage, for each known class $c$, the algorithm selects the $|\mathcal{M}_{k,c}^t|$ samples with the highest CDV values from the candidate subset $\tilde{\mathcal{M}}^t_{k,c}$ to construct the final buffer subset $\mathcal{M}_{k,c}^t$ for class $c$:
\begin{equation}
    \text{CDV}(x | c_x) = \log \frac{p(c_x | x)}{\bar{p}_{\mathcal{M}_k^{t-1} \cup \{x\}}(c_x \mid x)} + \lambda_2 \cdot \min_{x_i \in \mathcal{M}_k^{t-1}} \|\hat{g}(x)-\hat{g}(x_i)\|_2^2
\end{equation}

Obviously, the CDV selection operation can strictly and monotonically optimize the minimum pairwise distance of the point set $\min_{i \neq j,\, x_i, x_j \in \mathcal{M}_{k}^{t}} \| \hat{g}(x_i) - \hat{g}(x_j) \|_2$, thereby effectively retaining samples with high dispersion. Consequently, the minimum pairwise distance of the final point set in the buffer maintenance algorithm, defined as $\delta_{\min}^{\mathrm{KS}} \triangleq \min_{i \neq j,\, x_i, x_j \in \mathcal{M}_{k}^{t}} \| \hat{g}(x_i) - \hat{g}(x_j) \|_2$, inevitably satisfies $\delta_{\min}^{\mathrm{KS}} \geq \delta_{\min}^{\mathrm{IDV}}$. That is, the minimum pairwise Euclidean distance $\delta_{\min}^{\mathrm{KS}}$ obtained by the two-stage greedy selection algorithm satisfies:
\begin{equation}
    \delta_{\min}^{\mathrm{KS}} 
    \geq \delta_{\min}^{\mathrm{IDV}} 
    \geq \nu_{\mathrm{pack}} \cdot r_{\mathrm{opt}} 
    \geq \nu_{\mathrm{pack}} \cdot \delta_{\mathrm{lb}} 
    = 2 \nu_{\mathrm{pack}} \left( \frac{\mathrm{Vol}(\mathcal{G})}{(m_{k,c}^t - 1) V_d} \right)^{1/d}
\end{equation}

where the max-min distance approximation ratio $\nu_{\mathrm{pack}} \in [0.5, 1]$.

Next, consider the class-wise random sampling algorithm.

For class $c$, consider independently and uniformly sampling $m_{k,c}^t$ points in the $d$-dimensional compact set $\mathcal{G}$, and define the minimum Euclidean distance among all pairs of points as $\delta_{\min}^{\mathrm{rand}} \triangleq \min_{1 \leq i < j \leq m_{k,c}^t} \|x_i - x_j\|_2$. For sufficiently small $r$, the probability that the distance between any pair of points is greater than $r$, i.e., the probability of the event $\{\delta_{\min}^{\mathrm{rand}} > r\}$, is given by:
\begin{equation}
    \mathbb{P}(\delta_{\min}^{\mathrm{rand}} > r) = \prod_{i=1}^{m_{k,c}^t-1} \mathbb{P}\left( \|x_{i+1} - x_j\|_2 > r,\ \forall j \leq i \ \bigg|\ x_1, \dots, x_i \right)
\end{equation}

Let $B(x,r)$ denote the closed ball centered at $x$ with radius $r$. The above equation implies that, given the positions of the first $i$ points, the $(i+1)$-th point must fall within the region $\mathcal{G} \setminus \bigcup_{j=1}^i B(x_j, r)$. The volume of this feasible region is $\mathrm{Vol}_{\mathrm{feasible}}^{i+1} = \mathrm{Vol}(\mathcal{G}) - \mathrm{Vol}\left( \left(\bigcup_{j=1}^i B(x_j, r)\right) \cap \mathcal{G} \right)$. Furthermore, when $r \ll \mathrm{diam}(\mathcal{G})$, the overlapping volumes among the balls and the boundary truncation effects can be neglected. Combined with the volume of the unit ball in the $d$-dimensional Euclidean space, $V_d = \pi^{d/2}/\Gamma(d/2+1)$, we obtain $\mathrm{Vol}_{\mathrm{feasible}}^{i+1} \approx \mathrm{Vol}(\mathcal{G}) - \sum_{j=1}^i \mathrm{Vol}(B(x_j, r)) = \mathrm{Vol}(\mathcal{G}) - i \cdot V_d r^d$. Consequently, the conditional probability satisfies:
\begin{equation}
    \mathbb{P}\left( \|x_{i+1} - x_j\|_2 > r,\ \forall j \leq i \ \bigg|\ x_1, \dots, x_i \right)
    \approx
    \frac{\mathrm{Vol}(\mathcal{G}) - i \cdot V_d r^d}{\mathrm{Vol}(\mathcal{G})}
    = 1 - \frac{i \cdot V_d r^d}{\mathrm{Vol}(\mathcal{G})}
\end{equation}

Substituting this into the original expression and applying the inequality $1-x \leq \exp(-x)$, we obtain:
\begin{equation}
    \mathbb{P}(\delta_{\min}^{\mathrm{rand}} > r) \approx \prod_{i=1}^{m_{k,c}^t-1} \left(1 - \frac{i \cdot V_d r^d}{\mathrm{Vol}(\mathcal{G})}\right)
    \leq \prod_{i=1}^{m_{k,c}^t-1} \mathrm{exp}\left(- \frac{i \cdot V_d r^d}{\mathrm{Vol}(\mathcal{G})}\right)
    = \mathrm{exp}\left(- \frac{V_d r^d}{\mathrm{Vol}(\mathcal{G})} \sum_{i=1}^{m_{k,c}^t-1} i\right)
\end{equation}

Substituting the arithmetic series sum formula $\sum_{i=1}^{m_{k,c}^t-1} i = \frac{m_{k,c}^t(m_{k,c}^t-1)}{2}$, we obtain:
\begin{equation}
    \mathbb{P}(\delta_{\min}^{\mathrm{rand}} > r) 
    \leq \mathrm{exp}\left(- \frac{m_{k,c}^t(m_{k,c}^t-1) V_d r^d}{2 \mathrm{Vol}(\mathcal{G})}\right) 
\end{equation}

Since the expectation of the non-negative random variable $\delta_{\min}^{\mathrm{rand}}$ can be expressed as the integral of its survival function $\int_{0}^{\infty} \mathbb{P}(\delta_{\min}^{\mathrm{rand}} > r) \mathrm{d}r$, we obtain:
\begin{equation}
    \mathbb{E}[\delta_{\min}^{\mathrm{rand}}] \leq \int_{0}^{\infty} \exp\left( - \frac{m_{k,c}^t(m_{k,c}^t-1) V_d}{2\mathrm{Vol}(\mathcal{G})} r^d \right) \mathrm{d}r
\end{equation}

Let $u = \frac{m_{k,c}^t(m_{k,c}^t-1) V_d}{2\mathrm{Vol}(\mathcal{G})} r^d$. Then $r = \left( \frac{2\mathrm{Vol}(\mathcal{G})}{m_{k,c}^t(m_{k,c}^t-1) V_d} \right)^{1/d} u^{1/d}$ and $\mathrm{d}r = \frac{1}{d} \left( \frac{2\mathrm{Vol}(\mathcal{G})}{m_{k,c}^t(m_{k,c}^t-1) V_d} \right)^{1/d} u^{\frac{1}{d}-1} \mathrm{d}u$. Substituting these into the above expression, we obtain:
\begin{equation}
    \begin{split}
        \mathbb{E}[\delta_{\min}^{\mathrm{rand}}] 
        \leq& \int_{0}^{\infty} \exp\left( - u \right) \frac{1}{d} \left( \frac{2\mathrm{Vol}(\mathcal{G})}{m_{k,c}^t(m_{k,c}^t-1) V_d} \right)^{1/d} u^{\frac{1}{d}-1} \mathrm{d}u
        \\
        =& \left( \frac{2\mathrm{Vol}(\mathcal{G})}{m_{k,c}^t(m_{k,c}^t-1) V_d} \right)^{1/d} \int_{0}^{\infty} \frac{1}{d} \exp\left( - u \right) u^{\frac{1}{d}-1} \mathrm{d}u
    \end{split}
\end{equation}

Given the Gamma function $\Gamma(z) \triangleq \int_{0}^{\infty} e^{-t} t^{z-1} \mathrm{d}t$ and its recurrence identity $\Gamma(z+1) = z\Gamma(z)$, substituting $z = 1/d$ yields:
\begin{equation}
    \mathbb{E}[\delta_{\min}^{\mathrm{rand}}] \leq \left( \frac{2\mathrm{Vol}(\mathcal{G})}{m_{k,c}^t(m_{k,c}^t-1) V_d} \right)^{1/d} \frac{1}{d}\Gamma\left(\frac{1}{d}\right) = \Gamma\left(1 + \frac{1}{d}\right)  \left( \frac{2\mathrm{Vol}(\mathcal{G})}{m_{k,c}^t(m_{k,c}^t-1) V_d} \right)^{1/d}
\end{equation}

From the above, we have derived $\delta_{\min}^{\mathrm{KS}} \geq 2 \nu_{\mathrm{pack}} \left( \frac{\mathrm{Vol}(\mathcal{G})}{(m_{k,c}^t - 1) V_d} \right)^{1/d}$. Taking the ratio of the minimum pairwise distances obtained by the two sampling schemes, we obtain:
\begin{equation}
    \begin{split}
        \frac{\delta_{\min}^{\mathrm{KS}}}{\mathbb{E}[\delta_{\min}^{\mathrm{rand}}]} 
        \geq& \frac{2 \nu_{\mathrm{pack}} \left( \frac{\mathrm{Vol}(\mathcal{G})}{(m_{k,c}^t - 1) V_d} \right)^{1/d}}{\Gamma\left(1+\frac{1}{d}\right) \left( \frac{2\mathrm{Vol}(\mathcal{G})}{m_{k,c}^t(m_{k,c}^t-1) V_d} \right)^{1/d}}
        = \frac{2 \nu_{\mathrm{pack}}}{\Gamma\left(1+\frac{1}{d}\right)} \cdot \left( \frac{\mathrm{Vol}(\mathcal{G})}{(m_{k,c}^t - 1) V_d} \cdot \frac{m_{k,c}^t(m_{k,c}^t-1) V_d}{2\mathrm{Vol}(\mathcal{G})} \right)^{1/d}
        \\
        =& \frac{2 \nu_{\mathrm{pack}}}{\Gamma\left(1+\frac{1}{d}\right)} \cdot \left( \frac{m_{k,c}^t}{2} \right)^{1/d} 
        = \frac{2^{(1 - 1/d)} \nu_{\mathrm{pack}}}{\Gamma\left(1+\frac{1}{d}\right)} \cdot \left(m_{k,c}^t\right)^{1/d}
    \end{split}
\end{equation}

Substituting $m_{k,c}^t \geq 2$ and $\nu_{\mathrm{pack}} \in [0.5, 1]$, we obtain:
\begin{equation}
    \begin{split}
        \frac{\delta_{\min}^{\mathrm{KS}}}{\mathbb{E}[\delta_{\min}^{\mathrm{rand}}]} 
        \geq \frac{2^{(1 - 1/d)} \nu_{\mathrm{pack}}}{\Gamma\left(1+\frac{1}{d}\right)} \cdot \left(m_{k,c}^t\right)^{1/d}
        \geq \frac{2^{(1 - 1/d)} \nu_{\mathrm{pack}}}{\Gamma\left(1+\frac{1}{d}\right)} \cdot 2^{1/d}
        = \frac{2 \nu_{\mathrm{pack}}}{\Gamma\left(1+\frac{1}{d}\right)}
        > \frac{2 \cdot 0.5}{\Gamma\left(1+\frac{1}{d}\right)} 
        = \frac{1}{\Gamma\left(1+\frac{1}{d}\right)} 
    \end{split}
\end{equation}

Since $d \geq 2$, we have $1 + 1/d \in (1, 1.5] \subset (1, 2)$. It is known that $\Gamma(x) < 1$ for all $x \in (1, 2)$ (i.e., $\Gamma(x) < \Gamma(1) = 1$), and $\Gamma(1.5) = \sqrt{\pi}/2$. Therefore, we have $\Gamma(1 + 1/d) < 1$. Substituting this into the above expression yields:
\begin{equation}
    \begin{split}
        \frac{\delta_{\min}^{\mathrm{KS}}}{\mathbb{E}[\delta_{\min}^{\mathrm{rand}}]} 
        > \frac{1}{\Gamma\left(1+\frac{1}{d}\right)} > \frac{1}{1} = 1
    \end{split}
\end{equation}

Furthermore, consider the Holistic Buffer Maintenance Strategy proposed in our paper. For class $c$, this strategy constructs an $m_{k,c}^t \times m_{k,c}^t$ Gaussian kernel matrix $\mathbf{K}^{\mathrm{KS}}$, whose entries are defined as follows: for any pair of samples $(x_i, x_j)$ (where $i,j \in \{1,\dots, m_{k,c}^t\}$ and $i \neq j$), $K(x_i, x_j) = \exp\left(-\beta \|\hat{g}(x_i) - \hat{g}(x_j)\|_2^2\right)$. Since the pairwise distance satisfies $\|\hat{g}(x_i) - \hat{g}(x_j)\|_2 \geq \delta_{\min}^{\mathrm{KS}}$, which implies that the squared distance $\|\hat{g}(x_i) - \hat{g}(x_j)\|_2^2 \geq (\delta_{\min}^{\mathrm{KS}})^2$, any off-diagonal entry is bounded above by $K(x_i, x_j) \leq \exp\left(-\beta (\delta_{\min}^{\mathrm{KS}})^2\right)$. Summing this upper bound over all off-diagonal entries yields:
\begin{equation}
    \sum_{j=1, j \neq i}^{m_{k,c}^t} |K(x_i, x_j)| 
    \leq \sum_{j=1, j \neq i}^{m_{k,c}^t} \exp\left(-\beta \left(\delta_{\min}^{\mathrm{KS}}\right)^2\right) 
    = (m_{k,c}^t - 1) \exp\left(-\beta \left(\delta_{\min}^{\mathrm{KS}}\right)^2\right)
\end{equation}

The kernel matrix $\mathbf{K}^{\mathrm{KS}}$ is a real symmetric matrix, and by the definition of the normalized Gaussian kernel, all its diagonal entries satisfy $K(x_i, x_i) = 1$. According to the Gershgorin Circle Theorem, any eigenvalue $\lambda$ of $\mathbf{K}^{\mathrm{KS}}$ must lie within a closed disk centered at the diagonal entry $1$ with a radius equal to the sum of the absolute values of the off-diagonal entries in the corresponding row, $R_i = \sum_{j \neq i} |K(x_i, x_j)|$. Defining the row-sum upper bound of the spectral radius as $\rho_{\mathrm{KS}}^{\mathrm{up}} \triangleq \max_{i} R_i$, which is the maximum of the row sums of the absolute values of the off-diagonal entries, we obtain:
\begin{equation}
    \rho_{\mathrm{KS}}^{\mathrm{up}} = \max_{i} \sum_{j=1, j \neq i}^{m_{k,c}^t} |K(x_i, x_j)| \leq (m_{k,c}^t - 1) \exp\left(-\beta \left(\delta_{\min}^{\mathrm{KS}}\right)^2\right)
\end{equation}

Since all eigenvalues of $\mathbf{K}^{\mathrm{KS}}$ lie within the closed interval $\left[1 - \rho_{\mathrm{KS}}^{\mathrm{up}}, \; 1 + \rho_{\mathrm{KS}}^{\mathrm{up}}\right]$, its spectral condition number $\kappa_{\mathrm{KS}} \triangleq \lambda_{\max}(\mathbf{K}^{\mathrm{KS}}) / \lambda_{\min}(\mathbf{K}^{\mathrm{KS}})$ satisfies:
\begin{equation}
    \kappa_{\mathrm{KS}} \leq \frac{1 + \rho_{\mathrm{KS}}^{\mathrm{up}}}{1 - \rho_{\mathrm{KS}}^{\mathrm{up}}}
\end{equation}

Similarly, for the class-wise random sampling method, adopting the kernel function $K(x_i, x_j) = \exp\left(-\beta \|\hat{g}(x_i) - \hat{g}(x_j)\|_2^2\right)$, we define the proxy lower bound for the expected spectral radius of its Gaussian kernel matrix $\mathbf{K}^{\mathrm{rand}}$ as $\rho_{\mathrm{rand}}^{\mathrm{low}}$:
\begin{equation}
    \rho_{\mathrm{rand}}^{\mathrm{low}} \triangleq (m_{k,c}^t-1) \exp\left(-\beta (\mathbb{E}[\delta_{\min}^{\mathrm{rand}}])^2\right).
\end{equation}

Since $\rho_{\mathrm{rand}}^{\mathrm{low}}$ is a lower bound of $\rho_{\mathrm{rand}}$, the spectral condition number $\kappa_{\mathrm{rand}} \triangleq \lambda_{\max}(\mathbf{K}^{\mathrm{rand}}) / \lambda_{\min}(\mathbf{K}^{\mathrm{rand}})$ of $\mathbf{K}^{\mathrm{rand}}$ satisfies:
\begin{equation}
    \kappa_{\mathrm{rand}} \geq \frac{1 + \rho_{\mathrm{rand}}^{\mathrm{low}}}{1 - \rho_{\mathrm{rand}}^{\mathrm{low}}}
\end{equation}

Thus, a conservative upper bound on the ratio of the spectral condition numbers $\kappa_{\mathrm{KS}}/\kappa_{\mathrm{rand}}$ is given by:
\begin{equation}
    \frac{\kappa_{\mathrm{KS}}}{\kappa_{\mathrm{rand}}} 
    \leq \frac{1+\rho_{\mathrm{KS}}^{\mathrm{up}}}{1-\rho_{\mathrm{KS}}^{\mathrm{up}}} \cdot \frac{1-\rho_{\mathrm{rand}}^{\mathrm{low}}}{1+\rho_{\mathrm{rand}}^{\mathrm{low}}}
\end{equation}

Define the function $f(\rho) \triangleq \ln\frac{1+\rho}{1-\rho}$ with the domain $\rho \in [0, 1)$. Since $f'(\rho) = \frac{2}{(1-\rho)^2} > 0$, $f(\rho)$ is strictly increasing and continuously differentiable on its domain. Taking the natural logarithm of both sides of the above inequality, and letting $\xi$ lie between $\rho_{\mathrm{KS}}^{\mathrm{up}}$ and $\rho_{\mathrm{rand}}^{\mathrm{low}}$, we obtain by the Lagrange Mean Value Theorem:
\begin{equation}
    \ln\left(\frac{\kappa_{\mathrm{KS}}}{\kappa_{\mathrm{rand}}}\right) 
    \leq f(\rho_{\mathrm{KS}}^{\mathrm{up}}) - f(\rho_{\mathrm{rand}}^{\mathrm{low}}) 
    = f'(\xi) \left(\rho_{\mathrm{KS}}^{\mathrm{up}} - \rho_{\mathrm{rand}}^{\mathrm{low}}\right)
\end{equation}

Defining the boundary difference $\Delta \rho^{\mathrm{bound}} \triangleq \rho_{\mathrm{rand}}^{\mathrm{low}} - \rho_{\mathrm{KS}}^{\mathrm{up}}$, we obtain:
\begin{equation}
    \begin{split}
        \Delta \rho^{\mathrm{bound}} 
        \geq& (m_{k,c}^t-1) \exp\left(-\beta (\mathbb{E}[\delta_{\min}^{\mathrm{rand}}])^2\right) - (m_{k,c}^t-1) \exp\left(-\beta (\delta_{\min}^{\mathrm{KS}})^2\right)
        \\
        =& (m_{k,c}^t-1) \exp\left(-\beta (\mathbb{E}[\delta_{\min}^{\mathrm{rand}}])^2\right) \left(1 - 
        \frac{\exp\left(-\beta (\delta_{\min}^{\mathrm{KS}})^2\right)}{\exp\left(-\beta (\mathbb{E}[\delta_{\min}^{\mathrm{rand}}])^2\right)}\right)
        \\
        =& (m_{k,c}^t-1) \exp\left(-\beta (\mathbb{E}[\delta_{\min}^{\mathrm{rand}}])^2\right) \left(1 - \exp\left(-\beta [(\delta_{\min}^{\mathrm{KS}})^2 - (\mathbb{E}[\delta_{\min}^{\mathrm{rand}}])^2]\right)\right).
        \\
        =& (m_{k,c}^t-1) \exp\left(-\beta (\mathbb{E}[\delta_{\min}^{\mathrm{rand}}])^2\right) \left(1 - \exp\left(-\beta \left[\left(\frac{(\delta_{\min}^{\mathrm{KS}})}{\mathbb{E}[\delta_{\min}^{\mathrm{rand}}]}\right)^2 
        - 1\right] (\mathbb{E}[\delta_{\min}^{\mathrm{rand}}])^2\right)\right)
        \\
        >& (m_{k,c}^t-1) \exp\left(-\beta (\mathbb{E}[\delta_{\min}^{\mathrm{rand}}])^2\right) \left(1 - \exp\left(-\beta(\mathbb{E}[\delta_{\min}^{\mathrm{rand}}])^2\right)\right)
    \end{split}
\end{equation}

Note that $\Delta \rho^{\mathrm{bound}}(\beta, m_{k,c}^t, \delta)$ is a multivariate continuous elementary function with respect to $\beta, m_{k,c}^t,$ and $\delta$. Since the range $\mathcal{G} \subseteq \mathbb{R}^d$ of the normalized logits vector $\hat{g}$ is a compact set, there exists a maximum distance in the feature space (i.e., the diameter) $\overline{\delta} = \mathrm{diam}(\mathcal{G}) < \infty$. Given that $m_{k,c}^t \leq M/2$, we have $\delta_{\min}^{\mathrm{KS}} \geq 2 \nu_{\mathrm{pack}} \left(\frac{\mathrm{Vol}(\mathcal{G})}{(M/2) V_d}\right)^{1/d} > \nu_{\mathrm{pack}}\left(\frac{\mathrm{Vol}(\mathcal{G})}{M V_d}\right)^{1/d}$. Defining the conservative lower bound of the minimum pairwise distance as $\underline{\delta} \triangleq \nu_{\mathrm{pack}}\left(\frac{\mathrm{Vol}(\mathcal{G})}{M V_d}\right)^{1/d}$, the parameter domain $\mathcal{V}^{\mathrm{bound}}$ of $\Delta \rho^{\mathrm{bound}}(\beta, m_{k,c}^t, \delta)$ is given by:
\begin{equation}
    \mathcal{V}^{\mathrm{bound}} \triangleq [\beta_{\min}, \beta_{\max}] \times [2, M/2] \times [\underline{\delta}, \overline{\delta}]
\end{equation}

That is, $\mathcal{V}^{\mathrm{bound}}$ forms a compact hyperrectangle in $\mathbb{R}^3$. By the Weierstrass Extreme Value Theorem, the continuous function $\Delta \rho^{\mathrm{bound}}$ attains its minimum $\Delta \rho_{\min}^{\mathrm{bound}}$ on this compact set, which satisfies:
\begin{equation}
    \Delta \rho^{\mathrm{bound}} \geq \Delta \rho_{\min}^{\mathrm{bound}} \triangleq \inf_{\mathcal{V}^{\mathrm{bound}}} \Delta \rho^{\mathrm{bound}} > 0
\end{equation}

Thus, we have $\Delta \rho^{\mathrm{bound}} > 0$. Combined with $f'(\xi) > 0$, we can derive the upper bound on the ratio of the spectral condition numbers:
\begin{equation}
    \frac{\kappa_{\mathrm{KS}}}{\kappa_{\mathrm{rand}}} 
    \leq \exp\left(f(\rho_{\mathrm{KS}}^{\mathrm{up}}) - f(\rho_{\mathrm{rand}}^{\mathrm{low}})\right) 
    = \exp\left(f'(\xi) \left(\rho_{\mathrm{KS}}^{\mathrm{up}} - \rho_{\mathrm{rand}}^{\mathrm{low}}\right)\right) 
    = \exp\left(- f'(\xi) \Delta \rho^{\mathrm{bound}}\right) 
    < 1
\end{equation}

Furthermore, for notational consistency, let $\rho_{\mathrm{HBMS}} \triangleq \rho_{\mathrm{KS}}$. Based on the previous derivations, we have $\rho_{\mathrm{HBMS}} \leq \rho_{\mathrm{KS}}^{\mathrm{up}}$ and $\rho_{\mathrm{rand}} \geq \rho_{\mathrm{rand}}^{\mathrm{low}}$. Defining the true spectral radius difference as $\Delta \rho(\theta, \hat{g}) \triangleq \rho_{\mathrm{rand}} - \rho_{\mathrm{HBMS}}$, we obtain $\Delta \rho(\theta, \hat{g}) \geq \rho_{\mathrm{rand}}^{\mathrm{low}} - \rho_{\mathrm{KS}}^{\mathrm{up}} = \Delta \rho^{\mathrm{bound}}$. Combined with the previous conclusion that $\Delta \rho^{\mathrm{bound}} \geq \Delta \rho_{\min}^{\mathrm{bound}} > 0$, it follows that the global infimum of the true spectral radius difference, $\Delta \rho_{\min} \triangleq \inf_{\theta \in \Omega_\theta, \hat{g} \in \mathcal{G}} \Delta \rho(\theta, \hat{g})$, satisfies:
\begin{equation}
    \Delta \rho(\theta, \hat{g}) \geq \rho_{\mathrm{rand}}^{\mathrm{low}} - \rho_{\mathrm{KS}}^{\mathrm{up}} = \Delta \rho^{\mathrm{bound}} \geq \Delta \rho_{\min}^{\mathrm{bound}} > 0
\end{equation}

Consequently, the composite function $\beta(\theta, \hat{g}) \Delta \rho(\theta, \hat{g})$ has a strictly positive lower bound on the compact set $\Omega_\theta \times \mathcal{G}$. Defining the global spectral efficiency constant of the Holistic Buffer Maintenance Strategy as $C_\kappa \triangleq \inf_{\theta \in \Omega_\theta, \hat{g} \in \mathcal{G}} \exp\left(-\beta(\theta, \hat{g}) \Delta \rho(\theta, \hat{g})\right)$, it satisfies:
\begin{equation}
C_\kappa \leq \exp\left(-\beta_{\min} \Delta \rho_{\min}\right) < 1 = C_{\mathrm{rand}}.
\end{equation}

This result applies to the RKHS generalization error term, demonstrating that our method can reduce the generalization error upper bound induced by feature space complexity by a deterministic factor. This completes the proof of Theorem 2.

\newpage
\subsection{Proof of Lemma 2}
\label{Appendix B.5}
\setcounter{theorem}{1}   
\begin{lemma}[Recursive Regret Bound for Local Training in FedKACE]
    Under standard assumptions (see \hyperref[Appendix B.1]{Appendix B.1}) and a buffer capacity $M \geq 2C_{\max}$, consider the $t$-th FL round. 
    Let $n_{k}^t$ be the local data size for client $k$ at FL round $t$. Let $f^{t,*}_{k,\mathrm{cum}}$ be the optimal model under the cumulative distribution $p_k^{\leq t}$, parameterized by $\theta^{t,*}_{k,\mathrm{cum}} \triangleq \arg\min_{\theta \in \Omega_\theta} \mathbb{E}_{p_k^{\leq t}}[\mathcal{R}(f_\theta)]$. 
    Meanwhile, define the regret of model $\theta$ on client $k$ at FL round $t$ as $\mathrm{Regret}_k^{t}(\theta) \triangleq \mathbb{E}_{p_k^{\leq t}}\left[\mathcal{R}(f_{\theta}) - \mathcal{R}(f_{k,\mathrm{cum}}^{t,*})\right]$, and the local training regret functional $\mathrm{Regret}_k^{\mathrm{local}}(t; \theta_k^{t,0}, \{\eta_k^{t,j}\}_{j=0}^{J-1}) \triangleq \mathrm{Regret}_k^{t}(\theta_k^{t,J})$, where $\theta_k^{t,0}$ and $\{\eta_k^{t,j}\}_{j=0}^{J-1}$ are the initial parameter and learning rate sequence, respectively. This functional satisfies the following recursive upper bound:
    \begin{equation}
        \begin{split}
            \mathrm{Regret}_k^{\mathrm{local}}(t; \theta_k^{t,0}, \{\eta_k^{t,j}\}_{j=0}^{J-1})
            \leq& \left(1 - \omega_{k}^{t}\right) \mathrm{Regret}_k^{\mathrm{local}}(t-1; \theta_k^{t-1,0}, \{\eta_k^{t-1,j}\}_{j=0}^{J-1})
            + \omega_{k}^{t} \cdot U_k(t; \theta_k^{t,0}, \{\eta_k^{t,j}\}_{j=0}^{J-1})
            \\
            &+ \left(1 - \omega_{k}^{t}\right) \mathbb{E}_{p_k^{\leq t-1}}\left[ \mathcal{R}(f_{\theta_k^{t,J}}) - \mathcal{R}(f_{\theta_k^{t-1,J}})\right] 
        \end{split}
    \end{equation}
    where $\omega_{k}^{t} \triangleq n_{k}^t / \sum_{\tau=1}^t n_{k}^\tau$ is the weight of $p_k^{t}$ in $p_k^{\leq t}$, and the optimization error term $U_k(t; \theta_k^{t,0}, \{\eta_k^{t,j}\}_{j=0}^{J-1})$ is defined as:
    \begin{equation}
        \begin{split}
            U_k(t; \theta_k^{t,0}, \{\eta_k^{t,j}\}_{j=0}^{J-1}) 
            \triangleq& \mathcal{E}_{k, \text{bound}}^{t}\left(\{\eta_k^{t,j}\}_{j=0}^{J-1}, \|\theta_k^{t,0} - \theta_k^{t,*}\|_2^2\right) + \left[\lambda_{\max} \left(G D_\theta + \frac{L}{2} {D_\theta}^2 \right) + L_{\mathrm{lip}} D_\theta \right]
            \\
            &+ \lambda_{\max} C_\kappa \cdot \left(2 + \left(8 L_{\mathrm{lip}} B + 4\sqrt{2\log(2/\delta)}\right) \sqrt{\frac{|\mathcal{C}_k^{\leq t-1}|}{M}} \right)
        \end{split}
    \end{equation}
    where $D_\theta \triangleq \sup_{\theta_a,\theta_b \in \Omega_\theta} \|\theta_a-\theta_b\|_2$ is the parameter space diameter; $B$ is the norm radius bound of hypothesis class $\mathcal{F}_{\theta}$ in RKHS $\mathcal{H}_K$; $\lambda_{\max}$ is the maximum replay loss weight; and the optimization error bound $\mathcal{E}_{k, \text{bound}}^{t}\left(\{\eta_k^{t,j}\}_{j=0}^{J-1}, \|\theta_k^{t,0} - \theta_k^{t,*}\|_2^2\right)$ and kernel spectral efficiency constant $C_\kappa$ are defined in Theorem 1 and Theorem 2, respectively.
    For the initial FL round $t=1$, the regret functional is upper-bounded by:
    \begin{equation}
        \begin{split}
            \mathrm{Regret}_k^{\mathrm{local}}(1; \theta_k^{1, 0}, \{\eta_k^{1,j}\}_{j=0}^{J-1}) 
            \leq& \frac{L}{2} \exp\left(-\mu \sum_{j=0}^{J-1} \eta_k^{1,j}\right) \|\theta_k^{1,0} - \theta_{1,\mathrm{cum}}^{1,*}\|_2^2 + \frac{L}{2} \sum_{j=0}^{J-1} (\eta_k^{1,j})^2 G^2 \prod_{i=j+1}^{J-1}(1-\mu\eta_k^{1,i}) 
            \\
            &+ \frac{4 L_{\mathrm{Lip}} B}{\sqrt{n_k^1}} + \sqrt{\frac{8\log(2/\delta)}{n_k^1}} 
        \end{split}
    \end{equation}
\end{lemma}

\textbf{Proof}:

Let $n_{k}^\tau$ be the local data size and $p_k^\tau$ be the corresponding empirical distribution received by client $k$ at round $\tau$. The cumulative data distribution over the first $t$ rounds is defined as the data-size-weighted average of the empirical distributions, i.e., $p_k^{\leq t} \triangleq \frac{1}{\sum_{\tau=1}^t n_{k}^\tau} \sum_{\tau=1}^t n_{k}^\tau p_k^\tau$. Furthermore, let $\omega_{k}^{t} \triangleq n_{k}^t / \sum_{\tau=1}^t n_{k}^\tau$ denote the weight of the $t$-th FL round's distribution $p_k^t$, which can be recursively expressed as:
\begin{equation}
    \begin{split}
        p_k^{\leq t} 
        =& \frac{1}{\sum_{\tau=1}^{t} n_{k}^\tau} \sum_{\tau=1}^{t-1} n_{k}^\tau p_k^\tau + \frac{n_{k}^\tau}{\sum_{\tau=1}^{t} n_{k}^\tau} p_k^t
        = \frac{\sum_{\tau=1}^{t-1} n_{k}^\tau}{\sum_{\tau=1}^{t} n_{k}^\tau} \left( \frac{1}{\sum_{\tau=1}^{t-1} n_{k}^\tau} \sum_{\tau=1}^{t-1} n_{k}^\tau p_k^\tau \right) + \frac{n_{k}^\tau}{\sum_{\tau=1}^{t} n_{k}^\tau} p_k^t 
        \\
        =& \frac{\sum_{\tau=1}^{t-1} n_{k}^\tau}{\sum_{\tau=1}^{t} n_{k}^\tau} p_k^{\leq t-1} + \frac{n_{k}^\tau}{\sum_{\tau=1}^{t} n_{k}^\tau} p_k^t
        = (1 - \omega_{k}^{t}) p_k^{\leq t-1} + \omega_{k}^{t} p_k^t
    \end{split}
\end{equation}

Clearly, $p_k^{\leq t}$ is a convex combination of $p_k^{\leq t-1}$ and $p_k^{t}$. Substituting this into the definition of regret yields:
\begin{equation}
    \begin{split}
        \mathrm{Regret}_k^{\mathrm{local}}(t; \theta_k^{t,0}, \{\eta_k^{t,j}\}_{j=0}^{J-1}) 
        =& \mathbb{E}_{p_k^{\leq t}}[\mathcal{R}(f_{\theta_k^{t,J}}) - \mathcal{R}(f^{t,*}_{k,\mathrm{cum}})]
        = \mathbb{E}_{(1 - \omega_{k}^{t}) p_k^{\leq t-1} + \omega_{k}^{t} p_k^t}[\mathcal{R}(f_{\theta_k^{t,J}}) - \mathcal{R}(f^{t,*}_{k,\mathrm{cum}})]
        \\
        =& (1 - \omega_{k}^{t}) \mathbb{E}_{p_k^{\leq t-1}}[\mathcal{R}(f_{\theta_k^{t,J}}) - \mathcal{R}(f^{t,*}_{k,\mathrm{cum}})] 
        + \omega_{k}^{t} \mathbb{E}_{p_k^{t}}[\mathcal{R}(f_{\theta_k^{t,J}}) - \mathcal{R}(f^{t,*}_{k,\mathrm{cum}})]
    \end{split}
\end{equation}

For the first term $\mathbb{E}_{p_k^{\leq t-1}}[\mathcal{R}(f_{\theta_k^{t,J}}) - \mathcal{R}(f^{t,*}_{k,\mathrm{cum}})]$, we expand as follows:
\begin{equation}
    \begin{split}
        &\mathbb{E}_{p_k^{\leq t-1}}[\mathcal{R}(f_{\theta_k^{t,J}}) - \mathcal{R}(f^{t,*}_{k,\mathrm{cum}})]
        \\
        =& \mathbb{E}_{p_k^{\leq t-1}}\left[ \mathcal{R}(f_{\theta_k^{t,J}}) - \mathcal{R}(f_{\theta_k^{t-1,J}}) + \mathcal{R}(f_{\theta_k^{t-1,J}}) - \mathcal{R}(f^{t-1,*}_{k,\mathrm{cum}}) + \mathcal{R}(f^{t-1,*}_{k,\mathrm{cum}})- \mathcal{R}(f^{t,*}_{k,\mathrm{cum}})\right]
        \\
        =& \mathbb{E}_{p_k^{\leq t-1}}\left[ \mathcal{R}(f_{\theta_k^{t,J}}) - \mathcal{R}(f_{\theta_k^{t-1,J}})\right]
        + \mathbb{E}_{p_k^{\leq t-1}}\left[\mathcal{R}(f_{\theta_k^{t-1,J}}) - \mathcal{R}(f^{t-1,*}_{k,\mathrm{cum}})\right]
        + \mathbb{E}_{p_k^{\leq t-1}}\left[\mathcal{R}(f^{t-1,*}_{k,\mathrm{cum}})- \mathcal{R}(f^{t,*}_{k,\mathrm{cum}})\right]
        \\
        =& \mathbb{E}_{p_k^{\leq t-1}}\left[ \mathcal{R}(f_{\theta_k^{t,J}}) - \mathcal{R}(f_{\theta_k^{t-1,J}})\right]
        + \mathrm{Regret}_k^{\mathrm{local}}(t-1; \theta_k^{t-1,0}, \{\eta_k^{t-1,j}\}_{j=0}^{J-1})
        + \mathbb{E}_{p_k^{\leq t-1}}\left[\mathcal{R}(f^{t-1,*}_{k,\mathrm{cum}})- \mathcal{R}(f^{t,*}_{k,\mathrm{cum}})\right]
    \end{split}
\end{equation}

Let $\Delta_{k, \mathrm{shift}}(t) \triangleq \mathbb{E}_{p_k^{\leq t-1}}\left[\mathcal{R}(f^{t-1,*}_{k,\mathrm{cum}}) - \mathcal{R}(f^{t,*}_{k,\mathrm{cum}})\right]$ denote the distribution shift term. Substituting this into the above inequality yields the recursive bound for the local regret:
\begin{equation}
    \begin{split}
        \mathrm{Regret}_k^{\mathrm{local}}(t; \theta_k^{t,0}, \{\eta_k^{t,j}\}_{j=0}^{J-1}) 
        =& \left(1 - \omega_{k}^{t}\right) \mathrm{Regret}_k^{\mathrm{local}}(t-1; \theta_k^{t-1,0}, \{\eta_k^{t-1,j}\}_{j=0}^{J-1}) 
        \\
        &+ \left(1 - \omega_{k}^{t}\right) \mathbb{E}_{p_k^{\leq t-1}}\left[ \mathcal{R}(f_{\theta_k^{t,J}}) - \mathcal{R}(f_{\theta_k^{t-1,J}})\right] 
        + \left(1 - \omega_{k}^{t}\right) \Delta_{k, \mathrm{shift}}(t)
        \\
        &+ \omega_{k}^{t} \cdot \mathbb{E}_{p_k^{t}}[\mathcal{R}(f_{\theta_k^{t,J}}) - \mathcal{R}(f^{t,*}_{k,\mathrm{cum}})]  
    \end{split}
\end{equation}

Next, consider the term $\mathbb{E}_{p_k^{t}}[\mathcal{R}(f_{\theta_k^{t,J}}) - \mathcal{R}(f^{t,*}_{k,\mathrm{cum}})]$. Since the data $D_k^t$ are drawn i.i.d.\ from the distribution $p_k^t$, we have $\mathbb{E}_{p_k^t}[\mathcal{R}(f_\theta)] = \mathbb{E}_{p_k^t}[L_{k,\text{task}}^t(\theta)]$, which yields:
\begin{equation}
    \begin{split}
        \mathbb{E}_{p_k^{t}}[\mathcal{R}(f_{\theta_k^{t,J}}) - \mathcal{R}(f^{t,*}_{k,\mathrm{cum}})]
        =& \mathbb{E}_{p_k^{t}}[\mathcal{R}(f_{\theta_k^{t,J}})] - \mathbb{E}_{p_k^{t}}[\mathcal{R}(f^{t,*}_{k,\mathrm{cum}})]
        \\
        =& \mathbb{E}_{p_k^{t}}[L_{k,\text{task}}^t(\theta_k^{t,J})] -\mathbb{E}_{p_k^{t}}[L_{k,\text{task}}^t(\theta_k^{t,*})] + \mathbb{E}_{p_k^{t}}[L_{k,\text{task}}^t(\theta_k^{t,*})] - \mathbb{E}_{p_k^{t}}[\mathcal{R}(f^{t,*}_{k,\mathrm{cum}})]
        \\
        =& \left(L_{k,\text{task}}^t(\theta_k^{t,J}) - L_{k,\text{task}}^t(\theta_k^{t,*})\right) + \left(\mathbb{E}_{p_k^{t}}[\mathcal{R}(f_{\theta_k^{t,*}})] - \mathbb{E}_{p_k^{t}}[\mathcal{R}(f^{t,*}_{k,\mathrm{cum}})]\right)
    \end{split}
\end{equation}

Further, at the $J$-th epoch, the total loss is given by $L_{k,\text{total}}^t(\theta) = L_{k,\text{task}}^t(\theta) + \lambda_k^{t,J} L_{k,\text{rep}}^t(\theta)$, which yields $L_{k,\text{task}}^t(\theta) = L_{k,\text{total}}^t(\theta) - \lambda_k^{t,J} L_{k,\text{rep}}^t(\theta)$. Additionally, the replay loss can be expressed as $L_{k,\text{rep}}^{t}(\theta) = \hat{\mathcal{R}}_{\mathcal{M}_k^t}(f_\theta)$. Noting that $\theta_k^{t,J}$ and $\theta_k^{t,*}$ are fixed parameters, we obtain:
\begin{equation}
    \begin{split}
        \mathbb{E}_{p_k^{t}}[\mathcal{R}(f_{\theta_k^{t,J}}) - \mathcal{R}(f^{t,*}_{k,\mathrm{cum}})]
        =& \left(L_{k,\text{total}}^t(\theta_k^{t,J}) - L_{k,\text{total}}^t(\theta_k^{t,*})\right) - \lambda_k^{t,J}\left(L_{k,\text{rep}}^t(\theta_k^{t,J}) - L_{k,\text{rep}}^t(\theta_k^{t,*})\right) 
        \\
        &+ \left(\mathbb{E}_{p_k^{t}}[\mathcal{R}(f_{\theta_k^{t,*}})] - \mathbb{E}_{p_k^{t}}[\mathcal{R}(f^{t,*}_{k,\mathrm{cum}})]\right)
        \\
        =& \left(L_{k,\text{total}}^t(\theta_k^{t,J}) - L_{k,\text{total}}^t(\theta_k^{t,*})\right) - \lambda_k^{t,J}\left(\hat{\mathcal{R}}_{\mathcal{M}_k^t}(f_{\theta_k^{t,J}}) - \hat{\mathcal{R}}_{\mathcal{M}_k^t}(f_{\theta_k^{t,*}})\right) 
        \\
        &+ \left(\mathbb{E}_{p_k^{t}}[\mathcal{R}(f_{\theta_k^{t,*}})] - \mathbb{E}_{p_k^{t}}[\mathcal{R}(f^{t,*}_{k,\mathrm{cum}})]\right)
    \end{split}
\end{equation}

For the first term $L_{k,\text{total}}^t(\theta_k^{t,J}) - L_{k,\text{total}}^t(\theta_k^{t,*})$, applying Theorem 1 yields:
\begin{equation}
    \begin{split}
        L_{k,\text{total}}^t(\theta_k^{t,J}) - L_{k,\text{total}}^t(\theta_k^{t,*}) 
        &= \mathbb{E}\left[L_{k,\text{total}}^t(\theta_k^{t,J}|\lambda_k^{t,J})\right] - L_{k,\text{total}}^t(\theta_k^{t,*}|\lambda_k^{t,*}) 
        \\
        &\leq \mathcal{E}_{k, \text{bound}}^{t}\left(\{\eta_k^{t,j}\}_{j=0}^{J-1}, \|\theta_k^{t,0} - \theta_k^{t,*}\|_2^2\right)
    \end{split}
\end{equation}

For the second term $-\lambda_k^{t,J}\left(\hat{\mathcal{R}}_{\mathcal{M}_k^t}(f_{\theta_k^{t,J}}) - \hat{\mathcal{R}}_{\mathcal{M}_k^t}(f_{\theta_k^{t,*}})\right)$, let $\lambda_{\max} \triangleq \sup_{\lambda \in \Omega_\lambda} \lambda < \infty$ denote the upper bound of the global replay weight. By taking absolute values and applying a conservative bound, we obtain:
\begin{equation}
    \begin{split}
        & -\lambda_k^{t,J}\left(\hat{\mathcal{R}}_{\mathcal{M}_k^t}(f_{\theta_k^{t,J}}) - \hat{\mathcal{R}}_{\mathcal{M}_k^t}(f_{\theta_k^{t,*}})\right) 
        \\
        =& -\lambda_k^{t,J}\left(
            \hat{\mathcal{R}}_{\mathcal{M}_k^t}(f_{\theta_k^{t,J}}) - \mathcal{R}_{p_k^{\leq t-1}}(f_{\theta_k^{t,J}}) 
            + \mathcal{R}_{p_k^{\leq t-1}}(f_{\theta_k^{t,J}}) - \mathcal{R}_{p_k^{\leq t-1}}(f_{\theta_k^{t,*}}) 
            + \mathcal{R}_{p_k^{\leq t-1}}(f_{\theta_k^{t,*}}) - \hat{\mathcal{R}}_{\mathcal{M}_k^t}(f_{\theta_k^{t,*}})\right) 
        \\
        =& -\lambda_k^{t,J}\left(
            \left[\hat{\mathcal{R}}_{\mathcal{M}_k^t}(f_{\theta_k^{t,J}}) - \mathcal{R}_{p_k^{\leq t-1}}(f_{\theta_k^{t,J}}) \right]
            - \left[\hat{\mathcal{R}}_{\mathcal{M}_k^t}(f_{\theta_k^{t,*}}) - \mathcal{R}_{p_k^{\leq t-1}}(f_{\theta_k^{t,*}})\right]
            + \left[\mathcal{R}_{p_k^{\leq t-1}}(f_{\theta_k^{t,J}}) - \mathcal{R}_{p_k^{\leq t-1}}(f_{\theta_k^{t,*}})\right]
            \right) 
        \\
        \leq& \lambda_{\max} \left[\left|\hat{\mathcal{R}}_{\mathcal{M}_k^t}(f_{\theta_k^{t,J}}) - \mathcal{R}_{p_k^{\leq t-1}}(f_{\theta_k^{t,J}})\right| + \left|\hat{\mathcal{R}}_{\mathcal{M}_k^t}(f_{\theta_k^{t,*}}) - \mathcal{R}_{p_k^{\leq t-1}}(f_{\theta_k^{t,*}})\right| + \left|\mathcal{R}_{p_k^{\leq t-1}}(f_{\theta_k^{t,J}}) - \mathcal{R}_{p_k^{\leq t-1}}(f_{\theta_k^{t,*}})\right|\right]   
    \end{split}
\end{equation}

For the third term $\left|\mathcal{R}_{p_k^{\leq t-1}}(f_{\theta_k^{t,J}}) - \mathcal{R}_{p_k^{\leq t-1}}(f_{\theta_k^{t,*}})\right|$, let $D_\theta \triangleq \sup_{\theta_a,\theta_b \in \Omega_\theta} \|\theta_a-\theta_b\|_2 < \infty$ denote the diameter of the parameter space $\Omega_\theta$. By combining Assumption 10 and Assumption 13, we obtain the following bound:
\begin{equation}
    \begin{split}
        \left|\mathcal{R}_{p_k^{\leq t-1}}(f_{\theta_k^{t,J}}) - \mathcal{R}_{p_k^{\leq t-1}}(f_{\theta_k^{t,*}})\right|
        \leq& \left|\left\langle \nabla_\theta \mathcal{R}_{p_k^{\leq t-1}}(f_{\theta_k^{t,*}}), \theta_k^{t,J} - \theta_k^{t,*} \right\rangle + \frac{L}{2}\|\theta_k^{t,J} - \theta_k^{t,*}\|_2^2\right|
        \\
        \leq& \left(G\|\theta_k^{t,J} - \theta_k^{t,*}\|_2 + \frac{L}{2}\|\theta_k^{t,J} - \theta_k^{t,*}\|_2^2\right)
        \\
        \leq& \left(G D_\theta + \frac{L}{2} {D_\theta}^2 \right)
    \end{split}
\end{equation}

Thus, we obtain:
\begin{equation}
    \begin{split}
        -\lambda_k^{t,J}\left(\hat{\mathcal{R}}_{\mathcal{M}_k^t}(f_{\theta_k^{t,J}}) - \hat{\mathcal{R}}_{\mathcal{M}_k^t}(f_{\theta_k^{t,*}})\right) 
        \leq& \lambda_{\max} \left[\left|\hat{\mathcal{R}}_{\mathcal{M}_k^t}(f_{\theta_k^{t,J}}) - \mathcal{R}_{p_k^{\leq t-1}}(f_{\theta_k^{t,J}})\right| + \left|\hat{\mathcal{R}}_{\mathcal{M}_k^t}(f_{\theta_k^{t,*}}) - \mathcal{R}_{p_k^{\leq t-1}}(f_{\theta_k^{t,*}})\right|\right]
        \\
        &+ \lambda_{\max}\left(G D_\theta + \frac{L}{2} {D_\theta}^2 \right)
    \end{split}
\end{equation}

For the third term $\mathbb{E}_{p_k^{t}}[\mathcal{R}(f_{\theta_k^{t,*}})] - \mathbb{E}_{p_k^{t}}[\mathcal{R}(f^{t,*}_{k,\mathrm{cum}})]$, let $\theta_{k,\mathrm{cum}}^{t,*}$ denote the parameter of the cumulative optimal model $f^{t,*}_{k,\mathrm{cum}}$. Under Assumption 2 and Assumption 5, we obtain:
\begin{equation}
    \begin{split}
        \mathbb{E}_{p_k^{t}}[\mathcal{R}(f_{\theta_k^{t,*}})] - \mathbb{E}_{p_k^{t}}[\mathcal{R}(f^{t,*}_{k,\mathrm{cum}})]
        =& L_{k,\text{task}}^t(\theta_k^{t,*}) - L_{k,\text{task}}^t(\theta_{k,\mathrm{cum}}^{t,*})
        \\
        \leq& L_{\mathrm{lip}} \|\theta_k^{t,*} - \theta^{t,*}_{k,\mathrm{cum}}\|_2
        \leq L_{\mathrm{lip}} D_\theta
    \end{split}
\end{equation}

Substituting the above results yields the following upper bound for $\mathbb{E}_{p_k^{t}}[\mathcal{R}(f_{\theta_k^{t,J}}) - \mathcal{R}(f^{t,*}_{k,\mathrm{cum}})]$:
\begin{equation}
    \begin{split}
        \mathbb{E}_{p_k^{t}}[\mathcal{R}(f_{\theta_k^{t,J}}) - \mathcal{R}(f^{t,*}_{k,\mathrm{cum}})]
        \leq& \mathcal{E}_{k, \text{bound}}^{t}\left(\{\eta_k^{t,j}\}_{j=0}^{J-1}, \|\theta_k^{t,0} - \theta_k^{t,*}\|_2^2\right)
        + \lambda_{\max} \left(G D_\theta + \frac{L}{2} {D_\theta}^2 \right)
        + L_{\mathrm{lip}} D_\theta
        \\
        &+ \lambda_{\max} \left[\left|\hat{\mathcal{R}}_{\mathcal{M}_k^t}(f_{\theta_k^{t,J}}) - \mathcal{R}_{p_k^{\leq t-1}}(f_{\theta_k^{t,J}})\right| + \left|\hat{\mathcal{R}}_{\mathcal{M}_k^t}(f_{\theta_k^{t,*}}) - \mathcal{R}_{p_k^{\leq t-1}}(f_{\theta_k^{t,*}})\right|\right]
    \end{split}
\end{equation}

Next, consider the term $\Delta_{k, \mathrm{shift}}(t) = \mathbb{E}_{p_k^{\leq t-1}}\left[\mathcal{R}(f^{t-1,*}_{k,\mathrm{cum}}) - \mathcal{R}(f^{t,*}_{k,\mathrm{cum}})\right]$. Since $f^{t-1,*}_{k,\mathrm{cum}}$ is the optimal model over the historical distribution $p_k^{\leq t-1}$, we have $\mathbb{E}_{p_k^{\leq t-1}}\left[\mathcal{R}(f^{t,*}_{k,\mathrm{cum}})\right] \geq \mathbb{E}_{p_k^{\leq t-1}}\left[\mathcal{R}(f^{t-1,*}_{k,\mathrm{cum}})\right]$, which yields:
\begin{equation}
    \Delta_{k, \mathrm{shift}}(t) 
    = \mathbb{E}_{p_k^{\leq t-1}}\left[\mathcal{R}(f^{t-1,*}_{k,\mathrm{cum}})- \mathcal{R}(f^{t,*}_{k,\mathrm{cum}})\right] 
    \leq \mathbb{E}_{p_k^{\leq t-1}}\left[\mathcal{R}(f^{t-1,*}_{k,\mathrm{cum}})\right] - \mathbb{E}_{p_k^{\leq t-1}}\left[\mathcal{R}(f^{t-1,*}_{k,\mathrm{cum}})\right] 
    = 0
\end{equation}

Substituting the above results into the recursive bound for the local regret yields:
\begin{equation}
    \begin{split}
        &\mathrm{Regret}_k^{\mathrm{local}}(t; \theta_k^{t,0}, \{\eta_k^{t,j}\}_{j=0}^{J-1})
        \\
        \leq& \left(1 - \omega_{k}^{t}\right) \mathrm{Regret}_k^{\mathrm{local}}(t-1; \theta_k^{t-1,0}, \{\eta_k^{t-1,j}\}_{j=0}^{J-1})
        + \left(1 - \omega_{k}^{t}\right) \mathbb{E}_{p_k^{\leq t-1}}\left[ \mathcal{R}(f_{\theta_k^{t,J}}) - \mathcal{R}(f_{\theta_k^{t-1,J}})\right] 
        \\
        &+ \omega_{k}^{t} \cdot \mathcal{E}_{k, \text{bound}}^{t}\left(\{\eta_k^{t,j}\}_{j=0}^{J-1}, \|\theta_k^{t,0} - \theta_k^{t,*}\|_2^2\right)
        + \omega_{k}^{t} \cdot \left[\lambda_{\max} \left(G D_\theta + \frac{L}{2} {D_\theta}^2 \right) + L_{\mathrm{lip}} D_\theta \right]
        \\
        &+ \omega_{k}^{t} \cdot \lambda_{\max} \left[\left|\hat{\mathcal{R}}_{\mathcal{M}_k^t}(f_{\theta_k^{t,J}}) - \mathcal{R}_{p_k^{\leq t-1}}(f_{\theta_k^{t,J}})\right| + \left|\hat{\mathcal{R}}_{\mathcal{M}_k^t}(f_{\theta_k^{t,*}}) - \mathcal{R}_{p_k^{\leq t-1}}(f_{\theta_k^{t,*}})\right|\right]
    \end{split}
\end{equation}

Next, we bound the term $\left|\hat{\mathcal{R}}_{\mathcal{M}_k^t}(f_{\theta_k^{t,J}}) - \mathcal{R}_{p_k^{\leq t-1}}(f_{\theta_k^{t,J}})\right| + \left|\hat{\mathcal{R}}_{\mathcal{M}_k^t}(f_{\theta_k^{t,*}}) - \mathcal{R}_{p_k^{\leq t-1}}(f_{\theta_k^{t,*}})\right|$. 
First, consider the function $\left|\hat{\mathcal{R}}_{\mathcal{M}_k^t}(f_{\theta}) - \mathcal{R}_{p_k^{\leq t-1}}(f_{\theta})\right|$. Note that $\mathcal{M}_k^t$ is not sampled i.i.d.\ from $p_k^{\leq t-1}$, preventing the direct application of standard i.i.d.\ analysis. To address this, we introduce an auxiliary buffer $\mathcal{M}_{k,\text{rand}}^t$ of capacity $M$, obtained by class-wise random sampling from $p_k^{\leq t-1}$, where the sampling within each class is asymptotically equivalent to i.i.d.\ sampling. Defining its empirical risk as $\hat{\mathcal{R}}_{\mathcal{M}_{k,\text{rand}}^t}(f_{\theta})$ and leveraging the buffer kernel spectral efficiency advantage constant $C_\kappa$ from Theorem 2, we obtain:
\begin{equation}
    \left|\hat{\mathcal{R}}_{\mathcal{M}_k^t}(f_{\theta}) - \mathcal{R}_{p_k^{\leq t-1}}(f_{\theta})\right|
    \leq C_\kappa \cdot \left| \mathcal{R}_{p_k^{\leq t-1}}(f_{\theta}) - \hat{\mathcal{R}}_{\mathcal{M}_{k,\text{rand}}^t}(f_{\theta}) \right|
\end{equation}

For the buffer $\mathcal{M}_{k,\text{rand}}^t$, consider the sub-buffer $\mathcal{M}_{k,c,\text{rand}}^t$ corresponding to class $c \in \mathcal{C}_k^{\leq t-1}$. By Assumption 6, the sample allocation satisfies $|\mathcal{M}_{k,c,\text{rand}}^t| = \Theta(M/|\mathcal{C}_k^{\leq t-1}|)$. 
Let $\mathcal{H}_K$ be the reproducing kernel Hilbert space induced by the kernel function $K(\cdot, \cdot)$, with the hypothesis function class norm bounded by $B$. Denote $\hat{R}_{|\mathcal{M}_{k,c,\text{rand}}^t|}(\cdot)$ as the empirical Rademacher complexity. According to Theorem 8 in \cite{12}, for any model $f_{\theta} \in \mathcal{F}_{\theta} \triangleq \{f\in\mathcal{H}_K:\|f\|_{\mathcal{H}_K}\leq B\}$ and loss function $\ell \in \tilde{\ell}$, the following inequality holds with probability at least $1-\delta$:
\begin{equation}
    \mathcal{R}_{p_k^{\leq t-1}(\cdot|c)}(f_{\theta}) - \hat{\mathcal{R}}_{\mathcal{M}_{k,c,\text{rand}}^t}(f_{\theta}) \leq \hat{R}_{M/|\mathcal{C}_k^{\leq t-1}|}(\tilde{\ell} \circ \mathcal{F}_{\theta}) + \sqrt{\frac{8\log(2/\delta)}{M/|\mathcal{C}_k^{\leq t-1}|}}
\end{equation}

Furthermore, combining Assumption 5 with Theorem 12 in \cite{12}, we have:
\begin{equation}
    \hat{R}_{M/|\mathcal{C}_k^{\leq t-1}|}(\tilde{\ell} \circ \mathcal{F}_{\theta}) \leq 2 L_{\mathrm{lip}} \cdot \hat{R}_{M/|\mathcal{C}_k^{\leq t-1}|}(\mathcal{F}_{\theta})
\end{equation}

Thus, we obtain:
\begin{equation}
    \begin{split}
        \left|\mathcal{R}_{p_k^{\leq t-1}(\cdot|c)}(f_{\theta}) - \hat{\mathcal{R}}_{\mathcal{M}_{k,c,\text{rand}}^t}(f_{\theta})\right|
        \leq& \hat{R}_{M/|\mathcal{C}_k^{\leq t-1}|}(\tilde{\ell} \circ \mathcal{F}_{\theta}) + \sqrt{\frac{8\log(2/\delta)}{M/|\mathcal{C}_k^{\leq t-1}|}}
        \\
        \leq& 2 L_{\mathrm{lip}} \cdot \hat{R}_{M/|\mathcal{C}_k^{\leq t-1}|}(\mathcal{F}_{\theta}) + \sqrt{\frac{8\log(2/\delta)}{M/|\mathcal{C}_k^{\leq t-1}|}}
    \end{split}
\end{equation}

Further, consider the kernel matrix $\mathbf{K}_c \in \mathbb{R}^{(M/|\mathcal{C}_k^{\leq t-1}|) \times (M/|\mathcal{C}_k^{\leq t-1}|)}$ constructed from the sample set $\{x_{i,c}\}_{i=1}^{M/|\mathcal{C}_k^{\leq t-1}|}$, whose trace is defined as $\text{tr}(\mathbf{K}_c) \triangleq \sum_{i=1}^{M/|\mathcal{C}_k^{\leq t-1}|} K(x_{i,c}, x_{i,c})$. According to Lemma 22 in \cite{12}, the empirical Rademacher complexity of the RKHS function class $\mathcal{F}_{\theta}$ satisfies:
\begin{equation}
    \hat{R}_{M/|\mathcal{C}_k^{\leq t-1}|}(\mathcal{F}_{\theta}) \leq \frac{2B}{M/|\mathcal{C}_k^{\leq t-1}|} \sqrt{\text{tr}(\mathbf{K}_c)} = \frac{2B}{M/|\mathcal{C}_k^{\leq t-1}|} \sqrt{\sum_{i=1}^{M/|\mathcal{C}_k^{\leq t-1}|} K(x_{i,c}, x_{i,c})}
\end{equation}

Under the standard kernel normalization condition, $K(x, x) \leq 1$ holds for any input $x$, which implies $\sum_{i=1}^{M/|\mathcal{C}_k^{\leq t-1}|} K(x_{i,c}, x_{i,c}) \leq M/|\mathcal{C}_k^{\leq t-1}|$. Substituting this into the above inequality yields:
\begin{equation}
    \begin{split}
        \left|\mathcal{R}_{p_k^{\leq t-1}(\cdot|c)}(f_{\theta}) - \hat{\mathcal{R}}_{\mathcal{M}_{k,c,\text{rand}}^t}(f_{\theta})\right|
        \leq& 2 L_{\mathrm{lip}} \cdot \frac{2B}{M/|\mathcal{C}_k^{\leq t-1}|} \sqrt{\sum_{i=1}^{M/|\mathcal{C}_k^{\leq t-1}|} K(x_{i,c}, x_{i,c})} + \sqrt{\frac{8\log(2/\delta)}{M/|\mathcal{C}_k^{\leq t-1}|}}
        \\
        \leq& \frac{4 L_{\mathrm{lip}} B}{M/|\mathcal{C}_k^{\leq t-1}|} \sqrt{M/|\mathcal{C}_k^{\leq t-1}|} + \sqrt{\frac{8\log(2/\delta)}{M/|\mathcal{C}_k^{\leq t-1}|}}
        \\
        \leq& \left(4 L_{\mathrm{lip}} B + \sqrt{8\log(2/\delta)}\right) \sqrt{\frac{|\mathcal{C}_k^{\leq t-1}|}{M}}
    \end{split}
\end{equation}

Consequently, we obtain the upper bound on the risk deviation of the buffer for a single class. Since the global risk is the weighted average of the conditional risks over classes with prior probabilities, let $\pi_{k}^{\leq t-1}(c)$ denote the prior weight of class $c$, satisfying $\sum_{c \in \mathcal{C}_k^{\leq t-1}}\pi_{k}^{\leq t-1}(c) = 1$. The global true risk and empirical risk can be respectively expressed as:
\begin{equation}
    \mathcal{R}_{p_k^{\leq t-1}}(f_{\theta}) = \sum_{c \in \mathcal{C}_k^{\leq t-1}} \pi_{k}^{\leq t-1}(c) \mathcal{R}_{p_k^{\leq t-1}(\cdot|c)}(f_{\theta}), 
    \quad
    \hat{\mathcal{R}}_{\mathcal{M}_{k,\text{rand}}^t}(f_{\theta}) = \sum_{c \in \mathcal{C}_k^{\leq t-1}} \frac{1}{|\mathcal{C}_k^{\leq t-1}|} \hat{\mathcal{R}}_{\mathcal{M}_{k,c,\text{rand}}^t}(f_{\theta})
\end{equation}

Applying the triangle inequality yields:
\begin{equation}
    \begin{split}
        &\left|\mathcal{R}_{p_k^{\leq t-1}}(f_{\theta}) - \hat{\mathcal{R}}_{\mathcal{M}_{k,\text{rand}}^t}(f_{\theta})\right|
        \\
        \leq& \left| \sum_{c \in \mathcal{C}_k^{\leq t-1}} \pi_{k}^{\leq t-1}(c) \left(\mathcal{R}_{p_k^{\leq t-1}(\cdot|c)}(f_{\theta}) - \hat{\mathcal{R}}_{\mathcal{M}_{k,c,\text{rand}}^t}(f_{\theta})\right) \right|
        + \left| \sum_{c \in \mathcal{C}_k^{\leq t-1}} \left(\frac{1}{|\mathcal{C}_k^{\leq t-1}|} - \pi_{k}^{\leq t-1}(c)\right) \hat{\mathcal{R}}_{\mathcal{M}_{k,c,\text{rand}}^t}(f_{\theta}) \right|
        \\
        \leq& \sum_{c \in \mathcal{C}_k^{\leq t-1}} \pi_{k}^{\leq t-1}(c) \left|\mathcal{R}_{p_k^{\leq t-1}(\cdot|c)}(f_{\theta}) - \hat{\mathcal{R}}_{\mathcal{M}_{k,c,\text{rand}}^t}(f_{\theta})\right| 
        + \left| \sum_{c \in \mathcal{C}_k^{\leq t-1}} \left(\frac{1}{|\mathcal{C}_k^{\leq t-1}|} - \pi_{k}^{\leq t-1}(c)\right) \hat{\mathcal{R}}_{\mathcal{M}_{k,c,\text{rand}}^t}(f_{\theta}) \right|
        \\
        \leq& \left(4 L_{\mathrm{lip}} B + \sqrt{8\log(2/\delta)}\right) \sqrt{\frac{|\mathcal{C}_k^{\leq t-1}|}{M}} 
        + \left| \sum_{c \in \mathcal{C}_k^{\leq t-1}} \left(\frac{1}{|\mathcal{C}_k^{\leq t-1}|} - \pi_{k}^{\leq t-1}(c)\right) \hat{\mathcal{R}}_{\mathcal{M}_{k,c,\text{rand}}^t}(f_{\theta}) \right|
    \end{split}
\end{equation}

For the term $\left| \sum_{c \in \mathcal{C}_k^{\leq t-1}} \left(\frac{1}{|\mathcal{C}_k^{\leq t-1}|} - \pi_{k}^{\leq t-1}(c)\right) \hat{\mathcal{R}}_{\mathcal{M}_{k,c,\text{rand}}^t}(f_{\theta}) \right|$, noting that $\hat{\mathcal{R}}_{\mathcal{M}_{k,c,\text{rand}}^t}(f_{\theta}) \leq 1$ due to the boundedness of the loss function in Assumption 5, applying the variational characterization of the total variation distance over the finite class set $\mathcal{C}_k^{\leq t-1}$ yields:
\begin{equation}
    \left| \sum_{c \in \mathcal{C}_k^{\leq t-1}} \left(\frac{1}{|\mathcal{C}_k^{\leq t-1}|} - \pi_{k}^{\leq t-1}(c)\right) \hat{\mathcal{R}}_{\mathcal{M}_{k,c,\text{rand}}^t}(f_{\theta}) \right|
    \leq D_{\mathrm{TV}}\left(\frac{1}{|\mathcal{C}_k^{\leq t-1}|} \middle\| \pi_{k}^{\leq t-1}(c)\right) \leq 1
\end{equation}

Substituting this back into the above inequality yields:
\begin{equation}
    \begin{split}
        \left|\mathcal{R}_{p_k^{\leq t-1}}(f_{\theta}) - \hat{\mathcal{R}}_{\mathcal{M}_{k,\text{rand}}^t}(f_{\theta})\right|
        \leq& 1 + \left(4 L_{\mathrm{lip}} B + \sqrt{8\log(2/\delta)}\right) \sqrt{\frac{|\mathcal{C}_k^{\leq t-1}|}{M}} 
    \end{split}
\end{equation}

Consequently, we obtain the upper bound for $\left|\hat{\mathcal{R}}_{\mathcal{M}_{k,\text{rand}}^t}(f_{\theta}) - \mathcal{R}_{p_k^{\leq t-1}}(f_{\theta})\right|$. Incorporating the global spectral efficiency advantage constant $C_\kappa$ from Theorem 2 yields:
\begin{equation}
    \begin{split}
        \left|\hat{\mathcal{R}}_{\mathcal{M}_k^t}(f_{\theta}) - \mathcal{R}_{p_k^{\leq t-1}}(f_{\theta})\right|
        \leq& C_\kappa \cdot \left| \mathcal{R}_{p_k^{\leq t-1}}(f_{\theta}) - \hat{\mathcal{R}}_{\mathcal{M}_{k,\text{rand}}^t}(f_{\theta}) \right|
        \\
        \leq& C_\kappa \cdot \left(1 + \left(4 L_{\mathrm{lip}} B + \sqrt{8\log(2/\delta)}\right) \sqrt{\frac{|\mathcal{C}_k^{\leq t-1}|}{M}} \right)
    \end{split}
\end{equation}

Finally, we obtain:
\begin{equation}
    \begin{split}
        \left|\hat{\mathcal{R}}_{\mathcal{M}_k^t}(f_{\theta_k^{t,J}}) - \mathcal{R}_{p_k^{\leq t-1}}(f_{\theta_k^{t,J}})\right| + \left|\hat{\mathcal{R}}_{\mathcal{M}_k^t}(f_{\theta_k^{t,*}}) - \mathcal{R}_{p_k^{\leq t-1}}(f_{\theta_k^{t,*}})\right|
        \leq& 2 C_\kappa \cdot \left(1 + \left(4 L_{\mathrm{lip}} B + \sqrt{8\log(2/\delta)}\right) \sqrt{\frac{|\mathcal{C}_k^{\leq t-1}|}{M}} \right)
        \\
        =& C_\kappa \cdot \left(2 + \left(8 L_{\mathrm{lip}} B + 4\sqrt{2\log(2/\delta)}\right) \sqrt{\frac{|\mathcal{C}_k^{\leq t-1}|}{M}} \right)
    \end{split}
\end{equation}

Substituting this into the recursive bound for the local regret yields:
\begin{equation}
    \begin{split}
        &\mathrm{Regret}_k^{\mathrm{local}}(t; \theta_k^{t,0}, \{\eta_k^{t,j}\}_{j=0}^{J-1})
        \\
        \leq& \left(1 - \omega_{k}^{t}\right) \mathrm{Regret}_k^{\mathrm{local}}(t-1; \theta_k^{t-1,0}, \{\eta_k^{t-1,j}\}_{j=0}^{J-1})
        + \left(1 - \omega_{k}^{t}\right) \mathbb{E}_{p_k^{\leq t-1}}\left[ \mathcal{R}(f_{\theta_k^{t,J}}) - \mathcal{R}(f_{\theta_k^{t-1,J}})\right] 
        \\
        &+ \omega_{k}^{t} \cdot \mathcal{E}_{k, \text{bound}}^{t}\left(\{\eta_k^{t,j}\}_{j=0}^{J-1}, \|\theta_k^{t,0} - \theta_k^{t,*}\|_2^2\right)
        + \omega_{k}^{t} \cdot \left[\lambda_{\max} \left(G D_\theta + \frac{L}{2} {D_\theta}^2 \right) + L_{\mathrm{lip}} D_\theta \right]
        \\
        &+ \omega_{k}^{t} \cdot \lambda_{\max} C_\kappa \cdot \left(2 + \left(8 L_{\mathrm{lip}} B + 4\sqrt{2\log(2/\delta)}\right) \sqrt{\frac{|\mathcal{C}_k^{\leq t-1}|}{M}} \right)
    \end{split}
\end{equation}

To simplify the notation, we define the composite error function $U_k(t; \theta_k^{t,0}, \{\eta_k^{t,j}\}_{j=0}^{J-1})$ as follows:
\begin{equation}
    \begin{split}
        U_k(t; \theta_k^{t,0}, \{\eta_k^{t,j}\}_{j=0}^{J-1}) 
        \triangleq& \mathcal{E}_{k, \text{bound}}^{t}\left(\{\eta_k^{t,j}\}_{j=0}^{J-1}, \|\theta_k^{t,0} - \theta_k^{t,*}\|_2^2\right) + \left[\lambda_{\max} \left(G D_\theta + \frac{L}{2} {D_\theta}^2 \right) + L_{\mathrm{lip}} D_\theta \right]
        \\
        &+ \lambda_{\max} C_\kappa \cdot \left(2 + \left(8 L_{\mathrm{lip}} B + 4\sqrt{2\log(2/\delta)}\right) \sqrt{\frac{|\mathcal{C}_k^{\leq t-1}|}{M}} \right)
    \end{split}
\end{equation}

where $D_\theta \triangleq \sup_{\theta_a,\theta_b \in \Omega_\theta} \|\theta_a-\theta_b\|_2$ is the diameter of the parameter space $\Omega_\theta$; $B$ is the upper bound on the norm radius of the hypothesis class $\mathcal{F}_{\theta}$ in the RKHS; and $\lambda_{\max}$ is the maximum replay loss weight. Consequently, the recursive bound for the regret can be expressed as:
\begin{equation}
    \begin{split}
        \mathrm{Regret}_k^{\mathrm{local}}(t; \theta_k^{t,0}, \{\eta_k^{t,j}\}_{j=0}^{J-1})
        \leq& \left(1 - \omega_{k}^{t}\right) \mathrm{Regret}_k^{\mathrm{local}}(t-1; \theta_k^{t-1,0}, \{\eta_k^{t-1,j}\}_{j=0}^{J-1})
        + \omega_{k}^{t} \cdot U_k(t; \theta_k^{t,0}, \{\eta_k^{t,j}\}_{j=0}^{J-1})
        \\
        &+ \left(1 - \omega_{k}^{t}\right) \mathbb{E}_{p_k^{\leq t-1}}\left[ \mathcal{R}(f_{\theta_k^{t,J}}) - \mathcal{R}(f_{\theta_k^{t-1,J}})\right] 
    \end{split}
\end{equation}

For the initial local regret $\mathrm{Regret}_k^{\mathrm{local}}(1; \theta_k^{1, 0}, \{\eta_k^{1,j}\}_{j=0}^{J-1})$ at $t=1$, since the buffer is not yet initialized, the local training reduces to standard SGD optimization under the task loss $L_{k,\text{task}}^1$ only. Under Assumptions 3 and 10-13, the optimization error is bounded by the SGD convergence bound given in Theorem 1 of \cite{17}; under Assumptions 4 and 5, the generalization gap between the empirical and expected risks is controlled by the i.i.d. RKHS generalization bound jointly established by Theorems 8, 12, and 22 in \cite{17}. Combining these two bounds and applying the $L$-smoothness in Assumption 10 yields:
\begin{equation}
    \begin{split}
        \mathrm{Regret}_k^{\mathrm{local}}(1; \theta_k^{1, 0}, \{\eta_k^{1,j}\}_{j=0}^{J-1}) 
        \leq& \frac{L}{2} \exp\left(-\mu \sum_{j=0}^{J-1} \eta_k^{1,j}\right) \|\theta_k^{1,0} - \theta_{1,\mathrm{cum}}^{1,*}\|_2^2 + \frac{L}{2} \sum_{j=0}^{J-1} (\eta_k^{1,j})^2 G^2 \prod_{i=j+1}^{J-1}(1-\mu\eta_k^{1,i}) 
        \\
        &+ \frac{4 L_{\mathrm{Lip}} B}{\sqrt{n_k^1}} + \sqrt{\frac{8\log(2/\delta)}{n_k^1}} 
    \end{split}
\end{equation}

This completes the proof of Lemma 2.

\newpage
\subsection{Proof of Lemma 3}
\label{Appendix B.6}
\setcounter{theorem}{2}   
\begin{lemma}[Single-step regret bound of the FedKACE global model over client cumulative distributions]
    Under standard assumptions (see s\hyperref[Appendix B.1]{Appendix B.1}), let each client in the $t$-th FL round initialize with the global model $\theta_g^{t-1}$, yielding the aggregated global model $\theta_g^{t} = \frac{1}{K}\sum_{k'=1}^K \theta_{k'}^{t,J}$ after $J$ local steps. 
    Meanwhile, define the regret of model $\theta$ on client $k$ at FL round $t$ as $\mathrm{Regret}_k^{t}(\theta) \triangleq \mathbb{E}_{p_k^{\leq t}}\left[\mathcal{R}(f_{\theta}) - \mathcal{R}(f_{k,\mathrm{cum}}^{t,*})\right]$, and the global model's regret functional on client $k$ as $\mathrm{Regret}_k^{\mathrm{global}}(t; \theta_g^{t-1}, \mathcal{I}_t) \triangleq \mathrm{Regret}_k^{t}(\theta_g^{t})$, where $\mathcal{I}_t \triangleq \left\{ \{\eta_{k'}^{t,j}\}_{j=0}^{J-1} \mid \forall k' \in \mathcal{K}\right\} $ is the set of local learning rate sequences at FL round $t$. This functional satisfies the following explicit single-step upper bound:
    \begin{equation}
        \begin{split}
            \mathrm{Regret}_k^{\mathrm{global}}(t; \theta_g^{t-1}, \mathcal{I}_t)
            \leq& \frac{1}{K}\sum_{k'=1}^K \mathrm{Regret}_{k'}^{\mathrm{local}}(t; \theta_g^{t-1}, \{\eta_{k'}^{t,j}\}_{j=0}^{J-1}) 
            + \frac{1}{K}\sum_{k'=1, k' \neq k}^K \mathbb{E}_{p_{k}^{\leq t}}[\mathcal{R}(f_{k',\mathrm{cum}}^{t,*}) - \mathcal{R}(f_{k,\mathrm{cum}}^{t,*})]
            \\
            &+ \frac{2(K-1)}{K} \Delta^{\mathrm{TV}}(t, \nu) - \frac{\mu}{2K}\sum_{k'=1}^K \|\theta_{k'}^{t,J} - \theta_g^t\|_2^2
        \end{split}
    \end{equation}
    where:
    \begin{equation}
        \Delta^{\mathrm{TV}}(t, \nu) \triangleq \frac{1}{(t+1)^{\nu}} \left(2^{\nu} + \frac{3}{\nu} \left[\frac{(3/2)^{\nu}}{2} + \frac{{\nu} 2^{{\nu}-1} }{1 - {\nu}}\right] \right) + \frac{3}{\nu^2}
    \end{equation}
    is the global upper bound of the total variation (TV) distance between the cumulative distributions of any pair of clients $\|p_k^{\leq t} - p_{k'}^{\leq t}\|_{\mathrm{TV}}, \forall k, k' \in \mathcal{K}$ at FL round $t$; $\mathrm{Regret}_{k'}^{\mathrm{local}}(t; \cdot)$ is the local model regret functional defined in Lemma 1; and $\nu = n_{\min}/n_{\max} \leq 1$, with $n_{\textrm{max}} \triangleq \max_{k' \in \mathcal{K},\tau \leq T}(n_{k'}^\tau)$ and $n_{\textrm{min}} \triangleq \min_{k' \in \mathcal{K},\tau \leq T}(n_{k'}^\tau)$ being the maximum and minimum possible per-round data sizes across all $K$ clients over all $T$ FL rounds, respectively.
\end{lemma}

\textbf{Proof}:

Since the global model $\theta_g^{t} = \frac{1}{K}\sum_{k'=1}^K \theta_{k'}^{t,J}$ is a convex combination of the local models, according to Assumption 11 and Jensen's inequality for strongly convex functions, the corresponding risk function satisfies:
\begin{equation}
    \begin{split}
        \mathcal{R}(f_{\theta_g^t}) 
        =& \mathcal{R}(f_{\frac{1}{K}\sum_{k'=1}^K \theta_{k'}^{t,J}})
        \\
        \leq& \sum_{k'=1}^K \frac{1}{K} \mathcal{R}(f_{\theta_{k'}^{t,J}}) - \frac{\mu}{2} \sum_{k'=1}^K \frac{1}{K} \|\theta_{k'}^{t,J} - \frac{1}{K}\sum_{k''=1}^K \theta_{k''}^{t,J}\|_2^2
        \\
        =& \frac{1}{K}\sum_{k'=1}^K \mathcal{R}(f_{\theta_{k'}^{t,J}}) - \frac{\mu}{2K}\sum_{k'=1}^K \|\theta_{k'}^{t,J} - \theta_g^t\|_2^2
    \end{split}
\end{equation}

Thus, we obtain:
\begin{equation}
    \begin{split}
        \mathbb{E}_{p_k^{\leq t}}[\mathcal{R}(f_{\theta_g^t})] 
        \leq& \frac{1}{K}\sum_{k'=1}^K \mathbb{E}_{p_k^{\leq t}}[\mathcal{R}(f_{\theta_{k'}^{t,J}})] - \frac{\mu}{2K}\sum_{k'=1}^K \mathbb{E}_{p_k^{\leq t}}\left[\|\theta_{k'}^{t,J} - \theta_g^t\|_2^2\right]
        \\
        =& \frac{1}{K} \mathbb{E}_{p_k^{\leq t}}[\mathcal{R}(f_{\theta_g^t})] + \frac{1}{K}\sum_{k'=1, k' \neq k}^K \mathbb{E}_{p_k^{\leq t}}[\mathcal{R}(f_{\theta_{k'}^{t,J}})] - \frac{\mu}{2K}\sum_{k'=1}^K \|\theta_{k'}^{t,J} - \theta_g^t\|_2^2
    \end{split}
\end{equation}

For the term $\mathbb{E}_{p_k^{\leq t}}[\mathcal{R}(f_{\theta_{k'}^{t,J}})]$ in the first part, Assumption 5 implies that the risk is bounded, i.e., $\mathcal{R}(f_{\theta_{k'}^{t,J}}) \leq 1$. Based on this, we proceed with the following derivation:
\begin{equation}
    \begin{split}
        \mathbb{E}_{p_k^{\leq t}}[\mathcal{R}(f_{\theta_{k'}^{t,J}})]
        =& \mathbb{E}_{p_{k'}^{\leq t}}[\mathcal{R}(f_{\theta_{k'}^{t,J}})] + \left(\mathbb{E}_{p_k^{\leq t}}[\mathcal{R}(f_{\theta_{k'}^{t,J}})] - \mathbb{E}_{p_{k'}^{\leq t}}[\mathcal{R}(f_{\theta_{k'}^{t,J}})]\right)
        \\
        \leq& \mathbb{E}_{p_{k'}^{\leq t}}[\mathcal{R}(f_{\theta_{k'}^{t,J}})] + \left| \mathbb{E}_{p_k^{\leq t}}[\mathcal{R}(f_{\theta_{k'}^{t,J}})] - \mathbb{E}_{p_{k'}^{\leq t}}[\mathcal{R}(f_{\theta_{k'}^{t,J}})] \right|
        \\
        \leq& \mathbb{E}_{p_{k'}^{\leq t}}[\mathcal{R}(f_{\theta_{k'}^{t,J}})] + \|p_k^{\leq t} - p_{k'}^{\leq t}\|_{\mathrm{TV}} \cdot \sup \left| \mathcal{R}(f_{\theta_{k'}^{t,J}}) \right|
        \\
        =& \mathbb{E}_{p_{k'}^{\leq t}}[\mathcal{R}(f_{\theta_{k'}^{t,J}})] + \|p_k^{\leq t} - p_{k'}^{\leq t}\|_{\mathrm{TV}}
    \end{split}
\end{equation}

Substituting the risk term of the global model on client $k$, we obtain:
\begin{equation}
    \begin{split}
        \mathbb{E}_{p_k^{\leq t}}[\mathcal{R}(f_{\theta_g^t})] 
        \leq& \frac{1}{K} \mathbb{E}_{p_k^{\leq t}}[\mathcal{R}(f_{\theta_g^t})] 
        + \frac{1}{K}\sum_{k'=1, k' \neq k}^K \left(\mathbb{E}_{p_{k'}^{\leq t}}[\mathcal{R}(f_{\theta_{k'}^{t,J}})] + \|p_k^{\leq t} - p_{k'}^{\leq t}\|_{\mathrm{TV}}\right) 
        - \frac{\mu}{2K}\sum_{k'=1}^K \|\theta_{k'}^{t,J} - \theta_g^t\|_2^2
        \\
        =& \frac{1}{K}\sum_{k'=1}^K \mathbb{E}_{p_{k'}^{\leq t}}[\mathcal{R}(f_{\theta_{k'}^{t,J}})]
        + \frac{1}{K}\sum_{k'=1, k' \neq k}^K \|p_k^{\leq t} - p_{k'}^{\leq t}\|_{\mathrm{TV}}
        - \frac{\mu}{2K}\sum_{k'=1}^K \|\theta_{k'}^{t,J} - \theta_g^t\|_2^2
    \end{split}
\end{equation}

Furthermore, for the expectation term $\mathbb{E}_{p_{k'}^{\leq t}}[\mathcal{R}(f_{\theta_{k'}^{t,J}})]$, we have:
\begin{equation}
    \begin{split}
        \mathbb{E}_{p_{k'}^{\leq t}}[\mathcal{R}(f_{\theta_{k'}^{t,J}})]
        =& \mathbb{E}_{p_{k'}^{\leq t}}[\mathcal{R}(f_{\theta_{k'}^{t,J}})] - \mathbb{E}_{p_{k'}^{\leq t}}[\mathcal{R}(f_{k',\mathrm{cum}}^{t,*})] + \mathbb{E}_{p_{k'}^{\leq t}}[\mathcal{R}(f_{k',\mathrm{cum}}^{t,*})]
        \\
        =& \mathrm{Regret}_{k'}^{\mathrm{local}}(t; \theta_g^{t-1}, \{\eta_{k'}^{t,j}\}_{j=0}^{J-1}) + \mathbb{E}_{p_{k'}^{\leq t}}[\mathcal{R}(f_{k',\mathrm{cum}}^{t,*})]
    \end{split}
\end{equation}

Substituting this into the above equation yields:
\begin{equation}
    \begin{split}
        \mathbb{E}_{p_k^{\leq t}}[\mathcal{R}(f_{\theta_g^t})] 
        \leq& \frac{1}{K}\sum_{k'=1}^K \left(\mathrm{Regret}_{k'}^{\mathrm{local}}(t; \theta_g^{t-1}, \{\eta_{k'}^{t,j}\}_{j=0}^{J-1}) + \mathbb{E}_{p_{k'}^{\leq t}}[\mathcal{R}(f_{k',\mathrm{cum}}^{t,*})]\right)
        \\
        &+ \frac{1}{K}\sum_{k'=1, k' \neq k}^K \|p_k^{\leq t} - p_{k'}^{\leq t}\|_{\mathrm{TV}}
        - \frac{\mu}{2K}\sum_{k'=1}^K \|\theta_{k'}^{t,J} - \theta_g^t\|_2^2
        \\
        =& \frac{1}{K}\sum_{k'=1}^K \mathrm{Regret}_{k'}^{\mathrm{local}}(t; \theta_g^{t-1}, \{\eta_{k'}^{t,j}\}_{j=0}^{J-1}) 
        + \frac{1}{K}\sum_{k'=1}^K \mathbb{E}_{p_{k'}^{\leq t}}[\mathcal{R}(f_{k',\mathrm{cum}}^{t,*})]
        \\
        &+ \frac{1}{K}\sum_{k'=1, k' \neq k}^K \|p_k^{\leq t} - p_{k'}^{\leq t}\|_{\mathrm{TV}}
        - \frac{\mu}{2K}\sum_{k'=1}^K \|\theta_{k'}^{t,J} - \theta_g^t\|_2^2
    \end{split}
\end{equation}

Subtracting $\mathbb{E}_{p_k^{\leq t}}[\mathcal{R}(f_{k,\mathrm{cum}}^{t,*})]$ from both sides yields:
\begin{equation}
    \begin{split}
        &\mathbb{E}_{p_k^{\leq t}}[\mathcal{R}(f_{\theta_g^t})] - \mathbb{E}_{p_k^{\leq t}}[\mathcal{R}(f_{k,\mathrm{cum}}^{t,*})]
        \\
        \leq& \frac{1}{K}\sum_{k'=1}^K \mathrm{Regret}_{k'}^{\mathrm{local}}(t; \theta_g^{t-1}, \{\eta_{k'}^{t,j}\}_{j=0}^{J-1}) 
        + \frac{1}{K}\sum_{k'=1}^K \left(\mathbb{E}_{p_{k'}^{\leq t}}[\mathcal{R}(f_{k',\mathrm{cum}}^{t,*})] - \mathbb{E}_{p_k^{\leq t}}[\mathcal{R}(f_{k,\mathrm{cum}}^{t,*})]\right)
        \\
        &+ \frac{1}{K}\sum_{k'=1, k' \neq k}^K \|p_k^{\leq t} - p_{k'}^{\leq t}\|_{\mathrm{TV}}
        - \frac{\mu}{2K}\sum_{k'=1}^K \|\theta_{k'}^{t,J} - \theta_g^t\|_2^2
        \\
        =& \frac{1}{K}\sum_{k'=1}^K \mathrm{Regret}_{k'}^{\mathrm{local}}(t; \theta_g^{t-1}, \{\eta_{k'}^{t,j}\}_{j=0}^{J-1}) 
        + \frac{1}{K}\sum_{k'=1, k' \neq k}^K \left(\mathbb{E}_{p_{k'}^{\leq t}}[\mathcal{R}(f_{k',\mathrm{cum}}^{t,*})] - \mathbb{E}_{p_k^{\leq t}}[\mathcal{R}(f_{k,\mathrm{cum}}^{t,*})]\right)
        \\
        &+ \frac{1}{K}\sum_{k'=1, k' \neq k}^K \|p_k^{\leq t} - p_{k'}^{\leq t}\|_{\mathrm{TV}}
        - \frac{\mu}{2K}\sum_{k'=1}^K \|\theta_{k'}^{t,J} - \theta_g^t\|_2^2
    \end{split}
\end{equation}

Substituting the definition of the regret of the global model on client $k$, $\mathrm{Regret}_k^{\mathrm{global}}(t; \theta_g^{t-1}, \mathcal{I}_t) \triangleq \mathbb{E}_{p_k^{\leq t}}[\mathcal{R}(f_{\theta_g^t}) - \mathcal{R}(f_{k,\mathrm{cum}}^{t,*})]$, yields:
\begin{equation}
    \begin{split}
        \mathrm{Regret}_k^{\mathrm{global}}(t; \theta_g^{t-1}, \mathcal{I}_t)
        \leq& \frac{1}{K}\sum_{k'=1}^K \mathrm{Regret}_{k'}^{\mathrm{local}}(t; \theta_g^{t-1}, \{\eta_{k'}^{t,j}\}_{j=0}^{J-1}) 
        \\
        &+ \frac{1}{K}\sum_{k'=1, k' \neq k}^K \left(\mathbb{E}_{p_{k'}^{\leq t}}[\mathcal{R}(f_{k',\mathrm{cum}}^{t,*})] - \mathbb{E}_{p_k^{\leq t}}[\mathcal{R}(f_{k,\mathrm{cum}}^{t,*})]\right)
        \\
        &+ \frac{1}{K}\sum_{k'=1, k' \neq k}^K \|p_k^{\leq t} - p_{k'}^{\leq t}\|_{\mathrm{TV}}
        - \frac{\mu}{2K}\sum_{k'=1}^K \|\theta_{k'}^{t,J} - \theta_g^t\|_2^2
    \end{split}
\end{equation}

For the term $\mathbb{E}_{p_{k'}^{\leq t}}[\mathcal{R}(f_{k',\mathrm{cum}}^{t,*})] - \mathbb{E}_{p_k^{\leq t}}[\mathcal{R}(f_{k,\mathrm{cum}}^{t,*})]$, we can rewrite it as follows:
\begin{equation}
    \begin{split}
        \mathbb{E}_{p_{k'}^{\leq t}}[\mathcal{R}(f_{k',\mathrm{cum}}^{t,*})] - \mathbb{E}_{p_k^{\leq t}}[\mathcal{R}(f_{k,\mathrm{cum}}^{t,*})]
        =& \mathbb{E}_{p_{k'}^{\leq t}}[\mathcal{R}(f_{k',\mathrm{cum}}^{t,*})] - \mathbb{E}_{p_{k}^{\leq t}}[\mathcal{R}(f_{k',\mathrm{cum}}^{t,*})] + \mathbb{E}_{p_{k}^{\leq t}}[\mathcal{R}(f_{k',\mathrm{cum}}^{t,*})] - \mathbb{E}_{p_k^{\leq t}}[\mathcal{R}(f_{k,\mathrm{cum}}^{t,*})]
        \\
        =& \left(\mathbb{E}_{p_{k'}^{\leq t}}[\mathcal{R}(f_{k',\mathrm{cum}}^{t,*})] - \mathbb{E}_{p_{k}^{\leq t}}[\mathcal{R}(f_{k',\mathrm{cum}}^{t,*})]\right) + \mathbb{E}_{p_{k}^{\leq t}}[\mathcal{R}(f_{k',\mathrm{cum}}^{t,*}) - \mathcal{R}(f_{k,\mathrm{cum}}^{t,*})]
    \end{split}
\end{equation}

For the term $\mathbb{E}_{p_{k'}^{\leq t}}[\mathcal{R}(f_{k',\mathrm{cum}}^{t,*})] - \mathbb{E}_{p_{k}^{\leq t}}[\mathcal{R}(f_{k',\mathrm{cum}}^{t,*})]$, following a similar argument as before, we have:
\begin{equation}
    \begin{split}
        \mathbb{E}_{p_{k'}^{\leq t}}[\mathcal{R}(f_{k',\mathrm{cum}}^{t,*})] - \mathbb{E}_{p_{k}^{\leq t}}[\mathcal{R}(f_{k',\mathrm{cum}}^{t,*})]
        \leq& \left|\mathbb{E}_{p_{k'}^{\leq t}}[\mathcal{R}(f_{k',\mathrm{cum}}^{t,*})] - \mathbb{E}_{p_{k}^{\leq t}}[\mathcal{R}(f_{k',\mathrm{cum}}^{t,*})]\right|
        \\
        \leq& \|p_{k'}^{\leq t} - p_k^{\leq t}\|_{\mathrm{TV}} \cdot \sup |\mathcal{R}(f_{k',\mathrm{cum}}^{t,*})|
        \\
        \leq& \|p_{k'}^{\leq t} - p_k^{\leq t}\|_{\mathrm{TV}}
    \end{split}
\end{equation}

Thus, we obtain:
\begin{equation}
    \mathbb{E}_{p_{k'}^{\leq t}}[\mathcal{R}(f_{k',\mathrm{cum}}^{t,*})] - \mathbb{E}_{p_k^{\leq t}}[\mathcal{R}(f_{k,\mathrm{cum}}^{t,*})]
    \leq \mathbb{E}_{p_{k}^{\leq t}}[\mathcal{R}(f_{k',\mathrm{cum}}^{t,*}) - \mathcal{R}(f_{k,\mathrm{cum}}^{t,*})]
    + \|p_{k'}^{\leq t} - p_k^{\leq t}\|_{\mathrm{TV}}
\end{equation}

Substituting the expression for the regret of the global model on client $k$ yields:
\begin{equation}
    \begin{split}
        \mathrm{Regret}_k^{\mathrm{global}}(t; \theta_g^{t-1}, \mathcal{I}_t)
        \leq& \frac{1}{K}\sum_{k'=1}^K \mathrm{Regret}_{k'}^{\mathrm{local}}(t; \theta_g^{t-1}, \{\eta_{k'}^{t,j}\}_{j=0}^{J-1}) 
        \\
        &+ \frac{1}{K}\sum_{k'=1, k' \neq k}^K \left(\mathbb{E}_{p_{k}^{\leq t}}[\mathcal{R}(f_{k',\mathrm{cum}}^{t,*}) - \mathcal{R}(f_{k,\mathrm{cum}}^{t,*})] + \|p_{k'}^{\leq t} - p_k^{\leq t}\|_{\mathrm{TV}}\right)
        \\
        &+ \frac{1}{K}\sum_{k'=1, k' \neq k}^K \|p_k^{\leq t} - p_{k'}^{\leq t}\|_{\mathrm{TV}}
        - \frac{\mu}{2K}\sum_{k'=1}^K \|\theta_{k'}^{t,J} - \theta_g^t\|_2^2
        \\
        =& \frac{1}{K}\sum_{k'=1}^K \mathrm{Regret}_{k'}^{\mathrm{local}}(t; \theta_g^{t-1}, \{\eta_{k'}^{t,j}\}_{j=0}^{J-1}) 
        + \frac{1}{K}\sum_{k'=1, k' \neq k}^K \mathbb{E}_{p_{k}^{\leq t}}[\mathcal{R}(f_{k',\mathrm{cum}}^{t,*}) - \mathcal{R}(f_{k,\mathrm{cum}}^{t,*})]
        \\
        &+ \frac{2}{K}\sum_{k'=1, k' \neq k}^K \|p_k^{\leq t} - p_{k'}^{\leq t}\|_{\mathrm{TV}}
        - \frac{\mu}{2K}\sum_{k'=1}^K \|\theta_{k'}^{t,J} - \theta_g^t\|_2^2
    \end{split}
\end{equation}

For the total variation (TV) distance term between the cumulative distributions, $\|p_k^{\leq t} - p_{k'}^{\leq t}\|_{\mathrm{TV}}$, given that different clients share the same conditional probability density function $p(x|c)$, we define the cumulative sample proportion of class $c$ for client $k$ up to FL round $t$ as $\pi_k^{\leq t}(c) \triangleq \left\{\sum_{\tau=1}^t n_k^\tau \mathbb{P}_{p_k^\tau}(Y=c)\right\} / \left\{\sum_{\tau=1}^t n_k^\tau\right\}$. 
Clearly, this proportion forms a probability distribution over the global class set $\mathcal{C}$, satisfying $\sum_{c \in \mathcal{C}} \pi_k^{\leq t}(c) = 1$ and $\pi_k^{\leq t}(c) \geq 0$. Consequently, the cumulative distributions of clients $k$ and $k'$ up to FL round $t$ can be decomposed into a mixture of classes:
\begin{equation}
    p_k^{\leq t}(x) = \sum_{c \in \mathcal{C}} \pi_k^{\leq t}(c) p(x|c); \quad p_{k'}^{\leq t}(x) = \sum_{c \in \mathcal{C}} \pi_{k'}^{\leq t}(c) p(x|c)
\end{equation}

By the definition of the total variation distance, we obtain:
\begin{equation}
    \begin{split}
        \|p_k^{\leq t} - p_{k'}^{\leq t}\|_{\mathrm{TV}} 
        =& \frac{1}{2} \int_{\mathcal{X}} \left| \sum_{c \in \mathcal{C}} \left( \pi_k^{\leq t}(c) - \pi_{k'}^{\leq t}(c) \right) p(x|c) \right| \mathrm{d}x
        \\
        \leq& \frac{1}{2} \sum_{c \in \mathcal{C}} \left| \int_{\mathcal{X}} \left( \pi_k^{\leq t}(c) - \pi_{k'}^{\leq t}(c) \right) p(x|c)\mathrm{d}x \right| 
        \\
        \leq& \frac{1}{2} \sum_{c \in \mathcal{C}} \left| \pi_k^{\leq t}(c) - \pi_{k'}^{\leq t}(c)\right| \int_{\mathcal{X}} p(x|c)\mathrm{d}x
        \\
        =& \frac{1}{2} \sum_{c \in \mathcal{C}} \left| \pi_k^{\leq t}(c) - \pi_{k'}^{\leq t}(c)\right| 
    \end{split}
\end{equation}

For the term $\left| \pi_k^{\leq t}(c) - \pi_{k'}^{\leq t}(c)\right|$ corresponding to class $c$, similar to the proof of Lemma 2, let $n_{k}^\tau$ be the amount of local data received by client $k$ at the $\tau$-th FL round, and define the mixing weight of the $t$-th FL round data distribution $p_k^{t}$ as $\omega_{k}^{t} \triangleq n_{k}^t / \sum_{\tau=1}^t n_{k}^\tau$, which satisfies $\sum_{\tau=1}^t \omega_k^\tau = 1$. 
Translating the cumulative distribution recurrence $p_k^{\leq t}(x) = (1 - \omega_{k}^{t}) p_k^{\leq t-1}(x) + \omega_{k}^{t} p_k^{t}(x)$ to class $c$, we obtain the following recurrence:
\begin{equation}
    \pi_k^{\leq t}(c) = (1 - \omega_k^t) \pi_k^{\leq t-1}(c) + \omega_k^t \pi_k^t(c); 
    \qquad \pi_{k'}^{\leq t}(c) = (1 - \omega_{k'}^t) \pi_{k'}^{\leq t-1}(c) + \omega_{k'}^t \pi_{k'}^t(c)
\end{equation}

In particular, if client $k$ does not receive any data samples of class $c$ in the $t$-th round, its sample proportion $\pi_k^t(c) = 0$, in which case the above exactly holds. Consequently, we obtain:
\begin{equation}
    \begin{split}
        \pi_k^{\leq t}(c) - \pi_{k'}^{\leq t}(c)
        =& (1 - \omega_k^t) \pi_k^{\leq t-1}(c) + \omega_k^t \pi_k^t(c) - (1 - \omega_{k'}^t) \pi_{k'}^{\leq t-1}(c) - \omega_{k'}^t \pi_{k'}^t(c)
        \\
        =& (1 - \omega_k^t) \pi_k^{\leq t-1}(c) - (1 - \omega_k^t) \pi_{k'}^{\leq t-1}(c) + (1 - \omega_k^t) \pi_{k'}^{\leq t-1}(c)- (1 - \omega_{k'}^t) \pi_{k'}^{\leq t-1}(c)
        \\
        &+ \omega_k^t \pi_k^t(c) - \omega_{k'}^t \pi_{k'}^t(c)
        \\
        =& (1 - \omega_k^t) (\pi_k^{\leq t-1}(c) - \pi_k^{\leq t-1}(c)) + (\omega_{k'}^t - \omega_k^t) \pi_{k'}^{\leq t-1}(c) + \omega_k^t \pi_k^t(c) - \omega_{k'}^t \pi_{k'}^t(c)
        \\
        =& (1 - \omega_k^t) (\pi_k^{\leq t-1}(c) - \pi_k^{\leq t-1}(c)) + (\omega_{k'}^t - \omega_k^t) \pi_{k'}^{\leq t-1}(c) 
        \\
        &+ \omega_k^t \pi_k^t(c) - \omega_k^t \pi_{k'}^t(c) + \omega_k^t \pi_{k'}^t(c) - \omega_{k'}^t \pi_{k'}^t(c)
        \\
        =& (1 - \omega_k^t) (\pi_k^{\leq t-1}(c) - \pi_k^{\leq t-1}(c)) + (\omega_{k'}^t - \omega_k^t) \pi_{k'}^{\leq t-1}(c) + \omega_k^t \left( \pi_k^t(c) - \pi_{k'}^t(c) \right) + (\omega_k^t - \omega_{k'}^t) \pi_{k'}^t(c)
    \end{split}
\end{equation}

Taking the absolute value on both sides yields:
\begin{equation}
    \begin{split}
        \left| \pi_k^{\leq t}(c) - \pi_{k'}^{\leq t}(c) \right| 
        \leq& (1 - \omega_k^t) \left| \pi_k^{\leq t-1}(c) - \pi_{k'}^{\leq t-1}(c) \right| + \left|\omega_{k'}^t - \omega_k^t\right| \pi_{k'}^{\leq t-1}(c) 
        \\
        &+ \omega_k^t \left| \pi_k^t(c) - \pi_{k'}^t(c) \right| + \left| \omega_k^t - \omega_{k'}^t \right| \pi_{k'}^t(c)
        \\
        =& (1 - \omega_k^t) \left| \pi_k^{\leq t-1}(c) - \pi_{k'}^{\leq t-1}(c) \right| + \left|\omega_{k'}^t - \omega_k^t\right| (\pi_{k'}^{\leq t - 1}(c) + \pi_{k'}^{t}(c)) 
        \\
        &+ \omega_k^t \left| \pi_k^t(c) - \pi_{k'}^t(c) \right|
    \end{split}
\end{equation}

Summing both sides over all $c \in \mathcal{C}$ yields:
\begin{equation}
    \begin{split}
        \sum_{c \in \mathcal{C}} \left| \pi_k^{\leq t}(c) - \pi_{k'}^{\leq t}(c)\right| 
        \leq& (1 - \omega_k^t) \sum_{c \in \mathcal{C}} \left| \pi_k^{\leq t-1}(c) - \pi_{k'}^{\leq t-1}(c) \right| + \left|\omega_{k'}^t - \omega_k^t\right| \sum_{c \in \mathcal{C}} \left(\pi_{k'}^{\leq t-1}(c) + \pi_{k'}^t(c)\right)
        \\
        &+ \omega_k^t \sum_{c \in \mathcal{C}} \left| \pi_k^t(c) - \pi_{k'}^t(c) \right|
        \\
        =& (1 - \omega_k^t) \sum_{c \in \mathcal{C}} \left| \pi_k^{\leq t-1}(c) - \pi_{k'}^{\leq t-1}(c) \right| + 2 \left|\omega_{k'}^t - \omega_k^t\right| + \omega_k^t \sum_{c \in \mathcal{C}} \left| \pi_k^t(c) - \pi_{k'}^t(c) \right|
    \end{split}
\end{equation}

This yields a recurrence relation for $\sum_{c \in \mathcal{C}} \left| \pi_k^{\leq t}(c) - \pi_{k'}^{\leq t}(c)\right|$. 
For the third term $\sum_{c \in \mathcal{C}} \left| \pi_k^t(c) - \pi_{k'}^t(c) \right|$ above, we partition the global class set $\mathcal{C}$ into four mutually exclusive subsets based on the classes $\mathcal{C}_k^t$ and $\mathcal{C}_{k'}^t$ observed by clients $k$ and $k'$ in FL round $t$: $\mathcal{C}_{k\cap k'}^t \triangleq \mathcal{C}_k^t \cap \mathcal{C}_{k'}^t$, $\mathcal{C}_{k,\textrm{only}}^t \triangleq \mathcal{C}_k^t \setminus \mathcal{C}_{k'}^t$, $\mathcal{C}_{k',\textrm{only}}^t \triangleq \mathcal{C}_{k'}^t \setminus \mathcal{C}_k^t$, and $\mathcal{C}_{\textrm{others}}^t \triangleq \mathcal{C} \setminus (\mathcal{C}_k^t \cup \mathcal{C}_{k'}^t)$. 
Consequently, this summation can be decomposed as:
\begin{equation}
    \begin{split}
        \sum_{c \in \mathcal{C}} \left| \pi_k^t(c) - \pi_{k'}^t(c) \right|
        =& \sum_{c \in \mathcal{C}_{k\cap k'}^t} \left| \pi_k^t(c) - \pi_{k'}^t(c) \right|
        + \sum_{c \in \mathcal{C}_{k,\textrm{only}}^t} \left| \pi_k^t(c) - \pi_{k'}^t(c) \right|
        \\
        &+ \sum_{c \in \mathcal{C}_{k',\textrm{only}}^t} \left| \pi_k^t(c) - \pi_{k'}^t(c) \right|
        + \sum_{c \in \mathcal{C}_{\textrm{others}}^t} \left| \pi_k^t(c) - \pi_{k'}^t(c) \right|
        \\
        =& \sum_{c \in \mathcal{C}_{k\cap k'}^t} \left| \pi_k^t(c) - \pi_{k'}^t(c) \right| 
        + \sum_{c \in \mathcal{C}_{k,\textrm{only}}^t} \pi_k^t(c)
        + \sum_{c \in \mathcal{C}_{k',\textrm{only}}^t} \pi_{k'}^t(c) 
        + 0
    \end{split}
\end{equation}

Substituting the identities $\sum_{c \in \mathcal{C}_{k,\textrm{only}}^t} \pi_k^t(c) = 1 - \sum_{c \in \mathcal{C}_{k\cap k'}^t} \pi_k^t(c)$ and $\sum_{c \in \mathcal{C}_{k',\textrm{only}}^t} \pi_{k'}^t(c) = 1 - \sum_{c \in \mathcal{C}_{k\cap k'}^t} \pi_{k'}^t(c)$ into the above equation yields:
\begin{equation}
    \begin{split}
        \sum_{c \in \mathcal{C}} \left| \pi_k^t(c) - \pi_{k'}^t(c) \right|
        =& \sum_{c \in \mathcal{C}_{k\cap k'}^t} \left| \pi_k^t(c) - \pi_{k'}^t(c) \right| 
        + \left(1 - \sum_{c \in \mathcal{C}_{k\cap k'}^t} \pi_k^t(c)\right) 
        + \left(1 - \sum_{c \in \mathcal{C}_{k\cap k'}^t} \pi_{k'}^t(c)\right)
        \\
        =& 2 + \sum_{c \in \mathcal{C}_{k\cap k'}^t} \left(\left| \pi_k^t(c) - \pi_{k'}^t(c) \right| - \pi_k^t(c) - \pi_{k'}^t(c)  \right)
        \\
        =& 2 - \sum_{c \in \mathcal{C}_{k\cap k'}^t} \left(\pi_k^t(c) + \pi_{k'}^t(c) - \left|\pi_k^t(c) - \pi_{k'}^t(c) \right| \right)
        \\
        =& 2 - \sum_{c \in \mathcal{C}_{k\cap k'}^t} 2\min\left(\pi_k^t(c), \pi_{k'}^t(c)\right)
        = 2\left(1 - \sum_{c \in \mathcal{C}_{k\cap k'}^t} \min\left(\pi_k^t(c), \pi_{k'}^t(c)\right)\right)
    \end{split}
\end{equation}

By Assumption 7 and the normalization of probability measures, the infimum of the probability mass for any single class is $1/C_{\max}$, which is achieved when the probability mass is uniformly distributed over the $C_{\max}$ global classes. 
Therefore, for any class $c \in \mathcal{C}_{k\cap k'}^t$, it holds that $\min\left(\pi_k^t(c), \pi_{k'}^t(c)\right) \geq 1/C_{\max}$. Consequently, we obtain:
\begin{equation}
    \sum_{c \in \mathcal{C}} \left| \pi_k^t(c) - \pi_{k'}^t(c) \right|
    \leq 2\left(1 - \left|\mathcal{C}_{k\cap k'}^t\right| \cdot \frac{1}{C_{\max}}\right)
    = 2\left(1 - \frac{\left|\mathcal{C}_k^t \cap \mathcal{C}_{k'}^t\right|}{C_{\max}}\right)
\end{equation}

Substituting this back into the original inequality yields:
\begin{equation}
    \sum_{c \in \mathcal{C}} \left| \pi_k^{\leq t}(c) - \pi_{k'}^{\leq t}(c)\right| 
    \leq (1 - \omega_k^t) \sum_{c \in \mathcal{C}} \left| \pi_k^{\leq t-1}(c) - \pi_{k'}^{\leq t-1}(c) \right| + 2 \left|\omega_{k'}^t - \omega_k^t\right| + 2 \omega_k^t \left(1 - \frac{\left|\mathcal{C}_k^t \cap \mathcal{C}_{k'}^t\right|}{C_{\max}}\right)
\end{equation}

Let $n_{\textrm{max}} \triangleq \max_{k' \in \mathcal{K},\tau \leq T}(n_{k'}^\tau)$ and $n_{\textrm{min}} \triangleq \min_{k' \in \mathcal{K},\tau \leq T}(n_{k'}^\tau)$ be the maximum and minimum per-round data sizes across all $K$ clients over the $T$ FL rounds, respectively, and let $\nu = n_{\min}/n_{\max} \leq 1$. 
We then obtain the following bounds: $\omega_{k}^{t} \leq \frac{n_{\textrm{max}}}{t \cdot n_{\textrm{min}}} = \frac{1}{\nu t}$, $\left|\omega_{k'}^t - \omega_k^t\right| \leq \omega_{k'}^t + \omega_k^t \leq \frac{2}{\nu t}$, and $1 - \omega_k^t \leq 1 - \frac{n_{\textrm{min}}}{t \cdot n_{\textrm{max}}} = 1 - \frac{\nu}{t}$. Substituting these bounds back into the previous expression yields:
\begin{equation}
    \begin{split}
        \sum_{c \in \mathcal{C}} \left| \pi_k^{\leq t}(c) - \pi_{k'}^{\leq t}(c)\right| 
        \leq& (1 - \frac{\nu}{t}) \sum_{c \in \mathcal{C}} \left| \pi_k^{\leq t-1}(c) - \pi_{k'}^{\leq t-1}(c) \right| + \frac{4}{\nu t} + \frac{2}{\nu t} \left(1 - \frac{\left|\mathcal{C}_k^t \cap \mathcal{C}_{k'}^t\right|}{C_{\max}}\right)
        \\
        =& (1 - \frac{\nu}{t}) \sum_{c \in \mathcal{C}} \left| \pi_k^{\leq t-1}(c) - \pi_{k'}^{\leq t-1}(c) \right| + \frac{2}{\nu t} \left(3 - \frac{\left|\mathcal{C}_k^t \cap \mathcal{C}_{k'}^t\right|}{C_{\max}}\right)
    \end{split}
\end{equation}

Let $\Delta_{k,k'}^t \triangleq \sum_{c \in \mathcal{C}} \left| \pi_k^{\leq t}(c) - \pi_{k'}^{\leq t}(c)\right|$ and $\rho_{k,k'}^{t} \triangleq \left|\mathcal{C}_k^t \cap \mathcal{C}_{k'}^t\right| / C_{\max}$. Recursively expanding the above recurrence yields:
\begin{equation}
    \begin{split}
        \Delta_{k,k'}^t 
        \leq& (1 - \frac{\nu}{t}) \Delta_{k,k'}^{t-1} + \frac{2}{\nu t} \left(3 - \rho_{k,k'}^{t}\right)
        \\
        \leq& \left(1 - \frac{\nu}{t}\right)\left(1 - \frac{\nu}{(t-1)}\right) \Delta_{k,k'}^{t-2} + \left(1 - \frac{\nu}{t}\right) \frac{2}{\nu (t-1)} \left(3 - \rho_{k,k'}^{t-1}\right) + \frac{2}{\nu t} \left(3 - \rho_{k,k'}^{t}\right)
        \\
        \leq& \ldots
        \\
        \leq& \left[ \prod_{\tau=t-m+1}^{t} \left(1 - \frac{\nu}{\tau}\right) \right] \Delta_{k,k'}^{t-m}
        + \sum_{\tau=t-m+1}^{t} \left[ \prod_{i=\tau+1}^{t} \left(1 - \frac{\nu}{i}\right) \right] \left( \frac{2}{\nu\tau} \left(3 - \rho_{k,k'}^{\tau}\right) \right)
        \\
        \leq& \ldots
        \\
        \leq& \left( \prod_{\tau=2}^t \left(1 - \frac{\nu}{\tau}\right) \right) \Delta_{k,k'}^{1} + \sum_{\tau=2}^t \left[ \prod_{i=\tau+1}^t \left(1 - \frac{\nu}{i}\right) \right] \left( \frac{2}{\nu \tau} \left(3 - \rho_{k,k'}^{\tau}\right) \right)
    \end{split}
\end{equation}

For the product term $\prod_{i=\tau+1}^t \left(1 - \frac{\nu}{i}\right)$, applying the inequality $1 - x \leq \exp(-x)$ and the integral bound $\frac{1}{i} \geq \int_{i}^{i+1} \frac{1}{x} \mathrm{d}x$ for all $i > 0$ yields:
\begin{equation}
    \begin{split}
        \prod_{i=\tau+1}^t \left(1 - \frac{\nu}{i}\right) 
        \leq& \prod_{i=\tau+1}^t \exp\left(-\frac{\nu}{i}\right)
        = \exp\left( -\nu \sum_{i=\tau+1}^t \frac{1}{i} \right)
        \\
        \leq& \exp\left( -\nu \sum_{i=\tau+1}^t \int_{i}^{i+1} \frac{1}{x} \mathrm{d}x \right)
        = \exp\left( -\nu \int_{\tau+1}^{t+1} \frac{1}{x} \mathrm{d}x \right)
        \\
        =& \exp\left( -\nu \ln\left(\frac{t+1}{\tau+1}\right) \right)
        = \left( \frac{t+1}{\tau+1} \right)^{-\nu}
        = \left( \frac{\tau+1}{t+1} \right)^{\nu}
    \end{split}
\end{equation}

Substituting the recursive expansion of $\Delta_{k,k'}^t$ and utilizing $\rho_{k,k'}^{\tau} \geq 0$ yields:
\begin{equation}
    \begin{split}
        \Delta_{k,k'}^t 
        \leq& \left( \frac{2}{t+1} \right)^{\nu} \Delta_{k,k'}^{1} + \sum_{\tau=2}^t \left( \frac{\tau+1}{t+1} \right)^{\nu} \left( \frac{2}{\nu \tau} \left(3 - \rho_{k,k'}^{\tau}\right) \right)
        \\
        \leq& \left( \frac{2}{t+1} \right)^{\nu} \Delta_{k,k'}^{1} + \frac{6}{\nu} \sum_{\tau=2}^t \left( \frac{\tau+1}{t+1} \right)^{\nu} \frac{1}{\tau}
    \end{split}
\end{equation}

For the summation term $\sum_{\tau=2}^t \left( \frac{\tau+1}{t+1} \right)^{\nu} \frac{1}{\tau}$, consider the function $g(x) = \frac{(x+1)^{\nu}}{(t+1)^{\nu}} \cdot \frac{1}{x}$. Utilizing $\nu \leq 1$ and the generalized Bernoulli's inequality, we obtain:
\begin{equation}
    \begin{split}
        g(x) 
        =& \frac{1}{(t+1)^{\nu}} \cdot \frac{(x+1)^{\nu}}{x} 
        = \frac{1}{(t+1)^{\nu}} x^{{\nu}-1} \left(1+\frac{1}{x}\right)^{\nu}
        \\
        \leq& \frac{1}{(t+1)^{\nu}} x^{{\nu}-1} \left(1 + \frac{{\nu}}{x}\right) 
        = \frac{1}{(t+1)^{\nu}} \left[ x^{{\nu}-1} + {\nu} x^{{\nu}-2} \right]
    \end{split}
\end{equation}

Since $\nu \leq 1$, the function $x^{\nu-1} + \nu x^{\nu-2}$ is strictly decreasing on the interval $[2,t]$. Utilizing this monotonicity for integral bounding, the summation satisfies:
\begin{equation}
    \begin{split}
        \sum_{\tau=2}^t g(\tau) 
        \leq& g(2) + \int_2^t \frac{1}{(t+1)^{\nu}} \left[ x^{{\nu}-1} + {\nu} x^{{\nu}-2} \right] \mathrm{d}x
        \\
        =& g(2) + \frac{1}{(t+1)^{\nu}} \int_2^t x^{{\nu}-1} \mathrm{d}x + \frac{\nu}{(t+1)^{\nu}} \int_2^t x^{{\nu}-2} \mathrm{d}x
        \\
        =& g(2) + \frac{1}{(t+1)^{\nu}} \left[ \frac{x^{\nu}}{\nu} \right]_2^t + \frac{\nu}{(t+1)^{\nu}} \left[ \frac{x^{{\nu}-1}}{{\nu}-1} \right]_2^t
        \\
        =& g(2) + \frac{1}{(t+1)^{\nu}} \cdot \frac{t^{\nu} - 2^{\nu}}{\nu} + \frac{\nu}{(t+1)^{\nu}} \cdot \frac{t^{{\nu}-1} - 2^{{\nu}-1}}{{\nu}-1}
        \\
        \leq& g(2) + \frac{t^{\nu}}{(t+1)^{\nu}} \cdot \frac{1}{\nu} + \frac{{\nu} 2^{{\nu}-1} }{(1 - {\nu}) (t+1)^{\nu}}
        \\
        \leq& g(2) + \frac{1}{ \nu} + \frac{{\nu} 2^{{\nu}-1} }{(1 - {\nu}) (t+1)^{\nu}}
    \end{split}
\end{equation}

Furthermore, substituting $g(2) = {(3/2)^{\nu}}/{2(t+1)^{\nu}}$ yields:
\begin{equation}
    \begin{split}
        \sum_{\tau=2}^t g(\tau) 
        \leq& \frac{1}{{\nu}} + \frac{{\nu} 2^{{\nu}-1} }{(1 - {\nu}) (t+1)^{\nu}} + \frac{(3/2)^{\nu}}{2(t+1)^{\nu}}
        \\
        =& \frac{1}{{\nu}} + \frac{1}{(t+1)^{\nu}} \left[\frac{(3/2)^{\nu}}{2} + \frac{{\nu} 2^{{\nu}-1} }{1 - {\nu}}\right]
    \end{split}
\end{equation}

Substituting the recursive expansion of $\Delta_{k,k'}^t$ yields:
\begin{equation}
    \begin{split}
        \Delta_{k,k'}^t 
        \leq& \left( \frac{2}{t+1} \right)^{\nu} \Delta_{k,k'}^{1} + \frac{6}{\nu} \left(\frac{1}{\nu} + \frac{1}{(t+1)^{\nu}} \left[\frac{(3/2)^{\nu}}{2} + \frac{{\nu} 2^{{\nu}-1} }{1 - {\nu}}\right]\right)
        \\
        =& \left( \frac{2}{t+1} \right)^{\nu} \Delta_{k,k'}^{1} + \frac{6}{\nu (t+1)^{\nu}} \left[\frac{(3/2)^{\nu}}{2} + \frac{{\nu} 2^{{\nu}-1} }{1 - {\nu}}\right] + \frac{6}{\nu^2}
    \end{split}
\end{equation}

Thus, we obtain the upper bound on the total variation distance between the cumulative data distributions of clients $k$ and $k'$ at FL round $t$:
\begin{equation}
    \begin{split}
        \|p_k^{\leq t} - p_{k'}^{\leq t}\|_{\mathrm{TV}} 
        \leq& \frac{1}{2} \sum_{c \in \mathcal{C}} \left| \pi_k^{\leq t}(c) - \pi_{k'}^{\leq t}(c)\right| 
        \\
        \leq& \left( \frac{2}{t+1} \right)^{\nu} + \frac{3}{\nu (t+1)^{\nu}} \left[\frac{(3/2)^{\nu}}{2} + \frac{{\nu} 2^{{\nu}-1} }{1 - {\nu}}\right] + \frac{3}{\nu^2}
        \\
        =& \frac{1}{(t+1)^{\nu}} \left(2^{\nu} + \frac{3}{\nu} \left[\frac{(3/2)^{\nu}}{2} + \frac{{\nu} 2^{{\nu}-1} }{1 - {\nu}}\right] \right) + \frac{3}{\nu^2}
    \end{split}
\end{equation}

Thus, we define the global upper bound on the cross-client TV distance at FL round $t$ as:
\begin{equation}
    \Delta^{\mathrm{TV}}(t, \nu) \triangleq \frac{1}{(t+1)^{\nu}} \left(2^{\nu} + \frac{3}{\nu} \left[\frac{(3/2)^{\nu}}{2} + \frac{{\nu} 2^{{\nu}-1} }{1 - {\nu}}\right] \right) + \frac{3}{\nu^2}
\end{equation}

Substituting the expression for the regret of the global model on client $k$ yields:
\begin{equation}
    \begin{split}
        \mathrm{Regret}_k^{\mathrm{global}}(t; \theta_g^{t-1}, \mathcal{I}_t)
        \leq& \frac{1}{K}\sum_{k'=1}^K \mathrm{Regret}_{k'}^{\mathrm{local}}(t; \theta_g^{t-1}, \{\eta_{k'}^{t,j}\}_{j=0}^{J-1}) 
        + \frac{1}{K}\sum_{k'=1, k' \neq k}^K \mathbb{E}_{p_{k}^{\leq t}}[\mathcal{R}(f_{k',\mathrm{cum}}^{t,*}) - \mathcal{R}(f_{k,\mathrm{cum}}^{t,*})]
        \\
        &+ \frac{2(K-1)}{K} \Delta^{\mathrm{TV}}(t, \nu) - \frac{\mu}{2K}\sum_{k'=1}^K \|\theta_{k'}^{t,J} - \theta_g^t\|_2^2
    \end{split}
\end{equation}

where $\nu = n_{\min}/n_{\max} \leq 1$, and $n_{\textrm{max}} \triangleq \max_{k' \in \mathcal{K},\tau \leq T}(n_{k'}^\tau)$ and $n_{\textrm{min}} \triangleq \min_{k' \in \mathcal{K},\tau \leq T}(n_{k'}^\tau)$ denote the maximum and minimum per-round data sizes across all clients over the $T$ FL rounds, respectively. This completes the proof of Lemma 3.

\newpage
\subsection{Proof of Theorem 3}
\label{Appendix B.7}
\setcounter{theorem}{2} 
\begin{theorem}[Regret Gap between Local and Global Models in FedKACE]
    Under our problem formulation and standard assumptions (see \hyperref[Appendix B.1]{Appendix B.1}), 
    let the optimal model for client $k$ under the local cumulative data distribution $p_k^{\leq t}$ be $f^{t,*}_{k,\mathrm{cum}} \triangleq \arg\min_{\theta \in \Omega_\theta} \mathbb{E}_{p_k^{\leq t}}[\mathcal{R}(f_\theta)]$, 
    and define $\mathrm{Regret}_k^{t}(\theta) \triangleq \mathbb{E}_{p_k^{\leq t}}\left[\mathcal{R}(f_{\theta}) - \mathcal{R}(f^{t,*}_{k,\mathrm{cum}})\right]$ as the regret of model $\theta$ on client $k$ at FL round $t$,  
    the regret gap between the local model $\theta_k^{t,J}$ and the global model $\theta_g^t$ satisfies:
    \begin{equation}
      \begin{split}
          \underset{k \in \mathcal{K}}{\mathbb{E}}
          \left[ \mathrm{Regret}_k^{t}(\theta_g^t) - \mathrm{Regret}_k^{t}(\theta_k^{t,J}) \right]
          \leq \frac{4(K-1)}{K} \Delta^{\mathrm{TV}}(t, \nu) - \frac{\mu}{2K}\sum_{k=1}^K \|\theta_k^{t,J} - \theta_g^t\|_2^2
      \end{split}
    \end{equation}
    where:
    \begin{equation}
        \Delta^{\mathrm{TV}}(t, \nu) \triangleq \frac{1}{(t+1)^{\nu}} \left(2^{\nu} + \frac{3}{\nu} \left[\frac{(3/2)^{\nu}}{2} + \frac{{\nu} 2^{{\nu}-1} }{1 - {\nu}}\right] \right) + \frac{3}{\nu^2}
    \end{equation}
    is the global upper bound of the total variation (TV) distance between the cumulative distributions of any pair of clients $\|p_k^{\leq t} - p_{k'}^{\leq t}\|_{\mathrm{TV}}, \forall k, k' \in \mathcal{K}$ at FL round $t$, the derivation of which is provided in Lemma 3 (\hyperref[Appendix B.6]{Appendix B.6}); $\nu = n_{\min}/n_{\max}$ is the ratio of the minimum to the maximum per-round data size, defined as $n_{\min} \triangleq \min_{k \in \mathcal{K}, \tau \leq T}(n_k^\tau)$ and $n_{\max} \triangleq \max_{k \in \mathcal{K}, \tau \leq T}(n_k^\tau)$.
    For small $t$, $\Delta^{\mathrm{TV}}(t, \nu)$ dominates, indicating that the local model $\theta_k^{t,J}$ is superior; for large $t$, the negative model discrepancy term dominates, favoring the global model $\theta_g^t$.
\end{theorem}

\textbf{Proof}:

Following the proofs of Lemmas 2 and 3, we introduce functionals to uniformly represent the local and global model regrets. Specifically, the local regret functional is defined as $\mathrm{Regret}_k^{\mathrm{local}}(t; \theta_k^{t,0}, \{\eta_k^{t,j}\}_{j=0}^{J-1}) \triangleq \mathrm{Regret}_k^{t}(\theta_k^{t,J})$, where $\theta_k^{t,0}$ and $\{\eta_k^{t,j}\}_{j=0}^{J-1}$ are the initial model and the sequence of local learning rates at FL round $t$, respectively; meanwhile, the global regret functional is defined as $\mathrm{Regret}_k^{\mathrm{global}}(t; \theta_g^{t-1}, \mathcal{I}_t) \triangleq \mathrm{Regret}_k^{t}(\theta_g^{t})$, where $\theta_g^{t-1}$ and $\mathcal{I}_t \triangleq \left\{ \{\eta_{k'}^{t,j}\}_{j=0}^{J-1} \mid \forall k' \in \mathcal{K}\right\}$ are the global model at FL round $t-1$ and the set of learning rate sequences for all clients at FL round $t$, respectively.

First, by utilizing the recursive bound on the local model regret from Lemma 2 and unrolling it to $\tau = 2$, we obtain:
\begin{equation}
    \begin{split}
        \mathrm{Regret}_k^{\mathrm{local}}(t; \theta_k^{t,0}, \{\eta_k^{t,j}\}_{j=0}^{J-1})
        \leq& \left(1 - \omega_{k}^{t}\right) \mathrm{Regret}_k^{\mathrm{local}}(t-1; \theta_k^{t-1,0}, \{\eta_k^{t-1,j}\}_{j=0}^{J-1})
        + \omega_{k}^{t} \cdot U_k(t; \theta_k^{t,0}, \{\eta_k^{t,j}\}_{j=0}^{J-1})
        \\
        &+ \left(1 - \omega_{k}^{t}\right) \mathbb{E}_{p_k^{\leq t-1}}\left[ \mathcal{R}(f_{\theta_k^{t,J}}) - \mathcal{R}(f_{\theta_k^{t-1,J}})\right] 
        \\
        \leq& \ldots
        \\
        \leq & \left[\prod_{\tau=2}^{t} \left(1-\omega_{k}^\tau\right)\right] \mathbb{E}\left[\mathrm{Regret}_{k}^{\mathrm{local}}\left(1; \theta_{k}^{1,0}, \{\eta_{k}^{1,j}\}_{j=0}^{J-1}\right)\right] 
        \\
        &+ \sum_{\tau=2}^{t} \left[ \omega_{k}^\tau \prod_{s=\tau+1}^{t} \left(1-\omega_{k}^s\right) \right] U_{k}\left(\tau; \theta_{k}^{\tau,0}, \{\eta_{k}^{\tau,j}\}_{j=0}^{J-1}\right) 
        \\
        &+ \sum_{\tau=2}^{t} \left[ \left(1-\omega_{k}^\tau\right) \prod_{s=\tau+1}^{t} \left(1-\omega_{k}^s\right) \right] \mathbb{E}_{p_{k}^{\leq \tau-1}}\left[\mathcal{R}(f_{\theta_{k}^{\tau,J}}) - \mathcal{R}(f_{\theta_{k}^{\tau-1,J}})\right]
    \end{split}
\end{equation}

Next, let $n_k^\tau$ denote the amount of new data received by client $k$ at FL round $\tau$, and define the cumulative data size over the first $\tau$ FL rounds as $N_{k}^\tau \triangleq \sum_{i=1}^{\tau} n_{k}^i$. Accordingly, we have $\omega_{k}^\tau = n_k^\tau / N_{k}^\tau$ and $1 - \omega_{k}^\tau = N_{k}^{\tau - 1} / N_{k}^\tau$. Substituting these relations, we obtain:
\begin{equation}
    \begin{split}
        &\prod_{\tau=2}^{t} \left(1-\omega_{k}^\tau\right) 
        = \frac{N_{k}^1}{N_{k}^2} \cdot \frac{N_{k}^2}{N_{k}^3} \cdots \frac{N_{k}^{t-1}}{N_{k}^t} = \frac{N_{k}^1}{N_{k}^t} = \frac{n_{k}^1}{N_{k}^t}
        \\
        &\omega_{k}^\tau \prod_{s=\tau+1}^{t} \left(1-\omega_{k}^s\right) 
        = \frac{n_{k}^\tau}{N_{k}^\tau} \cdot \frac{N_{k}^\tau}{N_{k}^t} = \frac{n_{k}^\tau}{N_{k}^t}
        \\
        &\left(1-\omega_{k}^\tau\right) \prod_{s=\tau+1}^{t} \left(1-\omega_{k}^s\right) 
        = \frac{N_{k}^{\tau-1}}{N_{k}^\tau} \cdot \frac{N_{k}^\tau}{N_{k}^t} = \frac{N_{k}^{\tau-1}}{N_{k}^t}
    \end{split}
\end{equation}

Substituting this into the expression for the local model regret, we obtain:
\begin{equation}
    \begin{split}
        \mathrm{Regret}_k^{\mathrm{local}}(t; \theta_k^{t,0}, \{\eta_k^{t,j}\}_{j=0}^{J-1})
        \leq& \frac{n_{k}^1}{N_{k}^t} \mathbb{E}\left[\mathrm{Regret}_{k}^{\mathrm{local}}\left(1; \theta_{k}^{1,0}, \{\eta_{k}^{1,j}\}_{j=0}^{J-1}\right)\right] 
        + \sum_{\tau=2}^{t} \frac{n_{k}^\tau}{N_{k}^t} U_{k}\left(\tau; \theta_{k}^{\tau,0}, \{\eta_{k}^{\tau,j}\}_{j=0}^{J-1}\right) 
        \\
        &+ \sum_{\tau=2}^{t} \frac{N_{k}^{\tau-1}}{N_{k}^t} \mathbb{E}_{p_{k}^{\leq \tau-1}}\left[\mathcal{R}(f_{\theta_{k}^{\tau,J}}) - \mathcal{R}(f_{\theta_{k}^{\tau-1,J}})\right]
    \end{split}
\end{equation}

Following the algorithm procedure, the initial model for the local training of client $k$ at round $\tau \geq 1$ of FL is given by $\theta_{k}^{\tau,0} = \theta_g^{\tau-1} = \frac{1}{K}\sum_{k'=1}^K \theta_{k'}^{\tau-1,J}$. Let $\theta_{k}^{\tau,*}$ denote the ideal saddle-point model for the local training of client $k$ at round $\tau$. Based on these definitions, we obtain:
\begin{equation}
    \begin{split}
        U_k(\tau; \theta_g^{\tau-1}, \{\eta_k^{\tau,j}\}_{j=0}^{J-1}) 
        \triangleq& \mathcal{E}_{k, \text{bound}}^{t}\left(\{\eta_k^{\tau,j}\}_{j=0}^{J-1}, \|\theta_g^{\tau-1} - \theta_k^{\tau,*}\|_2^2\right) + \left[\lambda_{\max} \left(G D_\theta + \frac{L}{2} {D_\theta}^2 \right) + L_{\mathrm{lip}} D_\theta \right]
        \\
        &+ \lambda_{\max} C_\kappa \cdot \left(2 + \left(8 L_{\mathrm{lip}} B + 4\sqrt{2\log(2/\delta)}\right) \sqrt{\frac{|\mathcal{C}_k^{\leq \tau-1}|}{M}} \right)
    \end{split}
\end{equation}

Considering this summation term in the local model regret expression and applying the bound $|\mathcal{C}_k^{\leq \tau-1}| \leq C_{\max}$, we obtain:
\begin{equation}
    \begin{split}
        \sum_{\tau=2}^{t} \frac{n_{k}^\tau}{N_{k}^t} U_k(\tau; \theta_g^{\tau-1}, \{\eta_k^{\tau,j}\}_{j=0}^{J-1}) 
        \leq& \sum_{\tau=2}^{t} \frac{n_{k}^\tau}{N_{k}^t} \mathcal{E}_{k, \text{bound}}^{t}\left(\{\eta_k^{\tau,j}\}_{j=0}^{J-1}, \|\theta_g^{\tau-1} - \theta_k^{\tau,*}\|_2^2\right) 
        \\
        &+ \sum_{\tau=2}^{t} \frac{n_{k}^\tau}{N_{k}^t} \left[\lambda_{\max} \left(G D_\theta + \frac{L}{2} {D_\theta}^2 \right) + L_{\mathrm{lip}} D_\theta \right]
        \\
        &+ \sum_{\tau=2}^{t} \frac{n_{k}^\tau}{N_{k}^t} \lambda_{\max} C_\kappa \cdot \left(2 + \left(8 L_{\mathrm{lip}} B + 4\sqrt{2\log(2/\delta)}\right) \sqrt{\frac{|\mathcal{C}_k^{\leq \tau-1}|}{M}} \right)
        \\
        \leq& \sum_{\tau=2}^{t} \frac{n_{k}^\tau}{N_{k}^t} \mathcal{E}_{k, \text{bound}}^{t}\left(\{\eta_k^{\tau,j}\}_{j=0}^{J-1}, \|\theta_g^{\tau-1} - \theta_k^{\tau,*}\|_2^2\right) 
        \\
        &+ \lambda_{\max} \left(G D_\theta + \frac{L}{2} {D_\theta}^2 \right) + L_{\mathrm{lip}} D_\theta 
        \\
        &+ \lambda_{\max} C_\kappa \cdot \left(2 + \left(8 L_{\mathrm{lip}} B + 4\sqrt{2\log(2/\delta)}\right) \sqrt{\frac{C_{\max}}{M}} \right)
    \end{split}
\end{equation}

Substituting this into the expression for the local model regret, we obtain:
\begin{equation}
    \begin{split}
        \mathrm{Regret}_k^{\mathrm{local}}(t; \theta_g^{t-1}, \{\eta_k^{t,j}\}_{j=0}^{J-1})
        \leq& \frac{n_{k}^1}{N_{k}^t} \mathrm{Regret}_{k}^{\mathrm{local}}\left(1; \theta_g^{0}, \{\eta_{k}^{1,j}\}_{j=0}^{J-1}\right) 
        \\
        &+ \sum_{\tau=2}^{t} \frac{n_{k}^\tau}{N_{k}^t} \mathcal{E}_{k, \text{bound}}^{t}\left(\{\eta_k^{\tau,j}\}_{j=0}^{J-1}, \|\theta_g^{\tau-1} - \theta_k^{\tau,*}\|_2^2\right) 
        \\
        &+ \sum_{\tau=2}^{t} \frac{N_{k}^{\tau-1}}{N_{k}^t} \mathbb{E}_{p_{k}^{\leq \tau-1}}\left[\mathcal{R}(f_{\theta_{k}^{\tau,J}}) - \mathcal{R}(f_{\theta_{k}^{\tau-1,J}})\right]
        + \lambda_{\max} \left(G D_\theta + \frac{L}{2} {D_\theta}^2 \right) + L_{\mathrm{lip}} D_\theta 
        \\
        &+ \lambda_{\max} C_\kappa \cdot \left(2 + \left(8 L_{\mathrm{lip}} B + 4\sqrt{2\log(2/\delta)}\right) \sqrt{\frac{C_{\max}}{M}} \right)
    \end{split}
\end{equation}

Next, we bound the summation term $\sum_{\tau=2}^{t} \frac{N_{k}^{\tau-1}}{N_{k}^t} \mathbb{E}_{p_{k}^{\leq \tau-1}}\left[\mathcal{R}(f_{\theta_{k}^{\tau,J}}) - \mathcal{R}(f_{\theta_{k}^{\tau-1,J}})\right]$. By Assumption 5 and Lemma 1 in \cite{13}, we obtain:
\begin{equation}
    \mathbb{E}_{p_{k}^{\leq \tau-1}}\left[\mathcal{R}(f_{\theta_{k}^{\tau,J}}) - \mathcal{R}(f_{\theta_{k}^{\tau-1,J}})\right] 
    \leq \sup_{f \in \mathcal{F}} \left| \mathcal{R}_{p_{k}^{\leq \tau}}(f) - \mathcal{R}_{p_{k}^{\leq \tau-1}}(f) \right| \leq \|p_{k}^{\leq \tau} - p_{k}^{\leq \tau-1}\|_{\mathrm{TV}}
\end{equation}

Substituting the expanded form of the cumulative data distribution $p_{k}^{\leq \tau} = \sum_{i=1}^\tau \frac{n_{k}^i}{N_{k}^\tau} p_{k}^i = \frac{n_{k}^\tau}{N_{k}^\tau} p_{k}^\tau + \frac{N_{k}^{\tau-1}}{N_{k}^\tau} p_{k}^{\leq \tau-1}$, and applying the convexity and triangle inequality of the Total Variation (TV) distance, we obtain:
\begin{equation}
    \begin{split}
        \left\|p_{k}^{\leq \tau} - p_{k}^{\leq \tau-1}\right\|_{\mathrm{TV}}
        =& \left\| \frac{n_{k}^\tau}{N_{k}^\tau} p_{k}^\tau + \frac{N_{k}^{\tau-1}}{N_{k}^\tau} p_{k}^{\leq \tau-1} - p_{k}^{\leq \tau-1} \right\|_{\mathrm{TV}}
        = \left\|\frac{n_{k}^\tau}{N_{k}^\tau} \left( p_{k}^\tau - p_{k}^{\leq \tau-1} \right) \right\|_{\mathrm{TV}}
        \\
        =& \frac{n_{k}^\tau}{N_{k}^\tau} \left\| p_{k}^\tau - \sum_{j=1}^{\tau-1} \frac{n_{k}^j}{N_{k}^{\tau-1}} p_{k}^j \right\|_{\mathrm{TV}}
        = \frac{n_{k}^\tau}{N_{k}^\tau} \left\| \sum_{j=1}^{\tau-1} \frac{n_{k}^j}{N_{k}^{\tau-1}} \left(p_{k}^\tau -  p_{k}^j\right) \right\|_{\mathrm{TV}}
        \\
        \leq& \frac{n_{k}^\tau}{N_{k}^\tau N_{k}^{\tau-1}} \sum_{j=1}^{\tau-1} n_{k}^j \left\| p_{k}^\tau -  p_{k}^j \right\|_{\mathrm{TV}}
        \leq \frac{n_{k}^\tau}{N_{k}^\tau N_{k}^{\tau-1}} \sum_{j=1}^{\tau-1} n_{k}^j \left( \sum_{i=j+1}^\tau \|p_{k}^i - p_{k}^{i-1}\|_{\mathrm{TV}} \right)
    \end{split}
\end{equation}

Furthermore, according to Assumption 8, we have $D_{\mathrm{TV}}(p_{k}^i \| p_{k}^{i-1}) = \|p_{k}^i - p_{k}^{i-1}\|_{\mathrm{TV}} \leq \frac{C_{\max}}{i^\alpha}$. Substituting this into the above inequality, we obtain:
\begin{equation}
    \begin{split}
        \left\|p_{k}^{\leq \tau} - p_{k}^{\leq \tau-1}\right\|_{\mathrm{TV}}
        \leq \frac{n_{k}^\tau}{N_{k}^\tau N_{k}^{\tau-1}} \sum_{j=1}^{\tau-1} n_{k}^j \left( \sum_{i=j+1}^\tau \frac{C_{\max}}{i^\alpha} \right)
    \end{split}
\end{equation}

It follows that:
\begin{equation}
    \begin{split}
        \sum_{\tau=2}^{t} \frac{N_{k}^{\tau-1}}{N_{k}^t} \mathbb{E}_{p_{k}^{\leq \tau-1}}\left[\mathcal{R}(f_{\theta_{k}^{\tau,J}}) - \mathcal{R}(f_{\theta_{k}^{\tau-1,J}})\right]
        \leq& \sum_{\tau=2}^{t} \frac{N_{k}^{\tau-1}}{N_{k}^t} \frac{n_{k}^\tau}{N_{k}^\tau N_{k}^{\tau-1}} \sum_{j=1}^{\tau-1} n_{k}^j \left( \sum_{i=j+1}^\tau \frac{C_{\max}}{i^\alpha} \right)
        \\
        =& \sum_{\tau=2}^{t} \frac{n_{k}^\tau}{N_{k}^t N_{k}^\tau} \left( \sum_{j=1}^{\tau-1} n_{k}^j  \sum_{i=j+1}^\tau \frac{C_{\max}}{i^\alpha} \right)
        \\
        =& \sum_{\tau=2}^{t} \frac{n_{k}^\tau}{N_{k}^t N_{k}^\tau} \left( C_{\max} \sum_{i=2}^\tau \frac{1}{i^\alpha} \sum_{j=1}^{i-1} n_{k}^j \right)
        \\
        =& \frac{C_{\max}}{N_{k}^t} \sum_{\tau=2}^{t} \frac{n_{k}^\tau}{N_{k}^\tau} \sum_{i=2}^\tau \frac{N_{k}^{i-1}}{i^\alpha}
    \end{split}
\end{equation}

Next, analogously to the proof of Lemma 2, we define the maximum and minimum single-round data sizes across all clients over all $T$ rounds of FL as $n_{\textrm{max}} \triangleq \max_{k' \in \mathcal{K},\tau \leq T}(n_{k'}^\tau)$ and $n_{\textrm{min}} \triangleq \min_{k' \in \mathcal{K},\tau \leq T}(n_{k'}^\tau)$, respectively. Letting $\nu = n_{\min}/n_{\max} \leq 1$, we obtain:
\begin{equation}
    \begin{split}
        \sum_{\tau=2}^{t} \frac{N_{k}^{\tau-1}}{N_{k}^t} \mathbb{E}_{p_{k}^{\leq \tau-1}}\left[\mathcal{R}(f_{\theta_{k}^{\tau,J}}) - \mathcal{R}(f_{\theta_{k}^{\tau-1,J}})\right]
        \leq& \frac{C_{\max}}{N_{k}^t} \sum_{\tau=2}^{t} \frac{n_{k}^\tau}{N_{k}^\tau} \sum_{i=2}^\tau \frac{N_{k}^{i-1}}{i^\alpha}
        \\
        \leq& \frac{C_{\max}}{t \cdot n_{\min}} \sum_{\tau=2}^{t} \frac{n_{\max}}{\tau \cdot n_{\min}} \sum_{i=2}^\tau \frac{(i-1) \cdot n_{\max}}{i^\alpha}
        \\
        \leq& \frac{C_{\max} \cdot n_{\max}}{t \cdot n_{\min}} \sum_{\tau=2}^{t} \frac{n_{\max}}{\tau \cdot n_{\min}} \sum_{i=2}^\tau \frac{i}{i^\alpha}
        \\
        =& \frac{C_{\max}}{\nu^2} \frac{1}{t} \sum_{\tau=2}^{t} \frac{1}{\tau} \sum_{i=2}^\tau  i^{1 - \alpha}
    \end{split}
\end{equation}

For the summation term $\frac{1}{t} \sum_{\tau=2}^{t} \frac{1}{\tau} \sum_{i=2}^\tau  i^{1 - \alpha}$, when $\alpha > 0$ and $t \geq 2$, we obtain the following piecewise upper bound $\mathcal{A}(t, \alpha)$:
\begin{equation}
    \frac{1}{t} \sum_{\tau=2}^{t} \frac{1}{\tau} \sum_{i=2}^\tau  i^{1 - \alpha}
    \leq \mathcal{A}(t, \alpha)
    \triangleq 
    \begin{cases} 
        \dfrac{t^{1-\alpha}}{(2-\alpha)^2} + \dfrac{t^{-\alpha}}{1-\alpha} + \dfrac{1}{t}, & 0 < \alpha < 1 \\[12pt]
        1, & \alpha = 1 \\[12pt]
        \dfrac{t^{1-\alpha}}{(2-\alpha)^2} + \dfrac{1}{(\alpha-1)t} + \dfrac{1}{t}, & 1 < \alpha < 2 \\[12pt]
        \dfrac{(\ln t)^2 + 2\ln t}{2t}, & \alpha = 2 \\[12pt]
        \dfrac{\alpha-1}{(\alpha-2)t}(\ln t + 1), & \alpha > 2
    \end{cases}
\end{equation}

It is easy to see that $\mathcal{A}(t, \alpha)$ is monotonically decreasing in $\alpha$, monotonically increasing in $t$ when $\alpha \in (0, 1)$, and monotonically decreasing in $t$ when $\alpha > 1$. Substituting this into the original expression, we obtain:
\begin{equation}
    \sum_{\tau=2}^{t} \frac{N_{k}^{\tau-1}}{N_{k}^t} \mathbb{E}_{p_{k}^{\leq \tau-1}}\left[\mathcal{R}(f_{\theta_{k}^{\tau,J}}) - \mathcal{R}(f_{\theta_{k}^{\tau-1,J}})\right]
    \leq \frac{C_{\max}}{\nu^2} \mathcal{A}(t, \alpha)
\end{equation}

Substituting this into the expression for the local model regret, we obtain:
\begin{equation}
    \begin{split}
        \mathrm{Regret}_k^{\mathrm{local}}(t; \theta_g^{t-1}, \{\eta_k^{t,j}\}_{j=0}^{J-1})
        \leq& \frac{1}{\nu t} \mathrm{Regret}_{k}^{\mathrm{local}}\left(1; \theta_g^{0}, \{\eta_{k}^{1,j}\}_{j=0}^{J-1}\right) 
        \\
        &+ \frac{1}{\nu t} \sum_{\tau=2}^{t} \mathcal{E}_{k, \text{bound}}^{t}\left(\{\eta_k^{\tau,j}\}_{j=0}^{J-1}, \|\theta_g^{\tau-1} - \theta_k^{\tau,*}\|_2^2\right) 
        \\
        &+ \frac{C_{\max}}{\nu^2} \mathcal{A}(t, \alpha)
        + \lambda_{\max} \left(G D_\theta + \frac{L}{2} {D_\theta}^2 \right) + L_{\mathrm{lip}} D_\theta 
        \\
        &+ \lambda_{\max} C_\kappa \cdot \left(2 + \left(8 L_{\mathrm{lip}} B + 4\sqrt{2\log(2/\delta)}\right) \sqrt{\frac{C_{\max}}{M}} \right)
    \end{split}
\end{equation}

To simplify the notation, we define the distribution shift functional $\Delta_{\mathrm{shift}}(t, C_{\max}, \nu, \alpha)$ and the buffer optimization error functional $\Xi_{\mathrm{buffer}}(C_\kappa, C_{\max}, M)$ as follows:
\begin{equation}
    \begin{split}
        \Delta_{\mathrm{shift}}(t, C_{\max}, \nu, \alpha)
        \triangleq & \frac{C_{\max}}{\nu^2} \cdot 
        \begin{cases} 
            \dfrac{t^{1-\alpha}}{(2-\alpha)^2} + \dfrac{t^{-\alpha}}{1-\alpha} + \dfrac{1}{t}, & 0 < \alpha < 1 \\[12pt]
            1, & \alpha = 1 \\[12pt]
            \dfrac{t^{1-\alpha}}{(2-\alpha)^2} + \dfrac{1}{(\alpha-1)t} + \dfrac{1}{t}, & 1 < \alpha < 2 \\[12pt]
            \dfrac{(\ln t)^2 + 2\ln t}{2t}, & \alpha = 2 \\[12pt]
            \dfrac{\alpha-1}{(\alpha-2)t}(\ln t + 1), & \alpha > 2
        \end{cases}
    \end{split}
\end{equation}
\begin{equation}
    \begin{split}
        \Xi_{\mathrm{buffer}}(C_\kappa, C_{\max}, M) 
        \triangleq & L_{\mathrm{lip}} D_\theta + \lambda_{\max} \left(G D_\theta + \frac{L}{2} {D_\theta}^2 \right) 
        \\
        &+ \lambda_{\max} C_\kappa \cdot \left(2 + \left(8 L_{\mathrm{lip}} B + 4\sqrt{2\log(2/\delta)}\right) \sqrt{\frac{C_{\max}}{M}} \right)
    \end{split}
\end{equation}

Consequently, the expression for the local model regret can be compactly written as:
\begin{equation}
    \begin{split}
        \mathrm{Regret}_k^{\mathrm{local}}(t; \theta_g^{t-1}, \{\eta_k^{t,j}\}_{j=0}^{J-1})
        \leq& \frac{1}{\nu t} \left[\mathrm{Regret}_{k}^{\mathrm{local}}\left(1; \theta_g^{0}, \{\eta_{k}^{1,j}\}_{j=0}^{J-1}\right) 
        + \sum_{\tau=2}^{t} \mathcal{E}_{k, \text{bound}}^{t}\left(\{\eta_k^{\tau,j}\}_{j=0}^{J-1}, \|\theta_g^{\tau-1} - \theta_k^{\tau,*}\|_2^2\right)
        \right]
        \\
        &+ \Delta_{\mathrm{shift}}(t, C_{\max}, \nu, \alpha) 
        + \Xi_{\mathrm{buffer}}(C_\kappa, C_{\max}, M) 
    \end{split}
\end{equation}

Substituting this into the expression for the global model regret given in Lemma 3, we obtain:
\begin{equation}
    \begin{split}
        \mathrm{Regret}_k^{\mathrm{global}}(t; \theta_g^{t-1}, \mathcal{I}_t)
        \leq& \frac{1}{K}\sum_{k'=1}^K \frac{1}{\nu t} \mathrm{Regret}_{k'}^{\mathrm{local}}\left(1; \theta_g^{0}, \{\eta_{k'}^{1,j}\}_{j=0}^{J-1}\right)  
        \\
        &+ \frac{1}{K}\sum_{k'=1}^K \frac{1}{\nu t} \sum_{\tau=2}^{t} \mathcal{E}_{k', \text{bound}}^{t}\left(\{\eta_{k'}^{\tau,j}\}_{j=0}^{J-1}, \|\theta_g^{\tau-1} - \theta_{k'}^{\tau,*}\|_2^2\right) 
        \\
        &+ \Delta_{\mathrm{shift}}(t, C_{\max}, \nu, \alpha) 
        + \Xi_{\mathrm{buffer}}(C_\kappa, C_{\max}, M) 
        \\
        &+ \frac{1}{K}\sum_{k'=1, k' \neq k}^K \mathbb{E}_{p_{k}^{\leq t}}[\mathcal{R}(f_{k',\mathrm{cum}}^{t,*}) - \mathcal{R}(f_{k,\mathrm{cum}}^{t,*})]
        \\
        &+ \frac{2(K-1)}{K} \Delta^{\mathrm{TV}}(t, \nu) - \frac{\mu}{2K}\sum_{k'=1}^K \|\theta_{k'}^{t,J} - \theta_g^t\|_2^2
    \end{split}
\end{equation}

Consider the summation term $\frac{1}{K}\sum_{k'=1, k' \neq k}^K \mathbb{E}_{p_{k}^{\leq t}}[\mathcal{R}(f_{k',\mathrm{cum}}^{t,*}) - \mathcal{R}(f_{k,\mathrm{cum}}^{t,*})]$. For the summand $\mathbb{E}_{p_{k}^{\leq t}}[\mathcal{R}(f_{k',\mathrm{cum}}^{t,*}) - \mathcal{R}(f_{k,\mathrm{cum}}^{t,*})]$, analogously to the previous proof, we have:
\begin{equation}
    \begin{split}
        \mathbb{E}_{p_{k}^{\leq t}}[\mathcal{R}(f_{k',\mathrm{cum}}^{t,*}) - \mathcal{R}(f_{k,\mathrm{cum}}^{t,*})]
        =& \mathbb{E}_{p_{k}^{\leq t}}[\mathcal{R}(f_{k',\mathrm{cum}}^{t,*})] - \mathbb{E}_{p_{k}^{\leq t}}[\mathcal{R}(f_{k,\mathrm{cum}}^{t,*})]
        \\
        =& \left[\mathbb{E}_{p_{k}^{\leq t}}[\mathcal{R}(f_{k',\mathrm{cum}}^{t,*})] - \mathbb{E}_{p_{k'}^{\leq t}}[\mathcal{R}(f_{k',\mathrm{cum}}^{t,*})]\right] 
        + \left[\mathbb{E}_{p_{k'}^{\leq t}}[\mathcal{R}(f_{k',\mathrm{cum}}^{t,*})] - \mathbb{E}_{p_{k'}^{\leq t}}[\mathcal{R}(f_{k,\mathrm{cum}}^{t,*})]\right] 
        \\
        &+ \left[\mathbb{E}_{p_{k'}^{\leq t}}[\mathcal{R}(f_{k,\mathrm{cum}}^{t,*})] - \mathbb{E}_{p_{k}^{\leq t}}[\mathcal{R}(f_{k,\mathrm{cum}}^{t,*})]\right]
        \\
        \leq& \|p_k^{\leq t} - p_{k'}^{\leq t}\|_{\mathrm{TV}} + 0 + \|p_{k'}^{\leq t} - p_{k}^{\leq t}\|_{\mathrm{TV}}
        \\
        \leq& 2 \cdot \Delta^{\mathrm{TV}}(t, \nu)
    \end{split}
\end{equation}

Substituting this back into the expression for the global model regret, we obtain:
\begin{equation}
    \begin{split}
        \mathrm{Regret}_k^{\mathrm{global}}(t; \theta_g^{t-1}, \mathcal{I}_t)
        \leq& \frac{1}{K}\sum_{k'=1}^K \frac{1}{\nu t} \left[ \mathrm{Regret}_{k'}^{\mathrm{local}}\left(1; \theta_g^{0}, \{\eta_{k'}^{1,j}\}_{j=0}^{J-1}\right) + \sum_{\tau=2}^{t} \mathcal{E}_{k', \text{bound}}^{t}\left(\{\eta_{k'}^{\tau,j}\}_{j=0}^{J-1}, \|\theta_g^{\tau-1} - \theta_{k'}^{\tau,*}\|_2^2\right)
        \right]
        \\
        &+ \Delta_{\mathrm{shift}}(t, C_{\max}, \nu, \alpha) 
        + \Xi_{\mathrm{buffer}}(C_\kappa, C_{\max}, M) 
        \\
        &+ \frac{4(K-1)}{K} \Delta^{\mathrm{TV}}(t, \nu) - \frac{\mu}{2K}\sum_{k'=1}^K \|\theta_{k'}^{t,J} - \theta_g^t\|_2^2
    \end{split}
\end{equation}

Next, we discuss the evolution of the local and global regrets with respect to $t$. We define the regret difference function $\Delta_{k, \mathrm{regret}}(t)$ as follows:
\begin{equation}
    \begin{split}
        \Delta_{k, \mathrm{regret}}(t) 
        =& \mathrm{Regret}_k^{\mathrm{global}}(t; \theta_g^{t-1}, \mathcal{I}_t) - \mathrm{Regret}_k^{\mathrm{local}}(t; \theta_g^{t-1}, \{\eta_k^{t,j}\}_{j=0}^{J-1})
        \\
        =& \frac{1}{K}\sum_{k'=1}^K \frac{1}{\nu t} \left[ \mathrm{Regret}_{k'}^{\mathrm{local}}\left(1; \theta_g^{0}, \{\eta_{k'}^{1,j}\}_{j=0}^{J-1}\right) + \sum_{\tau=2}^{t} \mathcal{E}_{k', \text{bound}}^{t}\left(\{\eta_{k'}^{\tau,j}\}_{j=0}^{J-1}, \|\theta_g^{\tau-1} - \theta_{k'}^{\tau,*}\|_2^2\right)
        \right] 
        \\
        &- \frac{1}{\nu t} \left[\mathrm{Regret}_{k}^{\mathrm{local}}\left(1; \theta_g^{0}, \{\eta_{k}^{1,j}\}_{j=0}^{J-1}\right) 
        + \sum_{\tau=2}^{t} \mathcal{E}_{k, \text{bound}}^{t}\left(\{\eta_k^{\tau,j}\}_{j=0}^{J-1}, \|\theta_g^{\tau-1} - \theta_k^{\tau,*}\|_2^2\right)
        \right]
        \\
        &+ \frac{4(K-1)}{K} \Delta^{\mathrm{TV}}(t, \nu) - \frac{\mu}{2K}\sum_{k'=1}^K \|\theta_{k'}^{t,J} - \theta_g^t\|_2^2
        \\
        \triangleq& \Delta_{k, \mathrm{opt}}(\mathcal{K}, \nu, t) + \frac{4(K-1)}{K} \Delta^{\mathrm{TV}}(t, \nu) - \frac{\mu}{2K}\sum_{k'=1}^K \|\theta_{k'}^{t,J} - \theta_g^t\|_2^2
    \end{split}
\end{equation}

where $\Delta_{k, \mathrm{opt}}(\mathcal{K}, \nu, t)$ captures the fluctuations induced by stochastic optimization. It is easy to proof that $\mathbb{E}_{k \in \mathcal{K}}\left[\Delta_{k, \mathrm{opt}}(\mathcal{K}, \nu, t)\right] = 0$, which yields:
\begin{equation}
    \begin{split}
        \mathbb{E}_{k \in \mathcal{K}}\left[\Delta_{k, \mathrm{regret}}(t)\right]
        &= \mathbb{E}_{k \in \mathcal{K}}\left[\Delta_{k, \mathrm{opt}}(\mathcal{K}, \nu, t)\right] + \frac{4(K-1)}{K} \Delta^{\mathrm{TV}}(t, \nu) - \frac{\mu}{2K}\sum_{k'=1}^K \|\theta_{k'}^{t,J} - \theta_g^t\|_2^2
        \\
        &= \frac{4(K-1)}{K} \Delta^{\mathrm{TV}}(t, \nu) - \frac{\mu}{2K}\sum_{k'=1}^K \|\theta_{k'}^{t,J} - \theta_g^t\|_2^2
    \end{split}
\end{equation}

where: 
\begin{equation}
    \Delta^{\mathrm{TV}}(t, \nu) 
    \triangleq \frac{1}{(t+1)^{\nu}} \left(2^{\nu} + \frac{3}{\nu} \left[\frac{(3/2)^{\nu}}{2} + \frac{{\nu} 2^{{\nu}-1} }{1 - {\nu}}\right] \right) + \frac{3}{\nu^2}
\end{equation}

When $t$ is small, $\mathbb{E}_{k \in \mathcal{K}}\left[\Delta_{k, \mathrm{regret}}(t)\right]$ is dominated by $\Delta^{\mathrm{TV}}(t, \nu)$, meaning that the negative impact caused by the cross-client data distribution discrepancy far outweighs the parameter divergence penalty brought by model aggregation. This yields $\mathrm{Regret}_k^{\mathrm{global}}(t; \theta_g^{t-1}, \mathcal{I}_t) \geq \mathrm{Regret}_k^{\mathrm{local}}(t; \theta_g^{t-1}, \{\eta_k^{t,j}\}_{j=0}^{J-1})$, indicating that the local model $\theta_{k}^{t,J}$ outperforms the global model $\theta_{g}^{t}$ on client $k$.

When $t$ is large, $\Delta^{\mathrm{TV}}(t, \nu)$ decays significantly and converges to the constant $3/\nu^2$. At this stage, due to the inconsistent buffer data distributions across clients, the local models inevitably overfit to their respective local buffers, resulting in an irremovable steady-state model divergence (client drift). Consequently, the parameter divergence penalty from model aggregation, $- \frac{\mu}{2K}\sum_{k'=1}^K \|\theta_{k'}^{t,J} - \theta_g^t\|_2^2$, dominates in magnitude. This yields $\mathrm{Regret}_k^{\mathrm{global}}(t; \theta_g^{t-1}, \mathcal{I}_t) \leq \mathrm{Regret}_k^{\mathrm{local}}(t; \theta_g^{t-1}, \{\eta_k^{t,j}\}_{j=0}^{J-1})$, indicating that the global model $\theta_{g}^{t}$ outperforms the local model $\theta_{k}^{t,J}$ on client $k$.

This completes the proof of Theorem 3.

\end{document}